%% file: ms.tex
\documentclass[10pt,twocolumn,letterpaper]{article}

\usepackage{iccv}
\usepackage{times}
\usepackage{epsfig}
\usepackage{graphicx}
\usepackage{amsmath}
\usepackage{amssymb}
\usepackage{placeins}

\usepackage{makecell}
\usepackage{booktabs}
\usepackage{capt-of}
\usepackage{paralist}

\usepackage{mathtools}
\usepackage[pagebackref=true,breaklinks=true,letterpaper=true,colorlinks,bookmarks=false]{hyperref}

\iccvfinalcopy %

\def\iccvPaperID{****} %

\begin{document}

\title{Structure and Content-Guided Video Synthesis with Diffusion Models}

\author{
  Patrick Esser
  \qquad\qquad Johnathan Chiu
  \qquad\qquad Parmida Atighehchian \\
  Jonathan Granskog
  \qquad\qquad Anastasis Germanidis \\
  Runway \\
  \url{https://research.runwayml.com/gen1}
}

\twocolumn[{%
\maketitle%
\includegraphics[width=\textwidth]{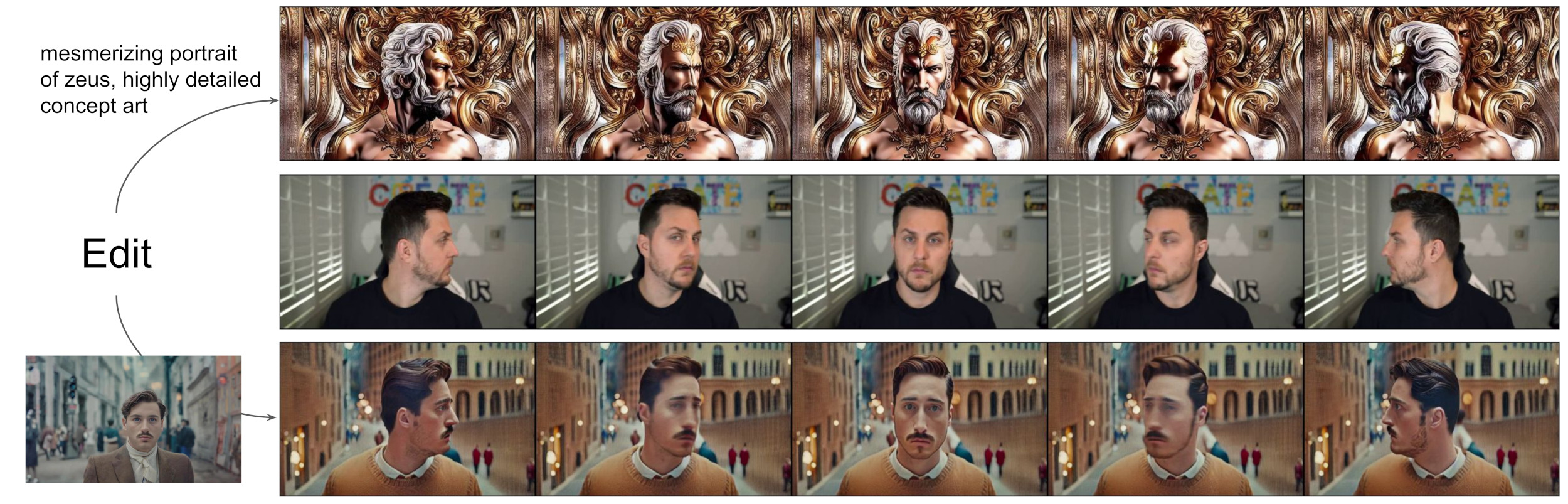}
\captionof{figure}{\textbf{Guided Video Synthesis} We present an approach based
on latent video diffusion models that synthesizes videos (top and bottom) guided by
content described through text (top) or images (bottom) while keeping the
structure of an input video (middle).}
\label{fig:teaser}
\vspace{2em}
}]

\begin{abstract}
Text-guided generative diffusion models unlock powerful image creation and editing tools.
While these have been extended to video generation, current approaches that
edit the content of existing footage while retaining structure
require expensive re-training for every
input or rely on error-prone propagation of image edits across frames.

In this work, we present a structure and content-guided video diffusion model
that edits videos based on visual or textual descriptions of the desired output.
Conflicts between user-provided content edits and structure representations occur due to
insufficient disentanglement between the two aspects.
As a solution, we show that training on monocular depth estimates with varying levels of detail
provides control over structure and content fidelity.
Our model is trained jointly on images and videos
which also exposes explicit control of temporal consistency through a novel guidance method.
Our experiments demonstrate a wide variety of successes; fine-grained control
over output characteristics, customization based on a few reference images, and
a strong user preference towards results by our model.
\end{abstract}

\newcommand{\RR}{\mathbb{R}}

\input{figcommands}
\input{tabcommands}
\input{macros}

\input{introduction}

\input{related-work}

\input{method}

\input{results}

\input{conclusion}
{\small
\bibliographystyle{ieee_fullname}
\bibliography{ms}
}

\input{supp}

\end{document}

%% file: figcommands.tex
\providecommand{\impath}[1]{}
\providecommand{\imwidth}{}
\providecommand{\imwidthB}{}
\providecommand{\imheight}{}
\providecommand{\cellwidth}{}

\newcommand{\figteaser}{%
  \centering%
    \includegraphics[width=0.5\textwidth]{figures/firstpage.jpg}%
}

\newcommand{\figcomparative}{
  \begin{figure*}
    \includegraphics[width=\textwidth]{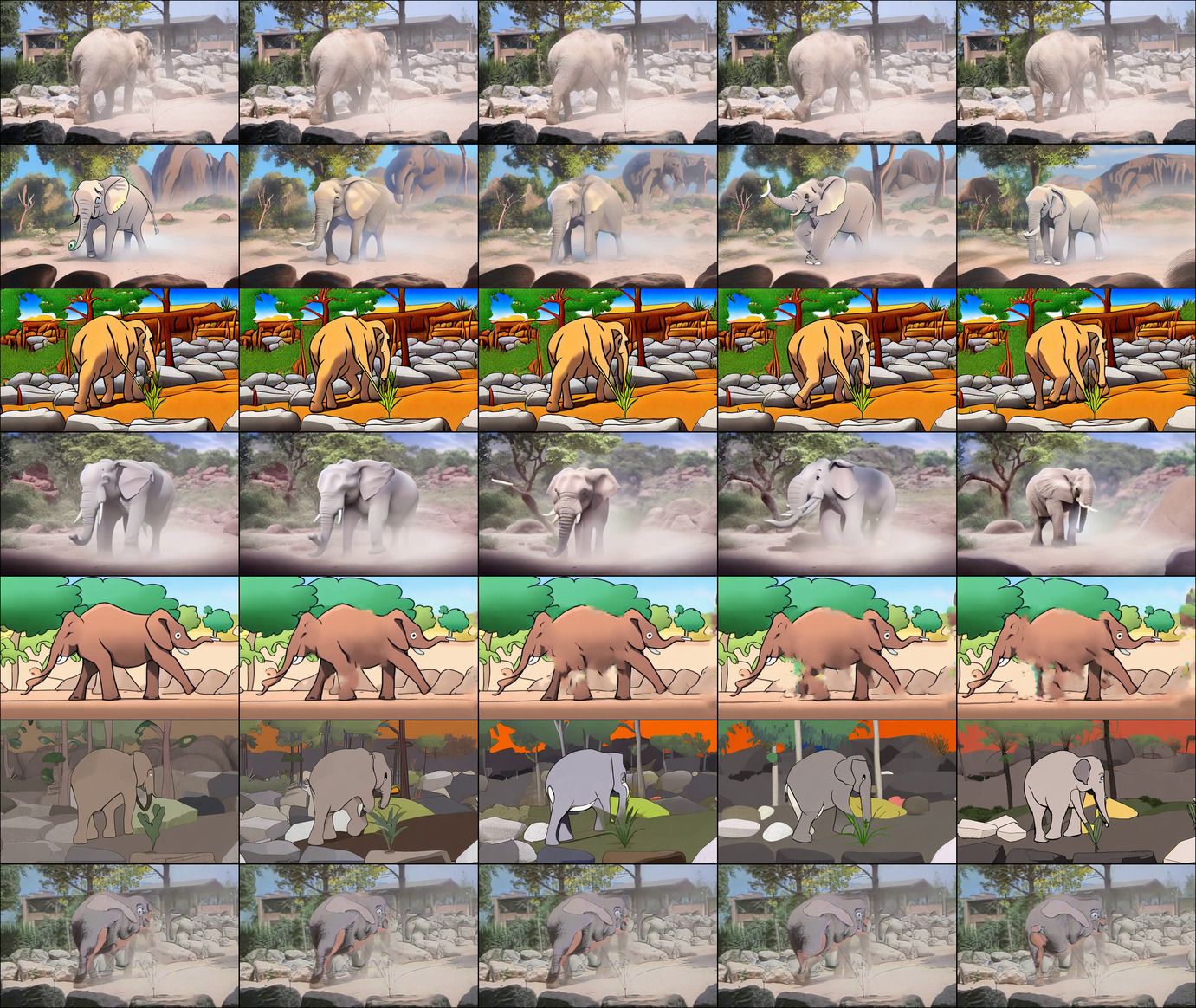}
  \caption{Visual comparison between evaluated methods. From top to bottom:
    input, Deforum, ours, SDEdit, IVS, Depth-SD, Text2Live.}
  \label{fig:compare}
  \end{figure*}
}

\newcommand{\figmaskededit}{
\begin{figure*}
  \centering
    \renewcommand{\cellwidth}{0.1\textwidth}
    \renewcommand{\imwidth}{0.8\textwidth}
    \begin{tabular}{cc}
    \parbox{\cellwidth}{\footnotesize Input} &
    \parbox{\imwidth}{\includegraphics[width=0.8\textwidth]{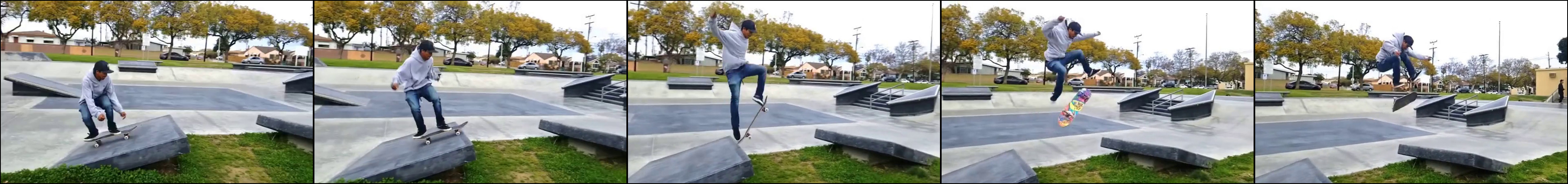}}
    \\
    \parbox{\cellwidth}{\footnotesize Mask} &
    \parbox{\imwidth}{\includegraphics[width=0.8\textwidth]{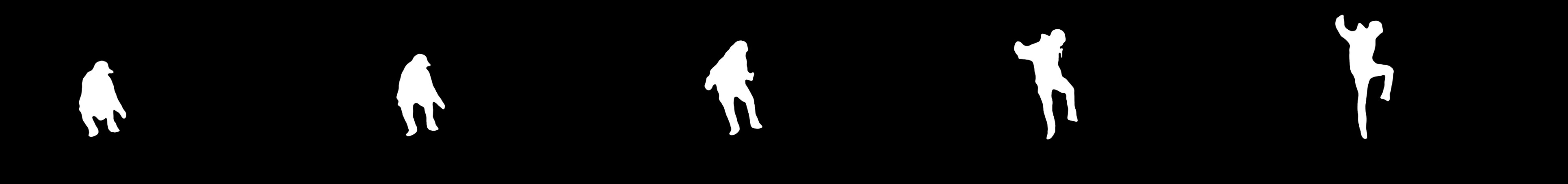}}
    \\
    \parbox{\cellwidth}{\footnotesize A snowboarder in a snow park on the mountain} &
    \parbox{\imwidth}{\includegraphics[width=0.8\textwidth]{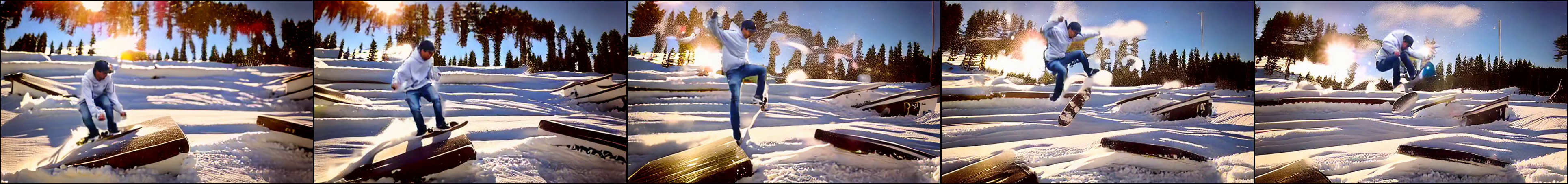}}
      \vspace{0.5em}
  \end{tabular}
  \caption{\textbf{Background Editing:} Masking the denoising process allows us
  to restrict edits to backgrounds for more control over results.}
  \label{fig:maskededit}
\end{figure*}%
}

\newcommand{\figarchitecture}{%
\begin{figure*}
  \includegraphics[width=\textwidth,trim={0.15cm 2.75cm 0.15cm 7.0cm},clip]{figures/architecture.pdf}
  \caption{Architecture.}
  \label{fig:modeloverview}
\end{figure*}
}

\newcommand{\figinference}{%
\begin{figure}
  \includegraphics[width=\columnwidth,trim={4.0cm 5.0cm 5.5cm 7.0cm},clip]{figures/inference.pdf}
  \caption{Inference}
  \label{fig:inferenceoverview}
\end{figure}
}

\newcommand{\figoverview}{%
\begin{figure*}
  \includegraphics[width=\textwidth]{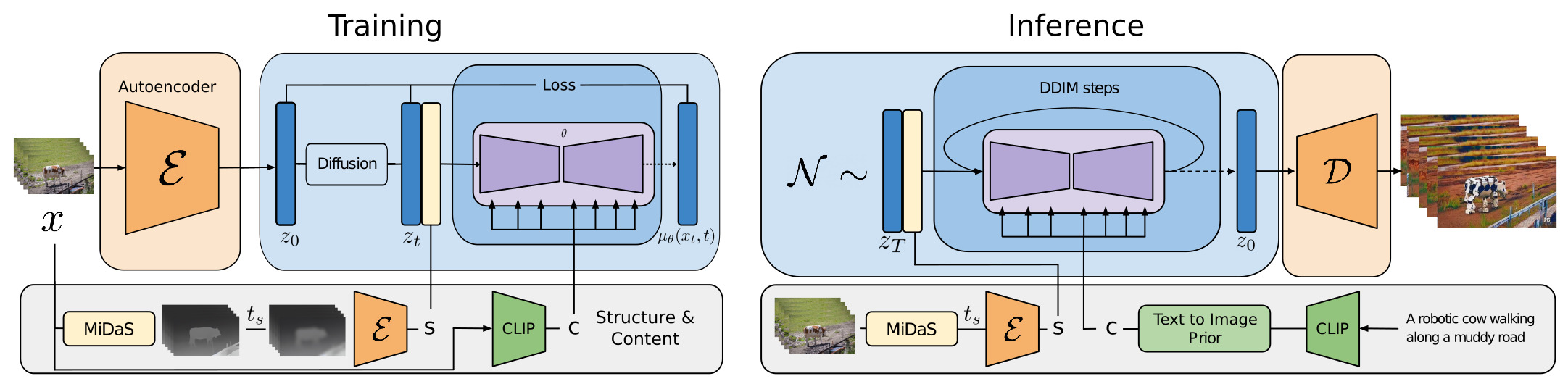}
  \caption{\textbf{Overview:} During training (left), input videos $x$ are encoded to $z_0$ with a fixed
  encoder $\mathcal{E}$ and diffused to $z_t$. We extract a structure
  representation $s$ by encoding depth maps obtained with MiDaS, and a content
  representation $c$ by encoding one of the frames with CLIP. The model then
  learns to reverse the diffusion process in the latent space, with the help of
  $s$, which gets concatenated to $z_t$, as well as $c$, which is provided via
  cross-attention blocks. During inference (right), the structure $s$ of an
  input video is provided in the same manner. To specify content via text, we
  convert CLIP text embeddings to image embeddings via a prior.}
  \label{fig:overview}
\end{figure*}
}

\newcommand{\figblocks}{%
\begin{figure}
  \includegraphics[width=\columnwidth]{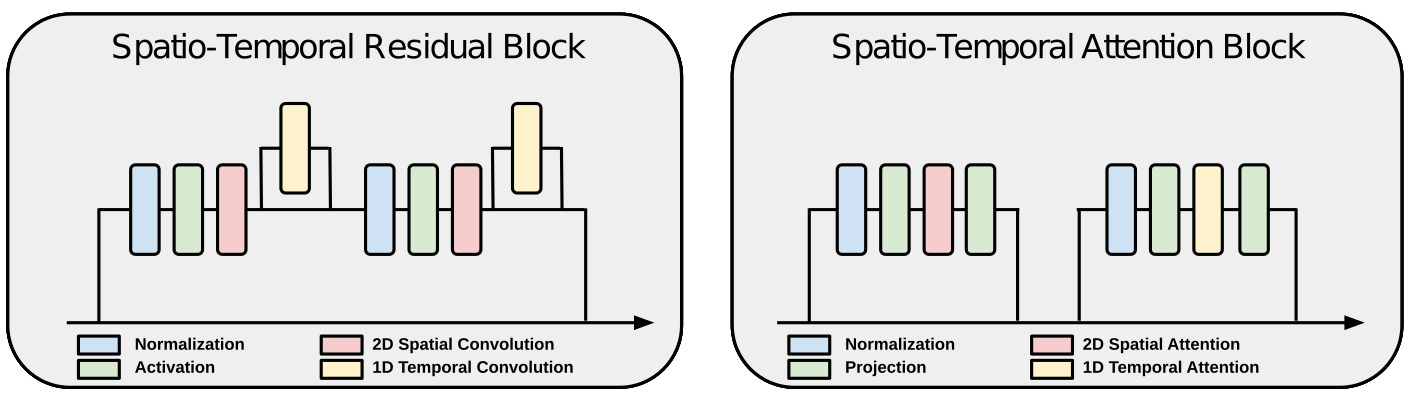}
  \caption{\textbf{Temporal Extension:} We extend an image-based UNet
  architecture to videos, by adding temporal layers in its building blocks. We
  add a 1D temporal convolution after each 2D spatial convolution in its
  residual blocks (left), and we add a 1D temporal attention block after each
  of its 2D spatial attention blocks (right).}
  \label{fig:blocks}
\end{figure}
}

\newcommand{\figtempcontrol}{%
\begin{figure*}
    \includegraphics[width=0.4\textwidth]{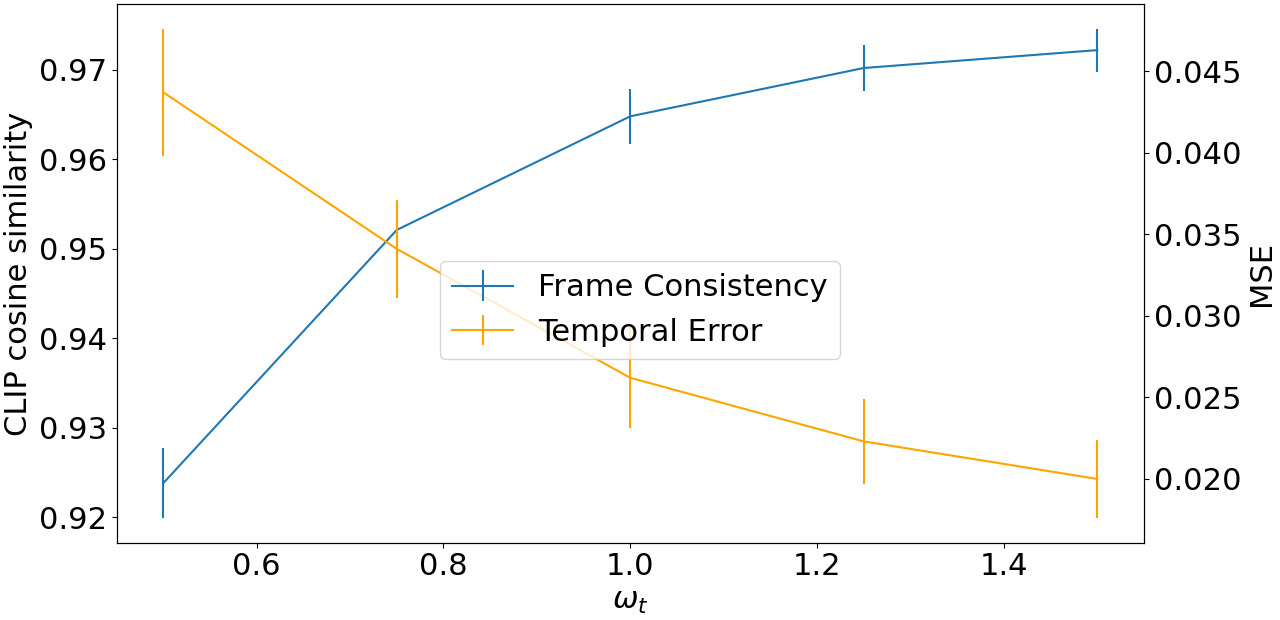}\vspace{-1em}\hfill
    \includegraphics[width=0.55\textwidth]{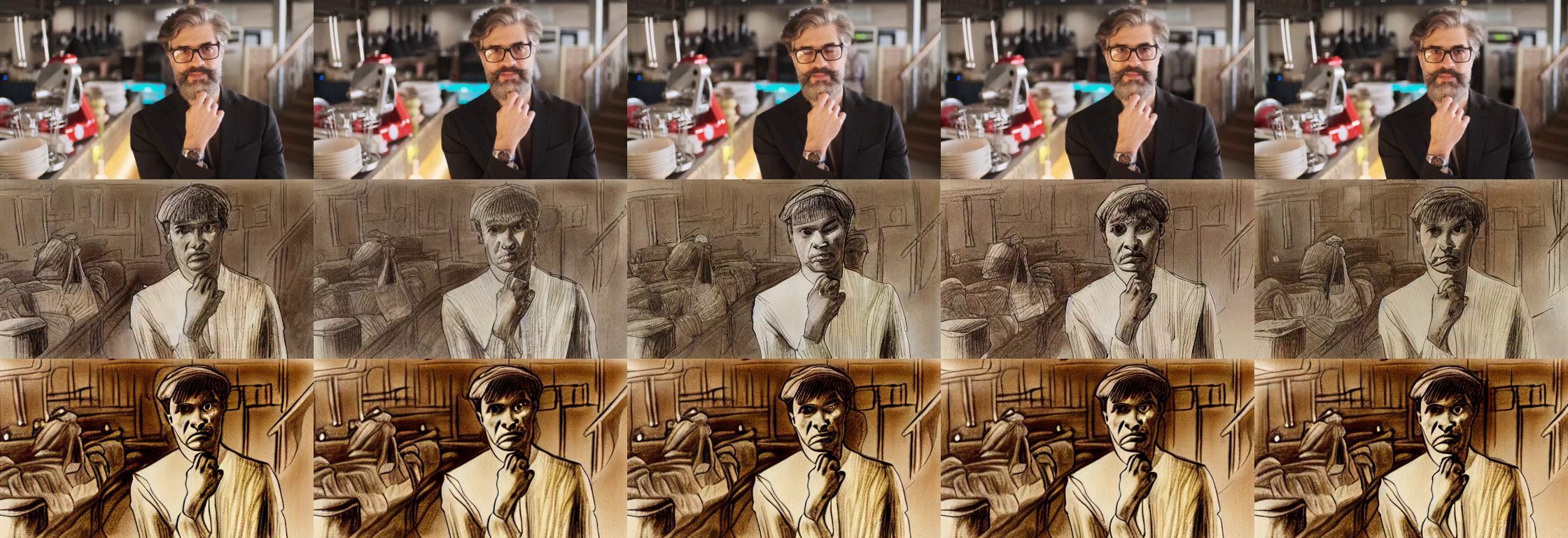}\vspace{1em}
    \caption{\textbf{Temporal Control:} By training image and video models jointly, we obtain explicit
control over the temporal consistency of edited videos via a temporal guidance
scale $\omega_t$. On the left, frame consistency measured via
CLIP cosine similarity of consecutive frames increases monotonically with
$\omega_t$, while mean squared error between frames warped with optical flow
decreases monotonically.
On the right, lower scales ($0.5$ in the middle row) achieve edits with a
"hand-drawn" look, whereas higher scales ($1.5$ in the bottom row) result in
smoother results. Top row shows the original input video, the two edits use
the prompt "pencil sketch of a man looking at the camera".}
\label{fig:tempcontrol}
\end{figure*}%
}

\newcommand{\figfidelitycontrol}{%
\begin{figure*}
    \includegraphics[width=\textwidth]{figures/fidelity_control.jpg}
    \caption{\textbf{Controlling Fidelity:} We obtain control over structure
    and appearance-fidelity. Each cell shows three frames produced with
    decreasing structure-fidelity $t_s$ (left-to-right) and increasing number
    of customization training steps in the rows (none, 250, 1500). The bottom
    row shows examples of reference images used for customization (red border)
    and the input image (blue border). Same driving video as in
    Fig.~\ref{fig:teaser}.}
    \label{fig:fidelitycontrol}
\end{figure*}%
}

\newcommand{\figfidelitycontrolsintel}{%
\begin{figure*}
    \includegraphics[width=\textwidth]{figures/fidelity_control_sintel_00.jpg}
    \caption{We obtain control over structure- and appearance-fidelity. Each
    cell shows three frames produced with decreasing structure-fidelity
    (left-to-right) and increasing number of customization training steps in
    the rows (none, 250, 2500).}
    \label{fig:fidelitycontrolsintel}
\end{figure*}%
}

\newcommand{\figfidelitycontrolsintelcolumn}{%
\begin{figure}
    \includegraphics[width=\columnwidth]{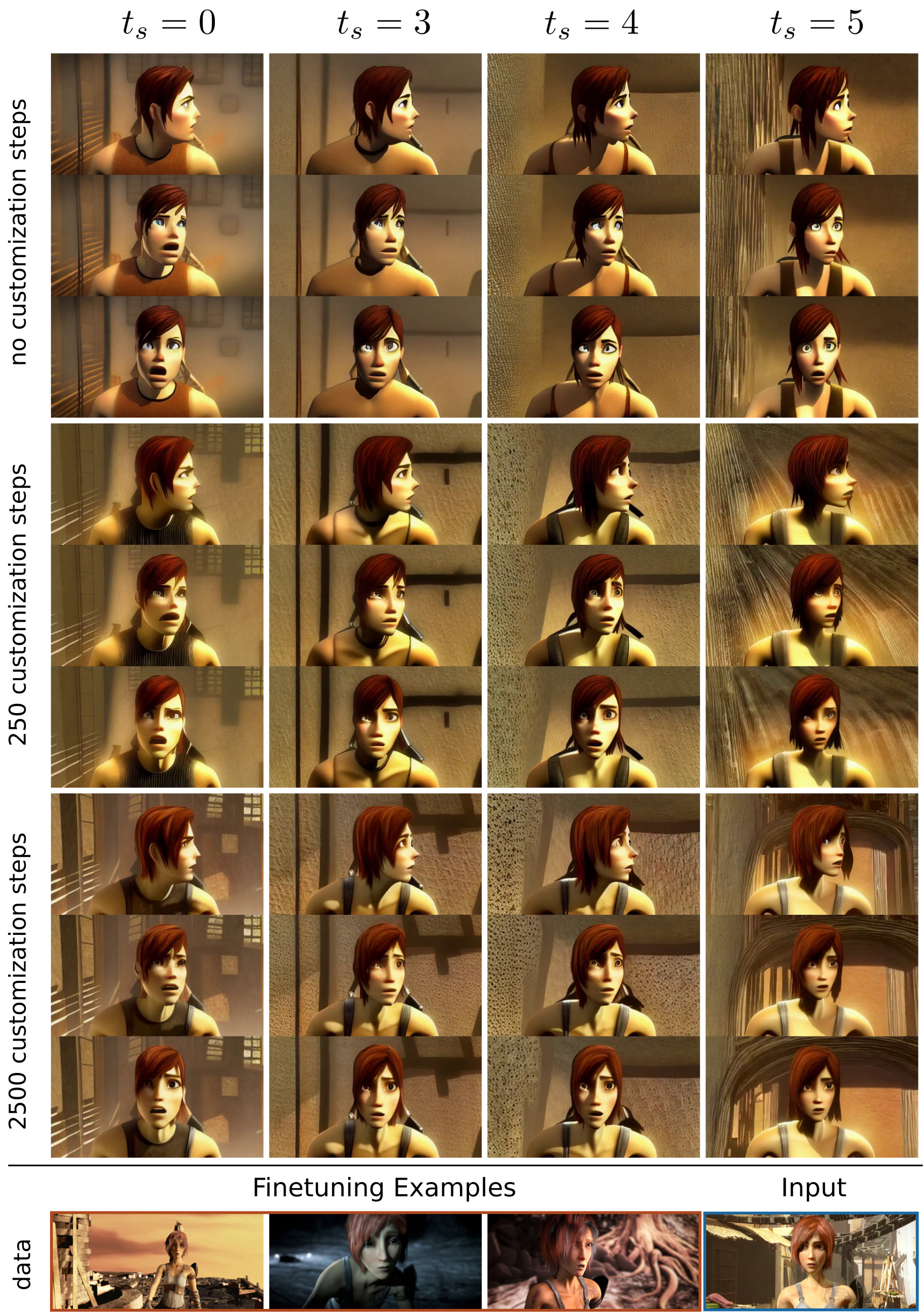}
    \caption{\textbf{Controlling Fidelity:} We obtain control over structure
    and appearance-fidelity. Each cell shows three frames produced with
    decreasing structure-fidelity $t_s$ (left-to-right) and increasing number
    of customization training steps (top-to-bottom). The bottom
    shows examples of images used for customization (red border)
    and the input image (blue border). Same driving video as in
    Fig.~\ref{fig:teaser}.}
    \label{fig:fidelitycontrolsintelcolumn}
\end{figure}%
}

\newcommand{\figfidelitycontroltheone}{%
\begin{figure*}
    \includegraphics[width=\textwidth]{figures/fidelity_control_theone_00.jpg}
    \caption{We obtain control over structure- and appearance-fidelity. Each
    cell shows three frames produced with decreasing structure-fidelity
    (left-to-right) and increasing number of customization training steps in
    the rows (none, 250, 1250).}
    \label{fig:fidelitycontroltheone}
\end{figure*}%
}

\newcommand{\figtexttovidedit}{%
\begin{figure*}
\centering
\renewcommand{\cellwidth}{0.1\textwidth}
\renewcommand{\imwidth}{0.8\textwidth}
\setcellgapes{0.3em}\makegapedcells
\begin{tabular}{cc}
  Prompt & Driving Video (top) and Result (bottom) \\ \hline
\parbox{\cellwidth}{a man using a laptop inside a train, anime style} &
\parbox{\imwidth}{\includegraphics[width=\imwidth]{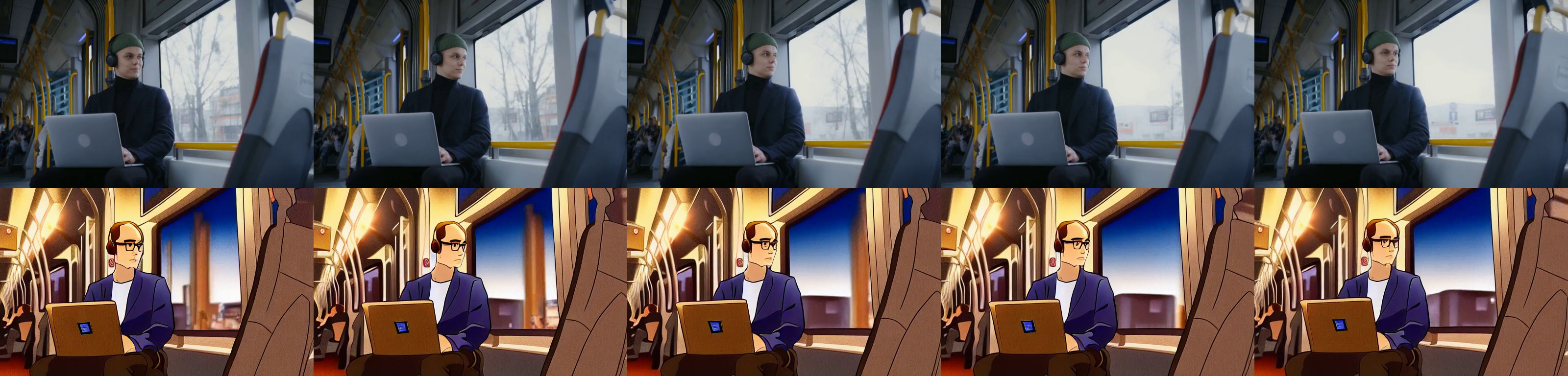}}
\\
\parbox{\cellwidth}{a woman and man take selfies while walking down the street, claymation} &
\parbox{\imwidth}{\includegraphics[width=\imwidth]{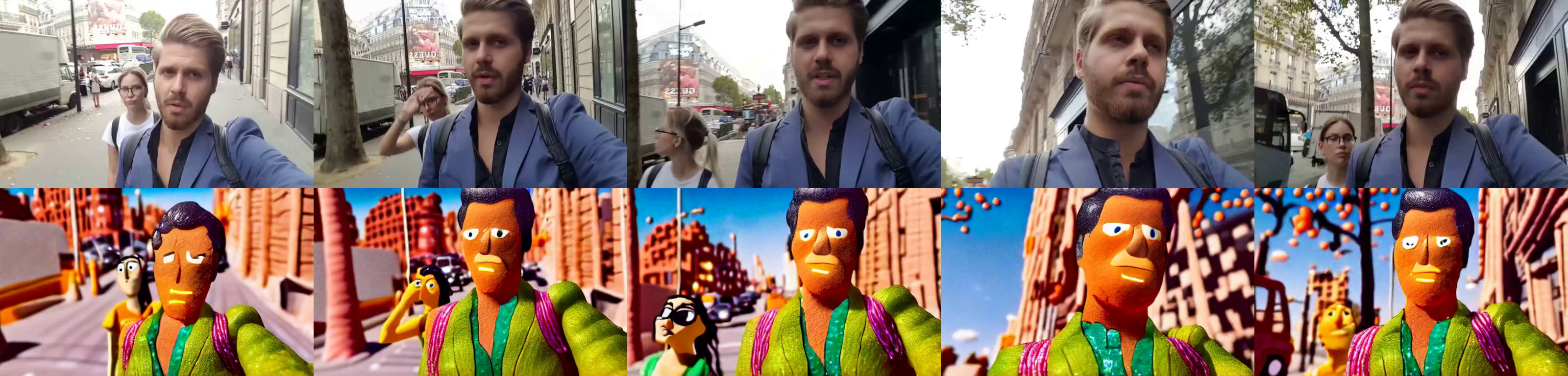}}
\\
\parbox{\cellwidth}{kite-surfer in the ocean at sunset} &
\parbox{\imwidth}{\includegraphics[width=\imwidth]{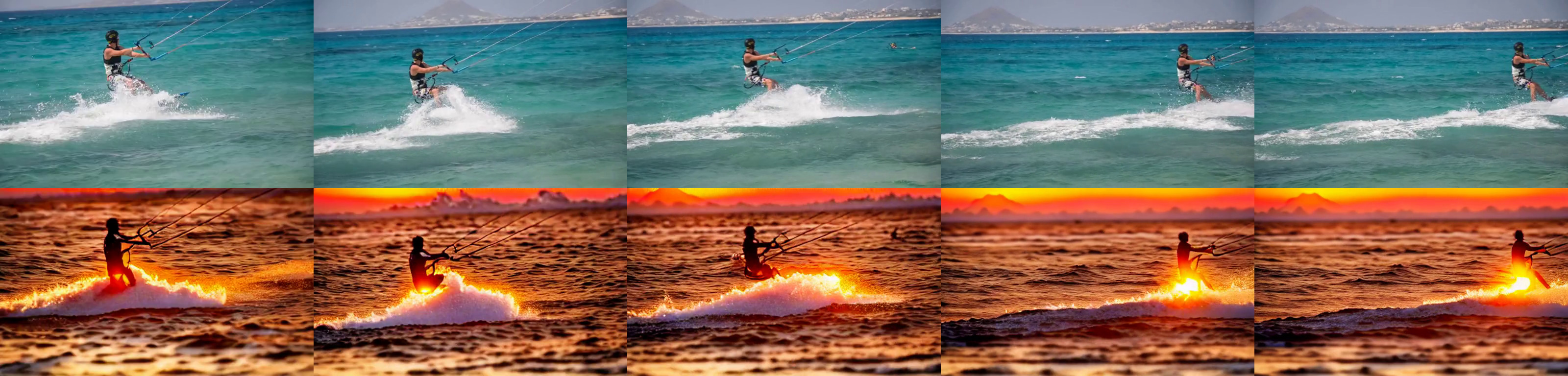}}
\\
\parbox{\cellwidth}{car on a snow-covered road in the countryside} &
\parbox{\imwidth}{\includegraphics[width=\imwidth]{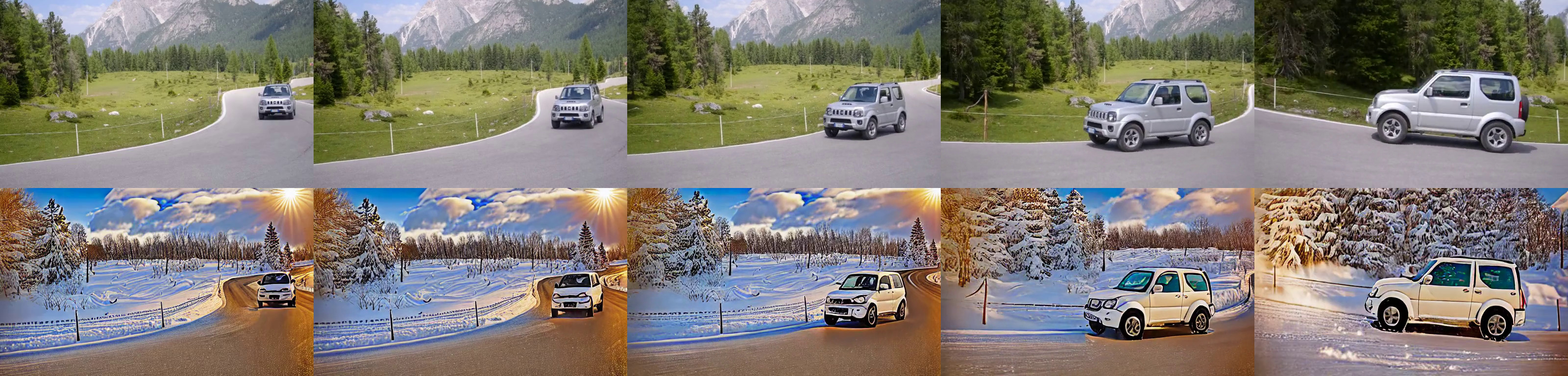}}
\\
\parbox{\cellwidth}{alien explorer hiking in the mountains} &
\parbox{\imwidth}{\includegraphics[width=\imwidth]{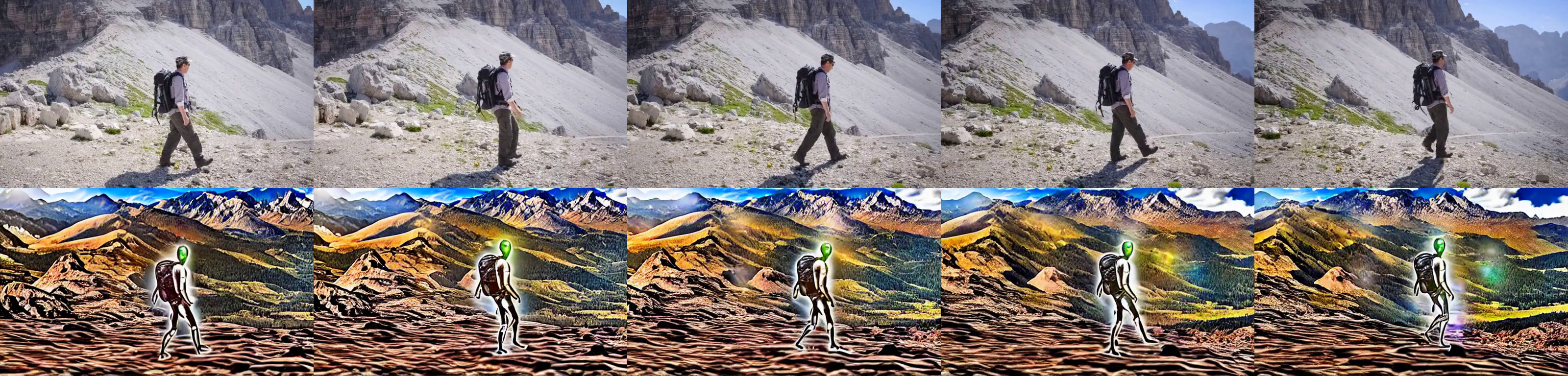}}
\\
\parbox{\cellwidth}{a space bear walking through the stars} &
\parbox{\imwidth}{\includegraphics[width=\imwidth]{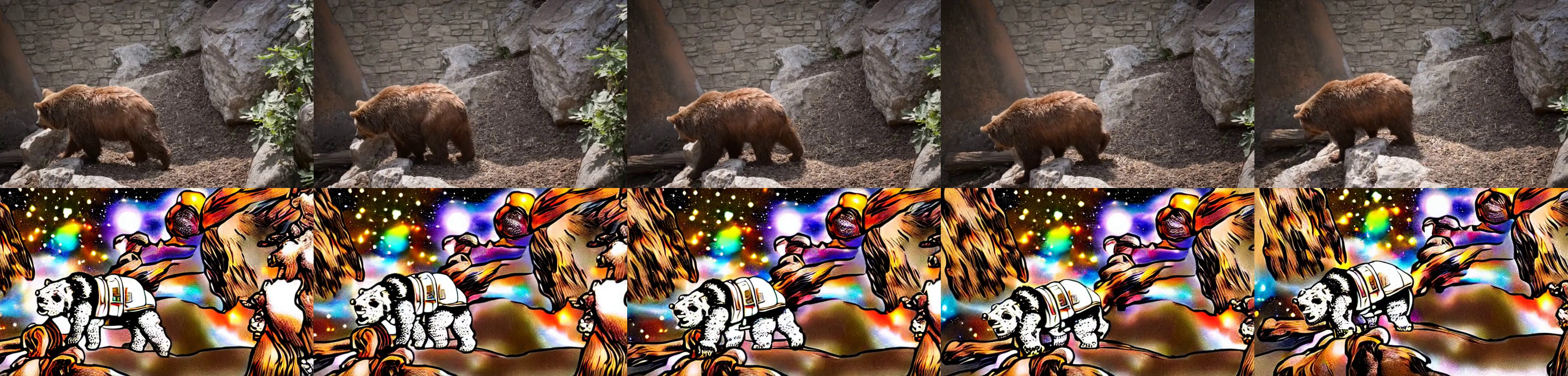}}
\vspace{1em}
\end{tabular}
  \caption{Our approach enables a wide range of video edits, including changes
to animation styles such as anime or claymation, changes of environment such
as day of time or season, and changing characters such as humans to aliens or
  move scenes from nature to outer space.}
  \label{fig:texttovidedit}
\end{figure*}%
}

\newcommand{\figtexttovideditA}{%
\begin{figure*}
  \centering

\renewcommand{\cellwidth}{0.1\textwidth}
\renewcommand{\imwidth}{0.9\textwidth}
\renewcommand{\arraystretch}{7}
\begin{tabular}{cc}
  Prompt & Driving Video (top) and Result (bottom) \vspace{-2em}\\ \hline
\parbox{\cellwidth}{pencil sketch of a man looking at the camera, black and white} &
\parbox{\imwidth}{\includegraphics[width=\imwidth]{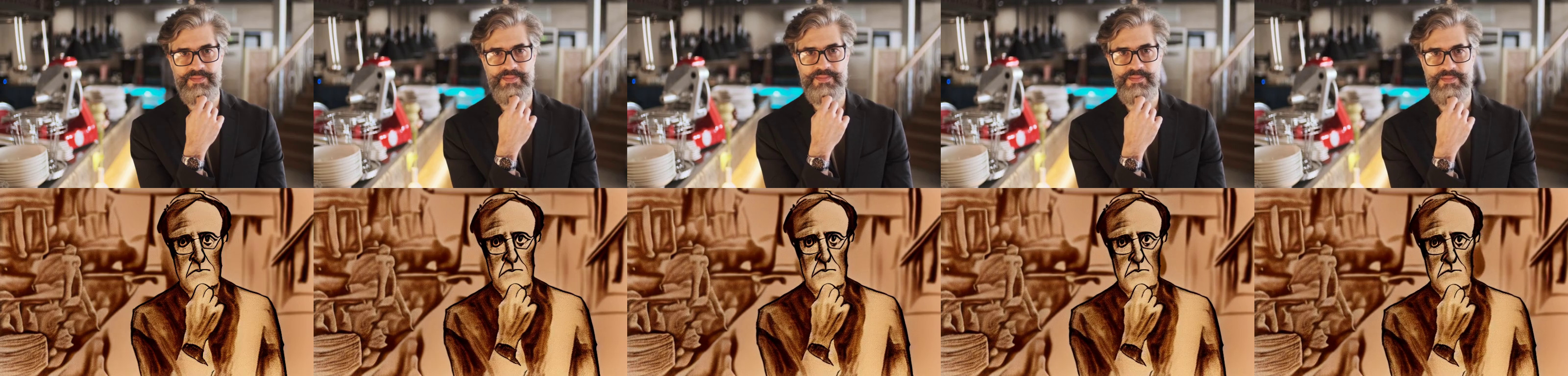}}
\\
\parbox{\cellwidth}{a man using a laptop inside a train, anime style} &
\parbox{\imwidth}{\includegraphics[width=\imwidth]{figures/texttovidedit/v00/00002.jpg}}
\\
\parbox{\cellwidth}{a woman and man take selfies while walking down the street, claymation} &
\parbox{\imwidth}{\includegraphics[width=\imwidth]{figures/texttovidedit/v00/00003.jpg}}
\\
\parbox{\cellwidth}{oil painting of a man driving} &
\parbox{\imwidth}{\includegraphics[width=\imwidth]{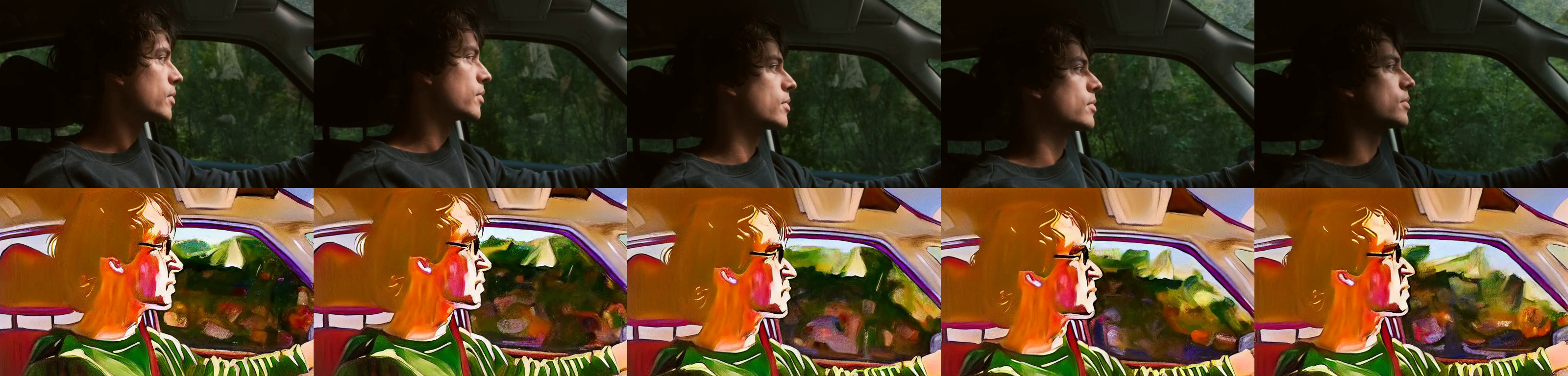}}
\\
\parbox{\cellwidth}{low-poly render of a man texting on the street} &
\parbox{\imwidth}{\includegraphics[width=\imwidth]{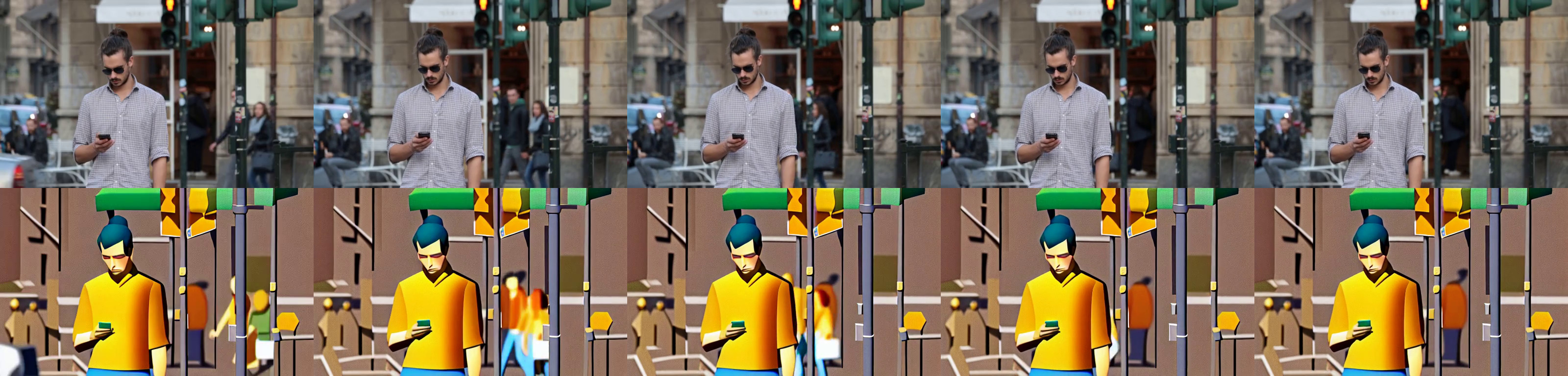}}

\vspace{1em}
\end{tabular}

  \caption{Additional results for text-to-video-editing.}
  \label{fig:texttovideditA}
\end{figure*}%
}

\newcommand{\figtexttovideditB}{%
\begin{figure*}
  \centering

\renewcommand{\cellwidth}{0.1\textwidth}
\renewcommand{\imwidth}{0.9\textwidth}
\renewcommand{\arraystretch}{7}
\begin{tabular}{cc}
  Prompt & Driving Video (top) and Result (bottom) \vspace{-2em}\\ \hline
\parbox{\cellwidth}{2D vector animation of a group of flamingos standing near some rocks and water} &
\parbox{\imwidth}{\includegraphics[width=\imwidth]{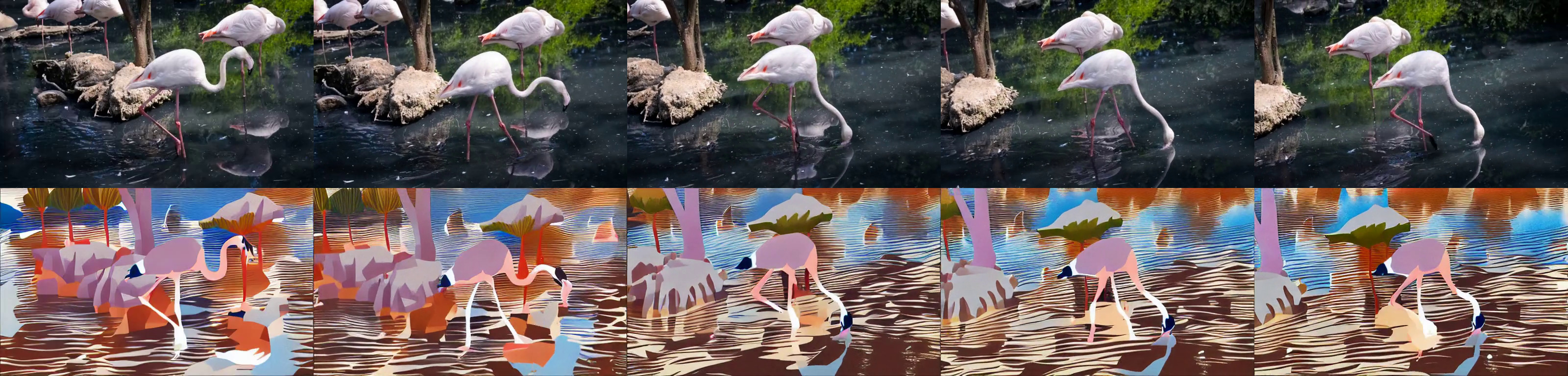}}
\\
\parbox{\cellwidth}{cartoon animation of an elephant walks through dirt surrounded by boulders} &
\parbox{\imwidth}{\includegraphics[width=\imwidth]{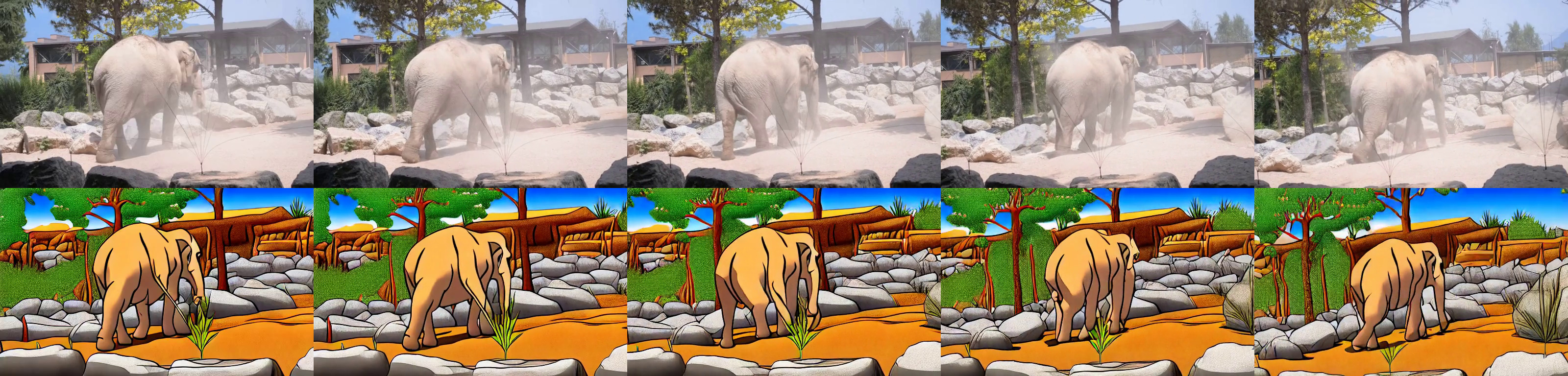}}
\\
\parbox{\cellwidth}{cyberpunk neon car on a road in the countryside} &
\parbox{\imwidth}{\includegraphics[width=\imwidth]{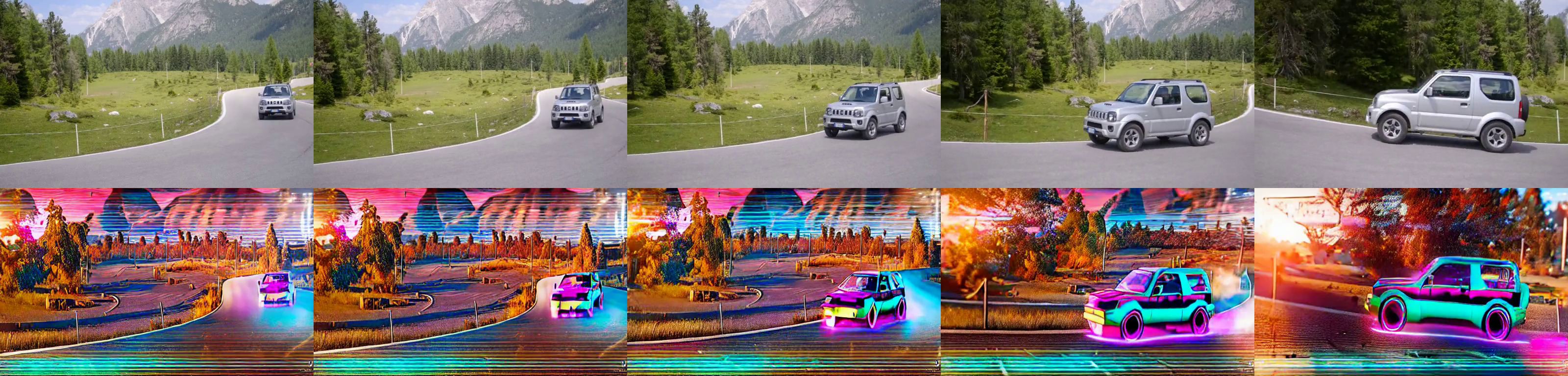}}
\\
\parbox{\cellwidth}{a crochet black swan swims in a pond with rocks and vegetation} &
\parbox{\imwidth}{\includegraphics[width=\imwidth]{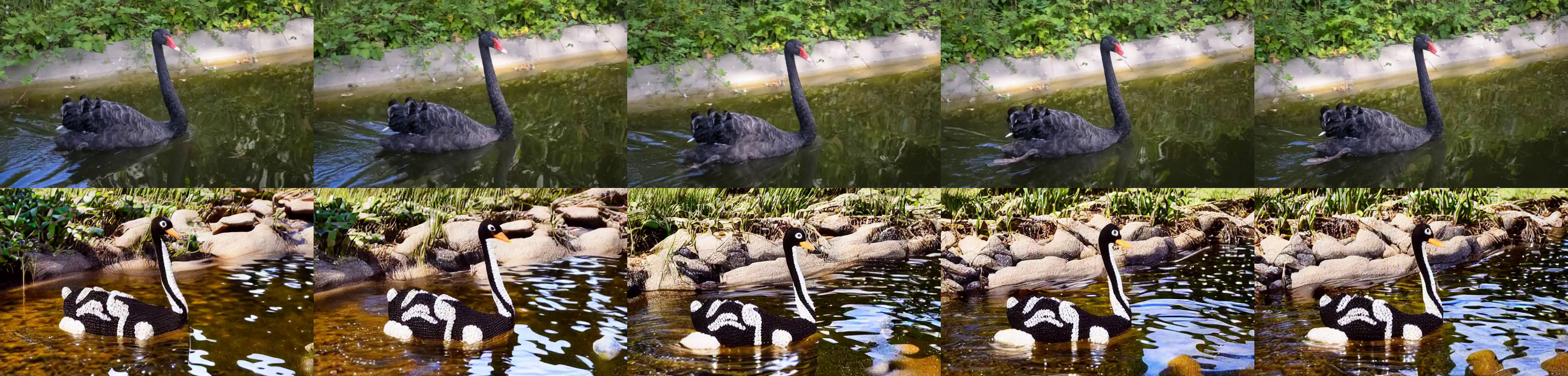}}
\\
\parbox{\cellwidth}{a dalmatian dog is walking away from a fence} &
\parbox{\imwidth}{\includegraphics[width=\imwidth]{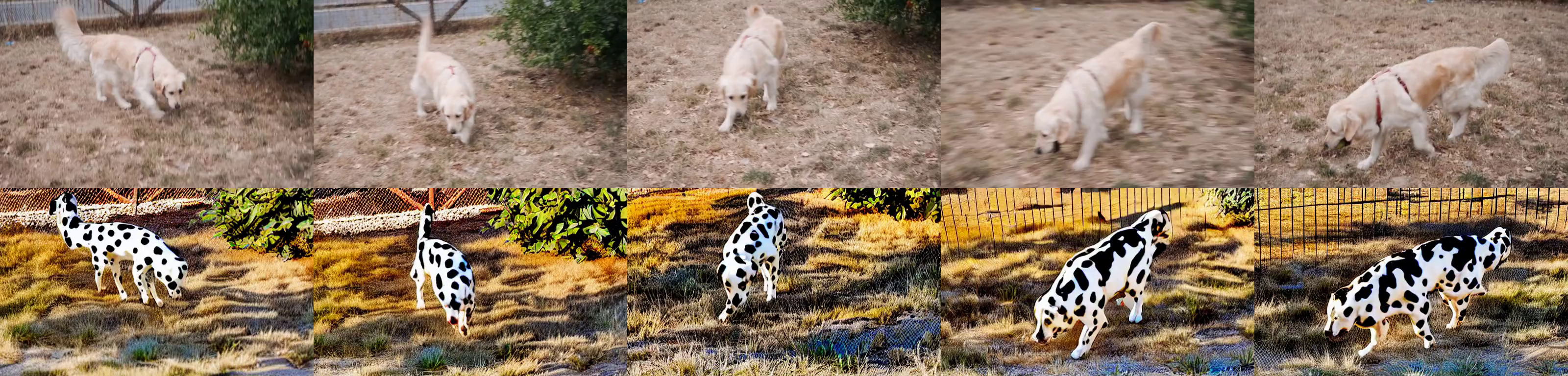}}

\vspace{1em}
\end{tabular}
  \caption{Additional results for text-to-video-editing.}
  \label{fig:texttovideditB}
\end{figure*}%
}

\newcommand{\figtexttovideditC}{%
\begin{figure*}
  \centering

\renewcommand{\cellwidth}{0.1\textwidth}
\renewcommand{\imwidth}{0.9\textwidth}
\renewcommand{\arraystretch}{7}
\begin{tabular}{cc}
  Prompt & Driving Video (top) and Result (bottom) \vspace{-2em}\\ \hline
\parbox{\cellwidth}{kite-surfer in the ocean at sunset} &
\parbox{\imwidth}{\includegraphics[width=\imwidth]{figures/texttovidedit/v00/00011.jpg}}
\\
\parbox{\cellwidth}{car on a snow-covered road in the countryside} &
\parbox{\imwidth}{\includegraphics[width=\imwidth]{figures/texttovidedit/v00/00012.jpg}}
\\
\parbox{\cellwidth}{small grey suv driving in front of apartment buildings at night} &
\parbox{\imwidth}{\includegraphics[width=\imwidth]{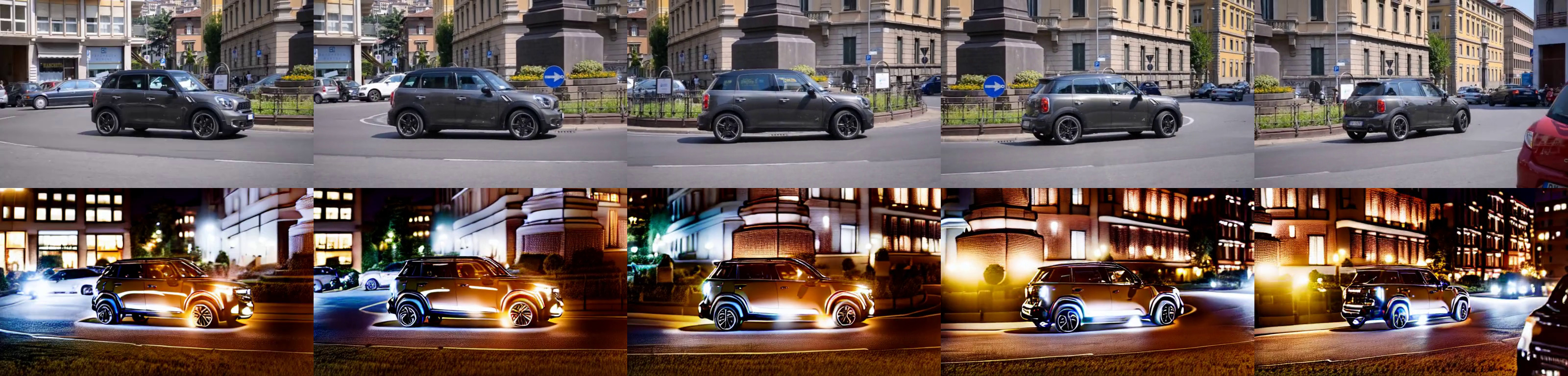}}
\\
\parbox{\cellwidth}{a space bear walking through the stars} &
\parbox{\imwidth}{\includegraphics[width=\imwidth]{figures/texttovidedit/v00/00014.jpg}}
\\
\parbox{\cellwidth}{white swan swimming in the water} &
\parbox{\imwidth}{\includegraphics[width=\imwidth]{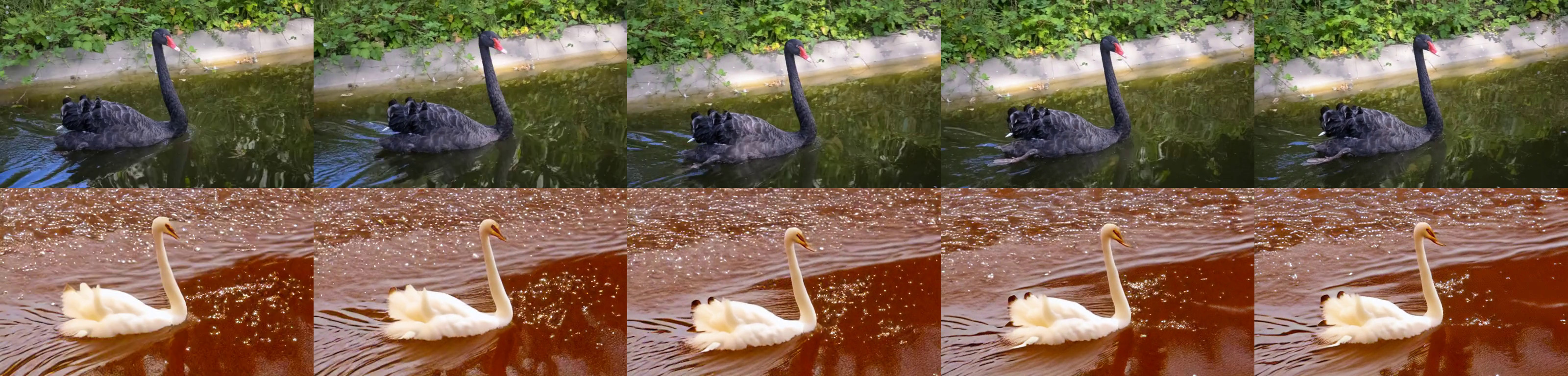}}

\vspace{1em}
\end{tabular}
    
  \caption{Additional results for text-to-video-editing.}
  \label{fig:texttovideditC}
\end{figure*}%
}

\newcommand{\figtexttovideditD}{%
\begin{figure*}
  \centering

\renewcommand{\cellwidth}{0.1\textwidth}
\renewcommand{\imwidth}{0.9\textwidth}
\renewcommand{\arraystretch}{7}
\begin{tabular}{cc}
  Prompt & Driving Video (top) and Result (bottom) \vspace{-2em}\\ \hline
\parbox{\cellwidth}{man riding a bicycle up the side of a dirt slope in a graphic novel style} &
\parbox{\imwidth}{\includegraphics[width=\imwidth]{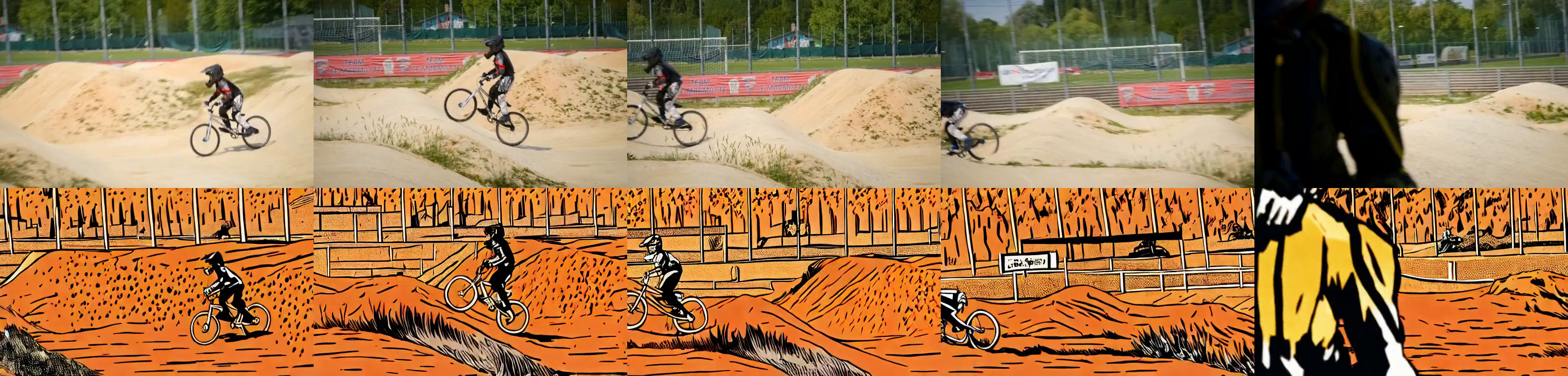}}
\\
\parbox{\cellwidth}{blue and white bus driving down a city street with a backdrop of snow-capped mountains} &
\parbox{\imwidth}{\includegraphics[width=\imwidth]{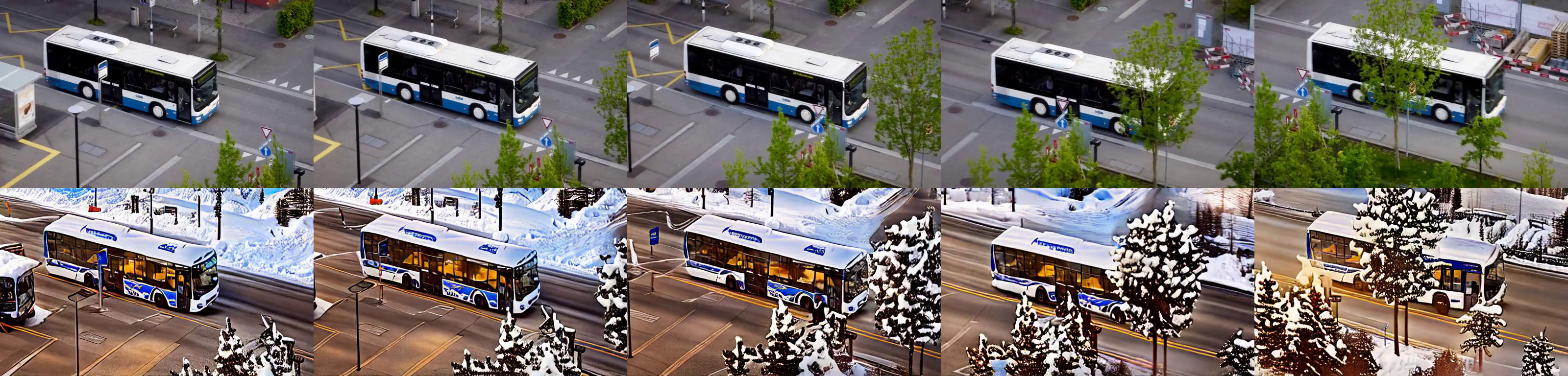}}
\\
\parbox{\cellwidth}{toy camel standing on dirt near a fence} &
\parbox{\imwidth}{\includegraphics[width=\imwidth]{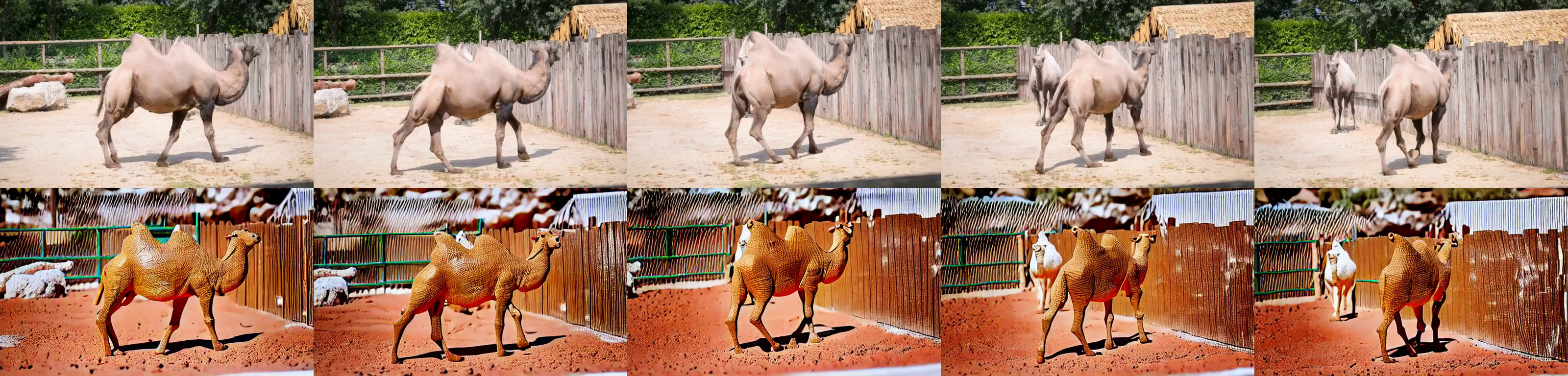}}
\\
\parbox{\cellwidth}{8-bit pixelated car driving down the road} &
\parbox{\imwidth}{\includegraphics[width=\imwidth]{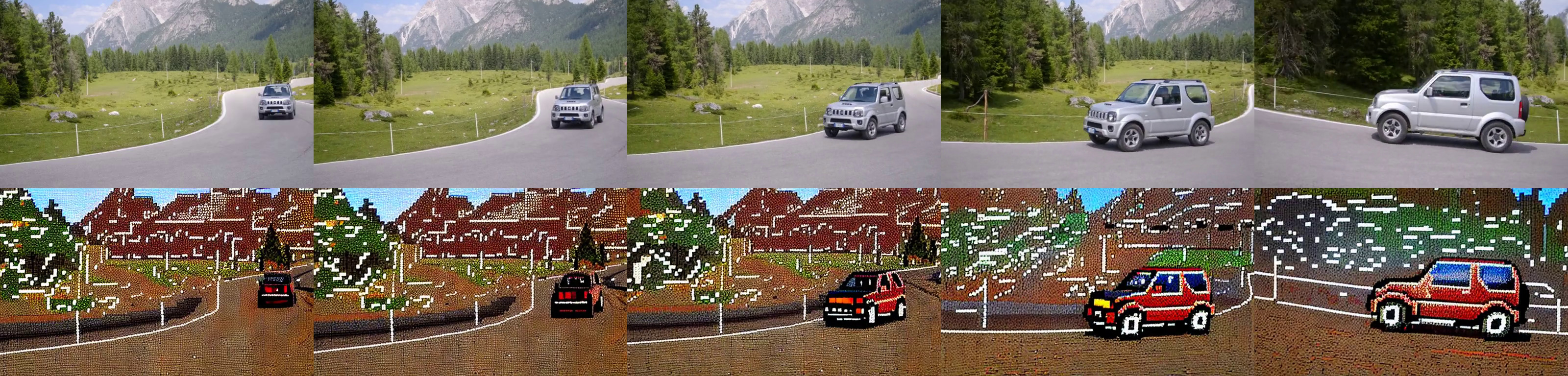}}
\\
\parbox{\cellwidth}{a robotic cow walking along a muddy road} &
\parbox{\imwidth}{\includegraphics[width=\imwidth]{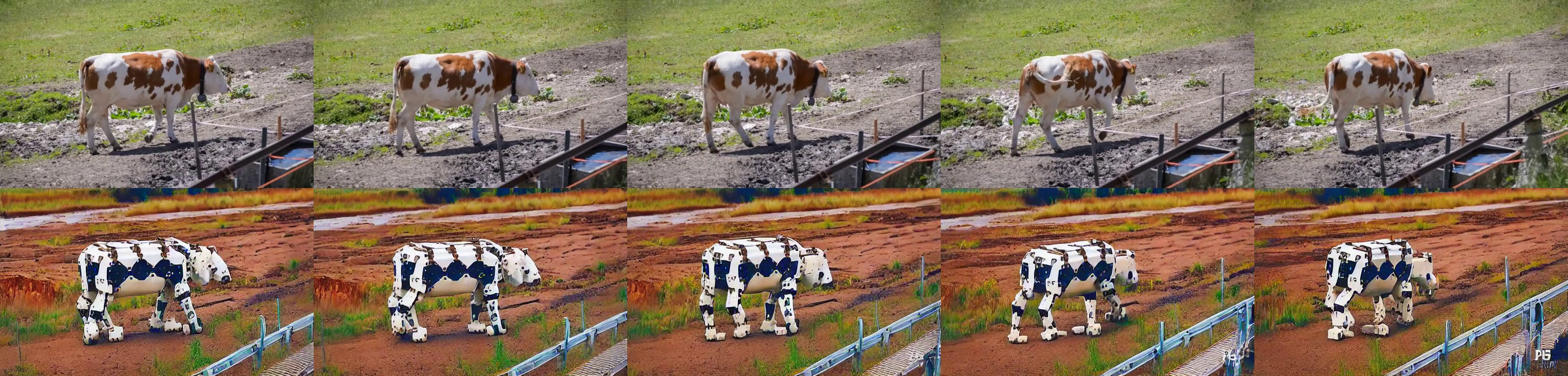}}

\vspace{1em}
\end{tabular}
    
  \caption{Additional results for text-to-video-editing.}
  \label{fig:texttovideditD}
\end{figure*}%
}

\newcommand{\figtexttovideditE}{%
\begin{figure*}
  \centering

\renewcommand{\cellwidth}{0.1\textwidth}
\renewcommand{\imwidth}{0.9\textwidth}
\renewcommand{\arraystretch}{7}
\begin{tabular}{cc}
  Prompt & Driving Video (top) and Result (bottom) \vspace{-2em}\\ \hline
\parbox{\cellwidth}{oil painting of four pink flamingos wading in water} &
\parbox{\imwidth}{\includegraphics[width=\imwidth]{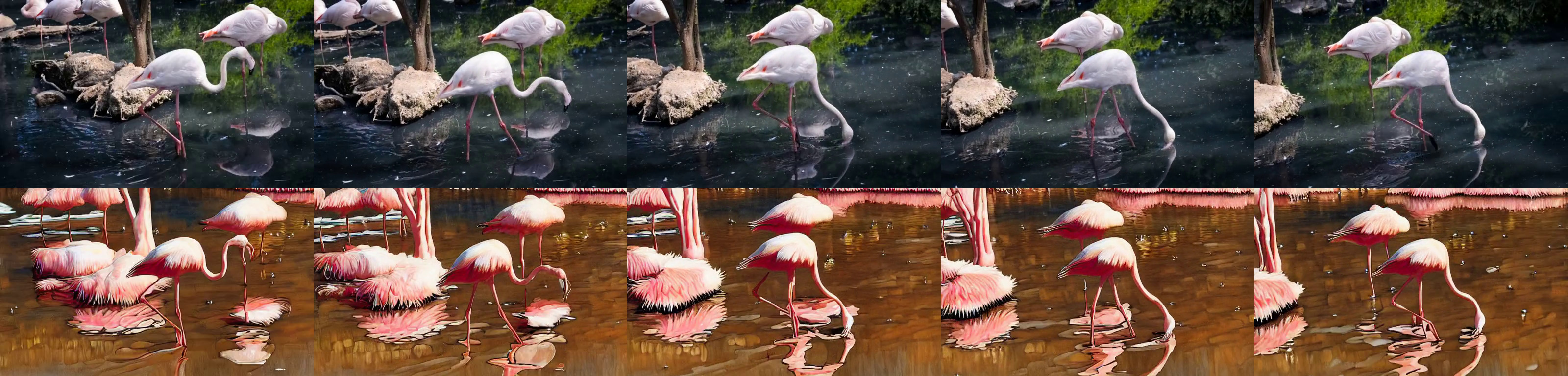}}
\\
\parbox{\cellwidth}{paper cut-out mountains with a hiker} &
\parbox{\imwidth}{\includegraphics[width=\imwidth]{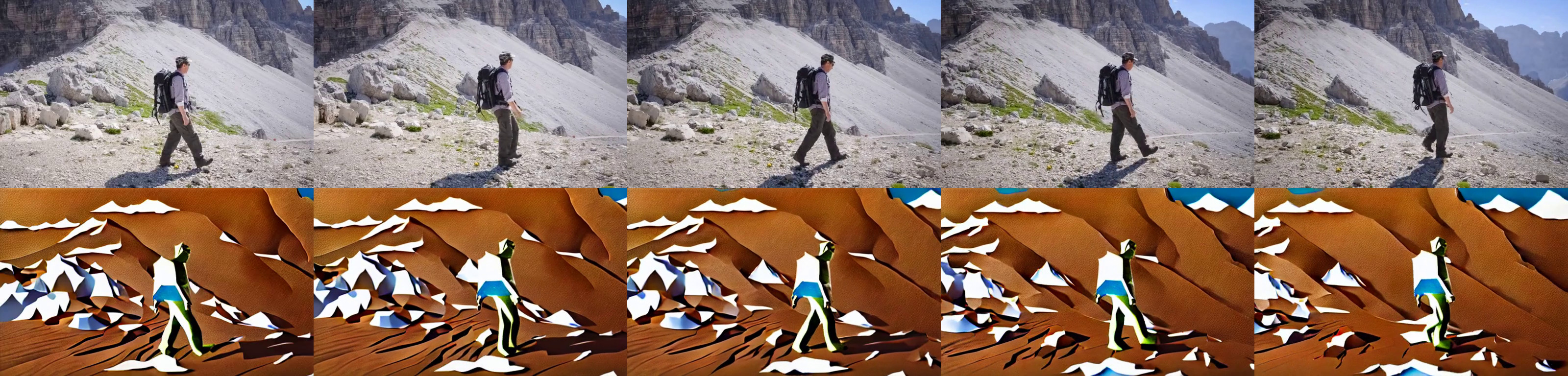}}
\\
\parbox{\cellwidth}{alien explorer hiking in the mountains} &
\parbox{\imwidth}{\includegraphics[width=\imwidth]{figures/texttovidedit/v00/00023.jpg}}
\\
\parbox{\cellwidth}{man hiking in the starry mountains} &
\parbox{\imwidth}{\includegraphics[width=\imwidth]{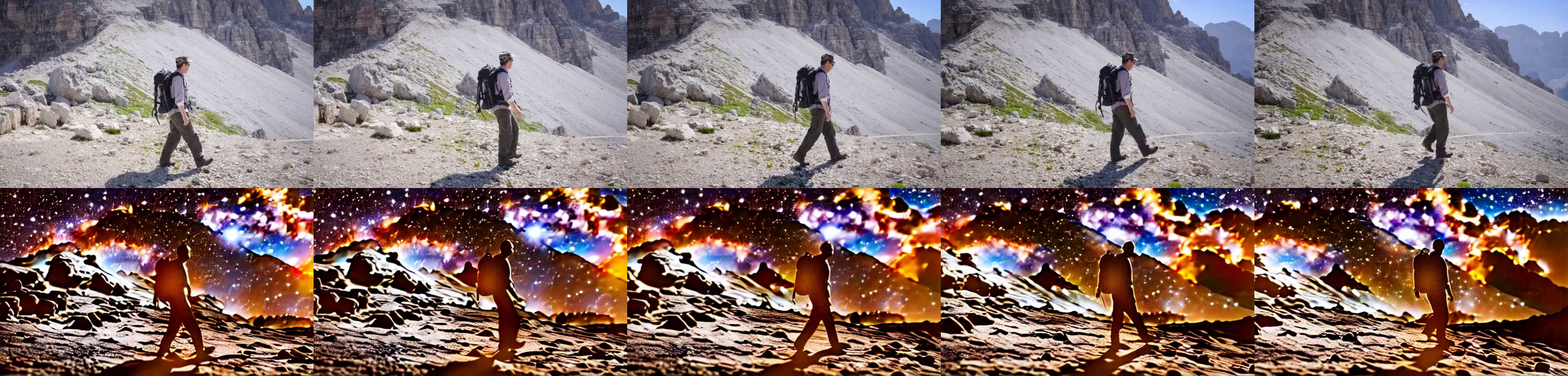}}
\\
\parbox{\cellwidth}{magical flying horse jumping over an obstacle} &
\parbox{\imwidth}{\includegraphics[width=\imwidth]{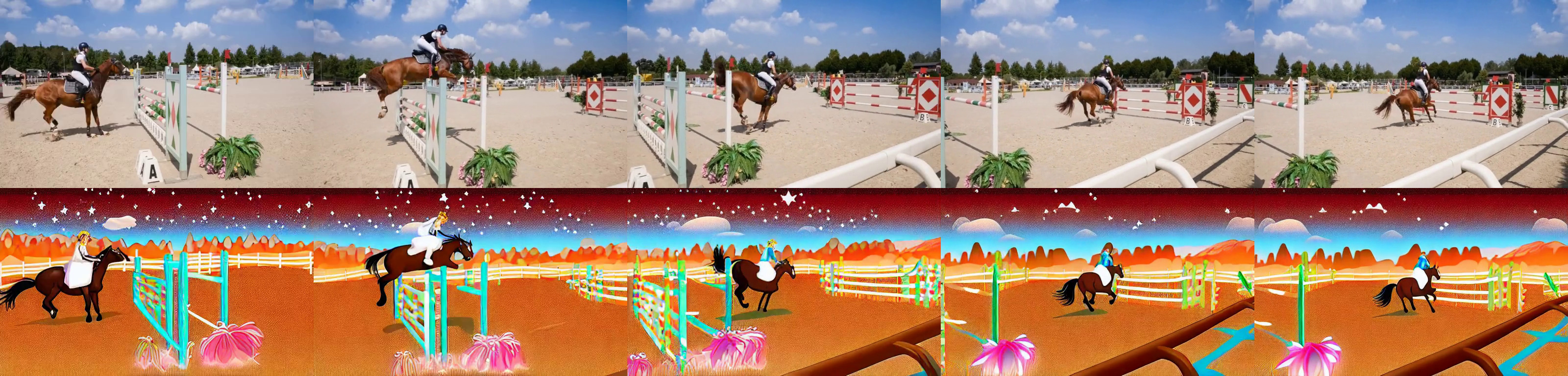}}

\vspace{1em}
\end{tabular}
    
  \caption{Additional results for text-to-video-editing.}
  \label{fig:texttovideditE}
\end{figure*}%
}

\newcommand{\figtexttovideditF}{%
\begin{figure*}
  \centering

\renewcommand{\cellwidth}{0.1\textwidth}
\renewcommand{\imwidth}{0.9\textwidth}
\renewcommand{\arraystretch}{7}
\begin{tabular}{cc}
  Prompt & Driving Video (top) and Result (bottom) \vspace{-2em}\\ \hline
\parbox{\cellwidth}{person rides on a horse while jumping over an obstacle with an aurora borealis in the background.} &
\parbox{\imwidth}{\includegraphics[width=\imwidth]{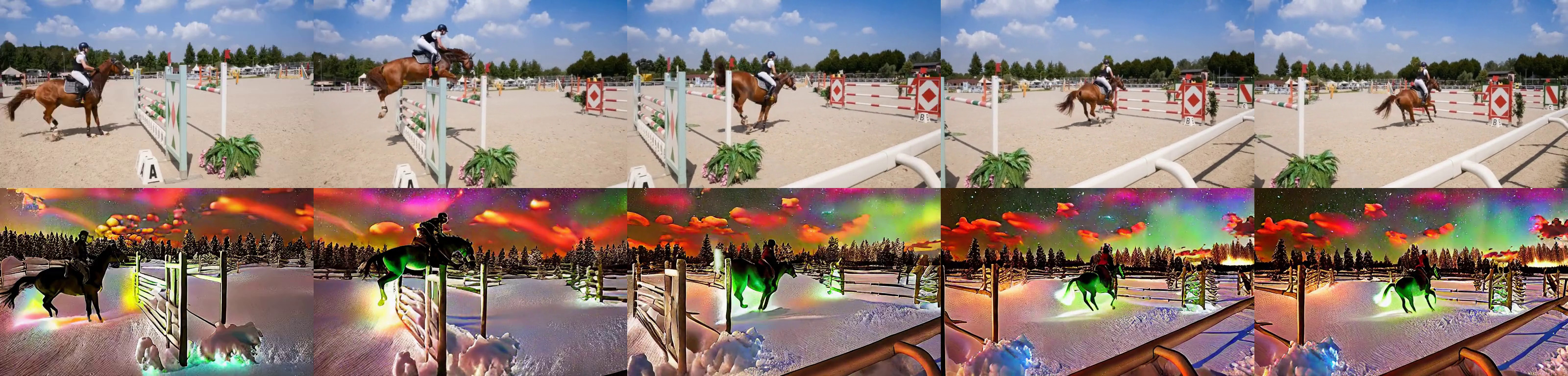}}
\\
\parbox{\cellwidth}{martial artists practicing on grassy mats while others watch} &
\parbox{\imwidth}{\includegraphics[width=\imwidth]{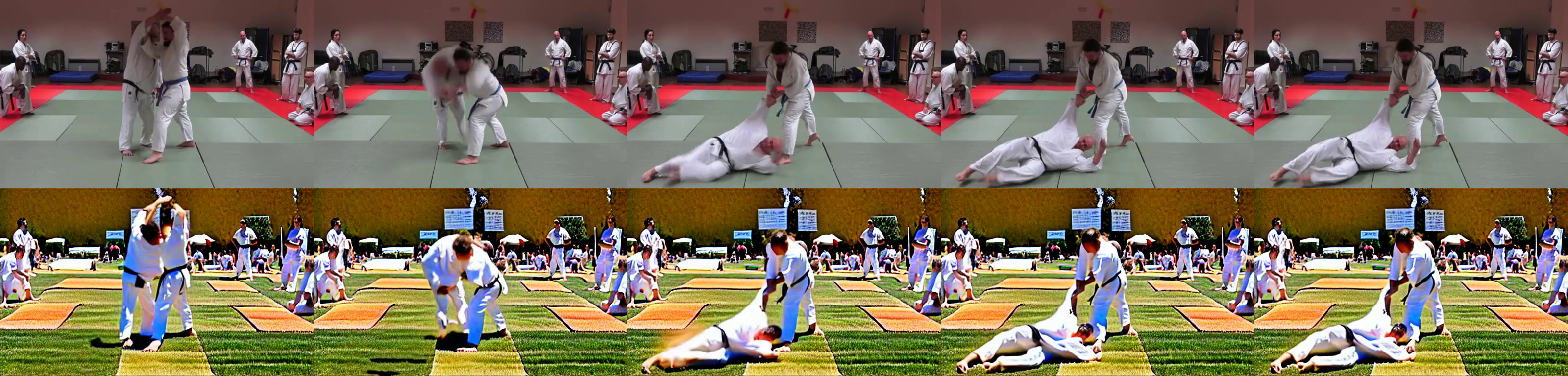}}
\\
\parbox{\cellwidth}{silhouetted martial artists practicing while others watch} &
\parbox{\imwidth}{\includegraphics[width=\imwidth]{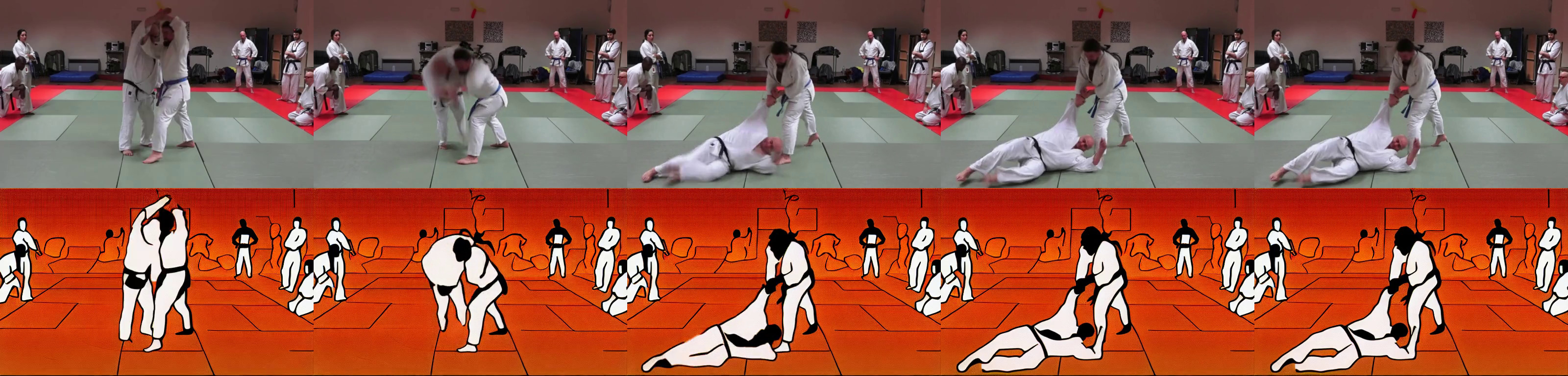}}
\\
\parbox{\cellwidth}{3D animation of a small dog running through grass} &
\parbox{\imwidth}{\includegraphics[width=\imwidth]{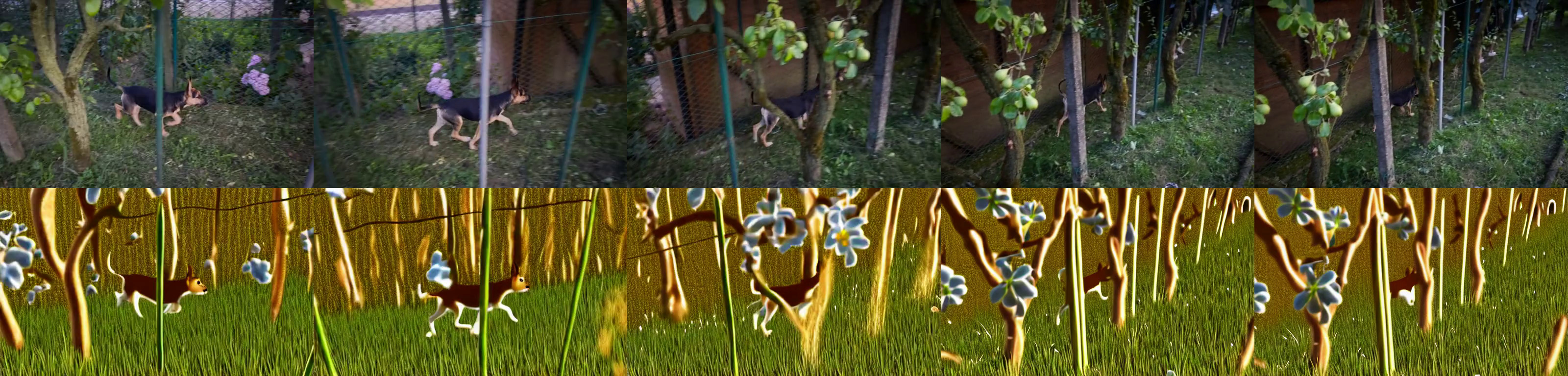}}
\\
\parbox{\cellwidth}{hyper-realistic painting of a person paragliding on a mountain} &
\parbox{\imwidth}{\includegraphics[width=\imwidth]{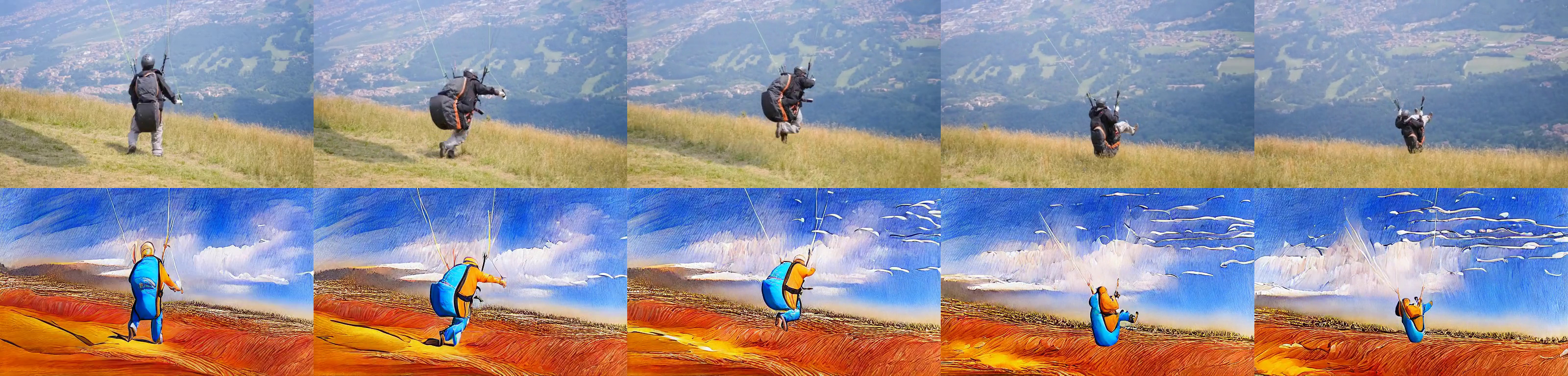}}

\vspace{1em}
\end{tabular}
    
  \caption{Additional results for text-to-video-editing.}
  \label{fig:texttovideditF}
\end{figure*}%
}

\newcommand{\figtexttovideditG}{%
\begin{figure*}
  \centering

\renewcommand{\cellwidth}{0.1\textwidth}
\renewcommand{\imwidth}{0.9\textwidth}
\renewcommand{\arraystretch}{7}
\begin{tabular}{cc}
  Prompt & Driving Video (top) and Result (bottom) \vspace{-2em}\\ \hline
\parbox{\cellwidth}{paraglider soaring on a mountain under a starry sky} &
\parbox{\imwidth}{\includegraphics[width=\imwidth]{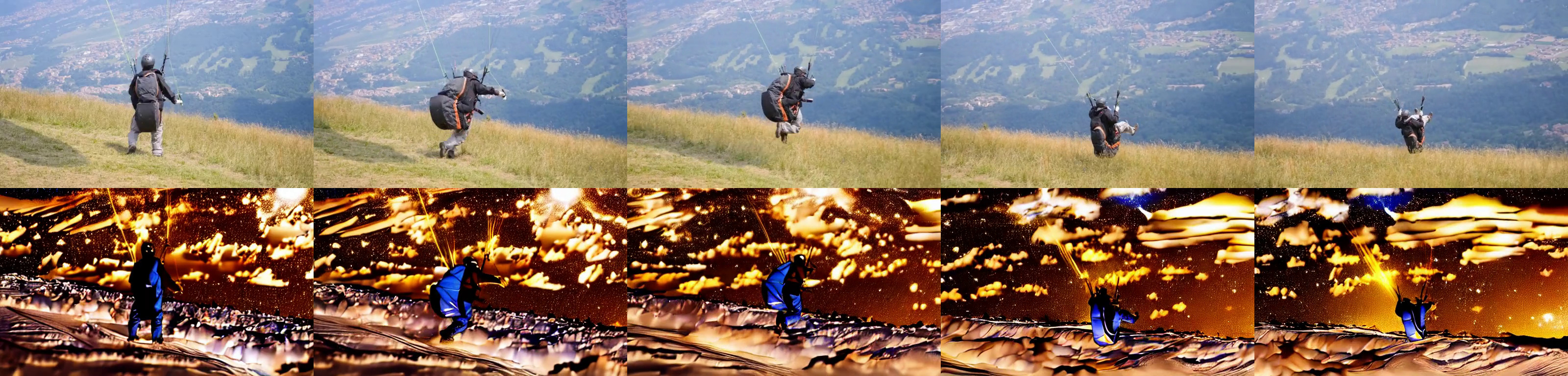}}
\\
\parbox{\cellwidth}{cartoon-style animation of a man riding a skateboard down a road} &
\parbox{\imwidth}{\includegraphics[width=\imwidth]{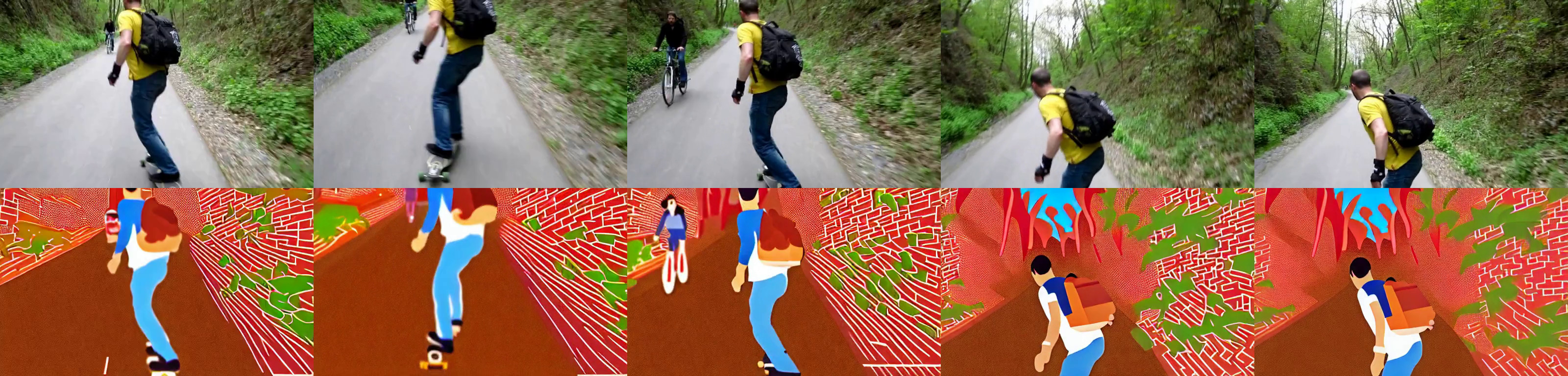}}
\\
\parbox{\cellwidth}{robot skateboarder riding down a road} &
\parbox{\imwidth}{\includegraphics[width=\imwidth]{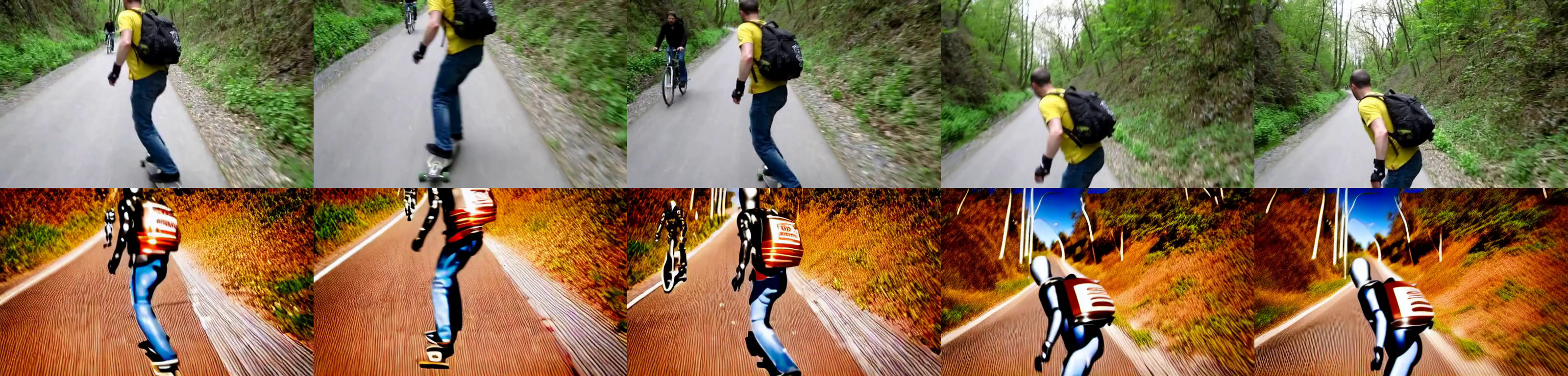}}
\\
\parbox{\cellwidth}{a man riding a skateboard down a magical river} &
\parbox{\imwidth}{\includegraphics[width=\imwidth]{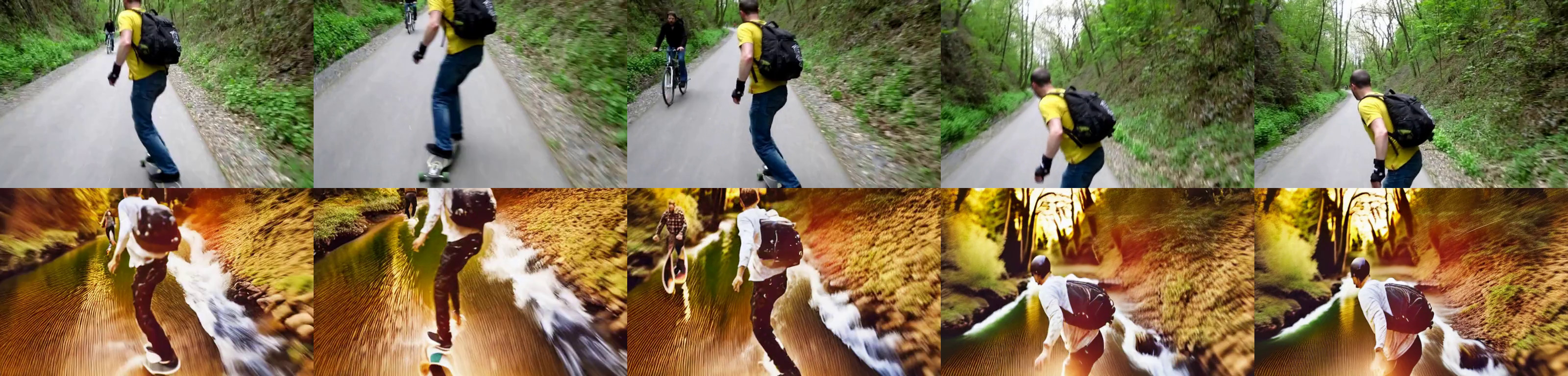}}
\\
\parbox{\cellwidth}{man playing tennis on the surface of the moon} &
\parbox{\imwidth}{\includegraphics[width=\imwidth]{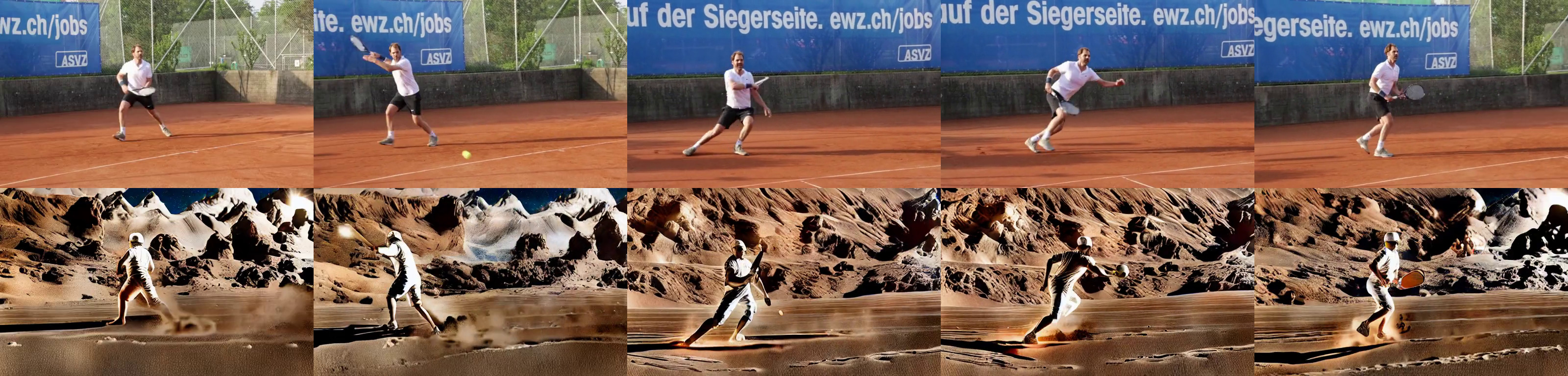}}

\vspace{1em}
\end{tabular}
    
  \caption{Additional results for text-to-video-editing.}
  \label{fig:texttovideditG}
\end{figure*}%
}

\newcommand{\figswapmatrix}{%
\begin{figure}
    \includegraphics[width=\columnwidth]{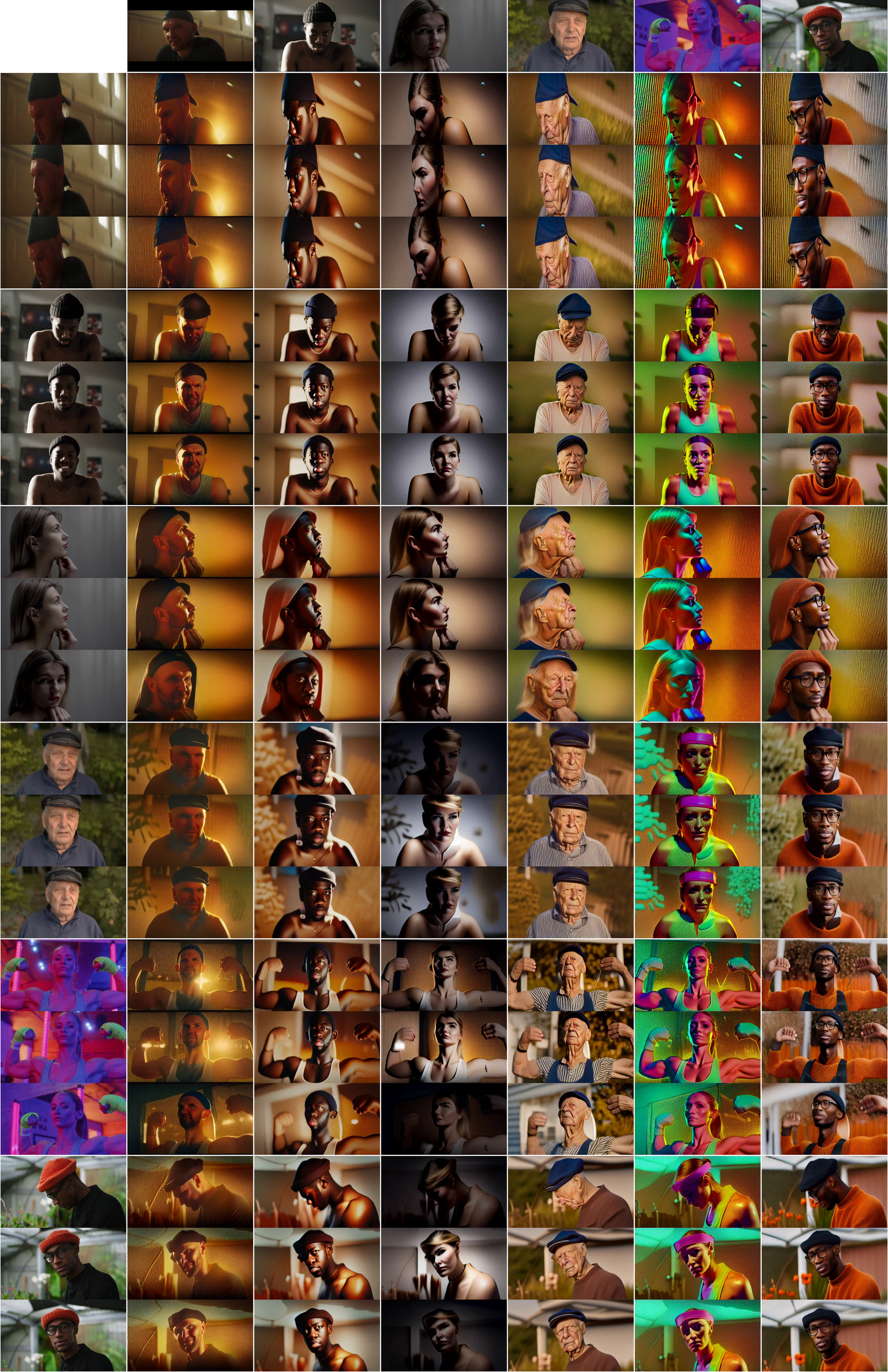}
    \caption{\textbf{Image Prompting:} We combine the structure of a driving video (first
    column) with content from other videos (first row).}
    \label{fig:swapmatrix}
\end{figure}%
}

\newcommand{\figimtovideditA}{%
\begin{figure*}
  \centering
\renewcommand{\cellwidth}{0.1\textwidth}
\renewcommand{\imwidth}{0.9\textwidth}
\renewcommand{\imwidthB}{0.1\textwidth}
\renewcommand{\arraystretch}{7}
\begin{tabular}{cc}
  Prompt & Driving Video (top) and Result (bottom) \vspace{-2em}\\ \hline
\parbox{\imwidthB}{\includegraphics[width=\imwidthB]{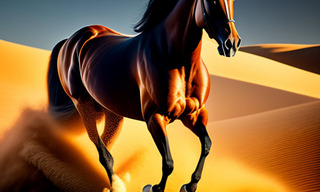}} & 
\parbox{\imwidth}{\includegraphics[width=\imwidth]{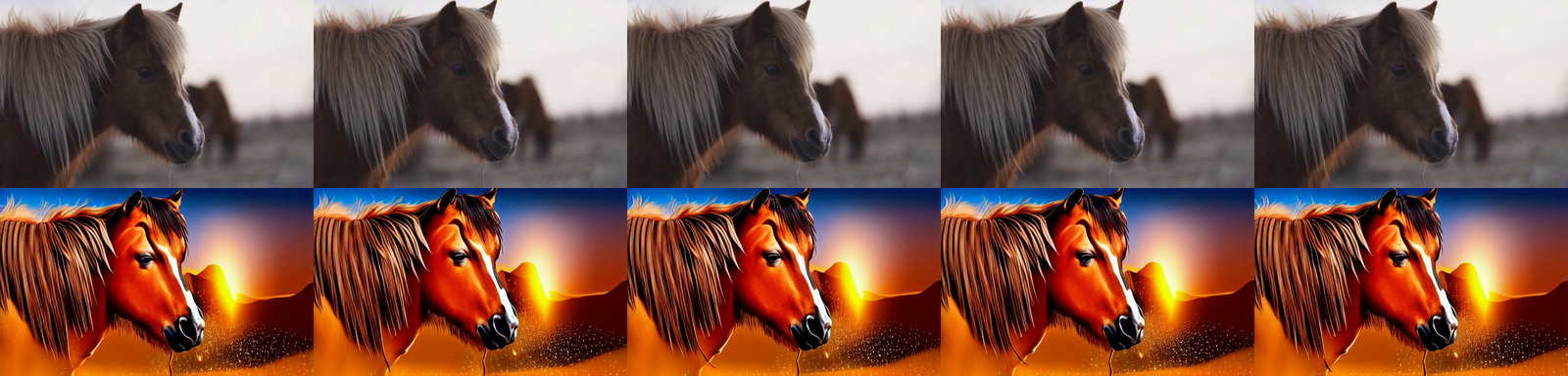}}
\\
\parbox{\imwidthB}{\includegraphics[width=\imwidthB]{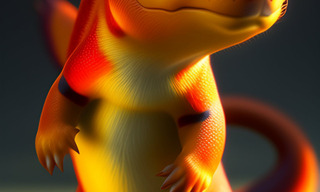}} & 
\parbox{\imwidth}{\includegraphics[width=\imwidth]{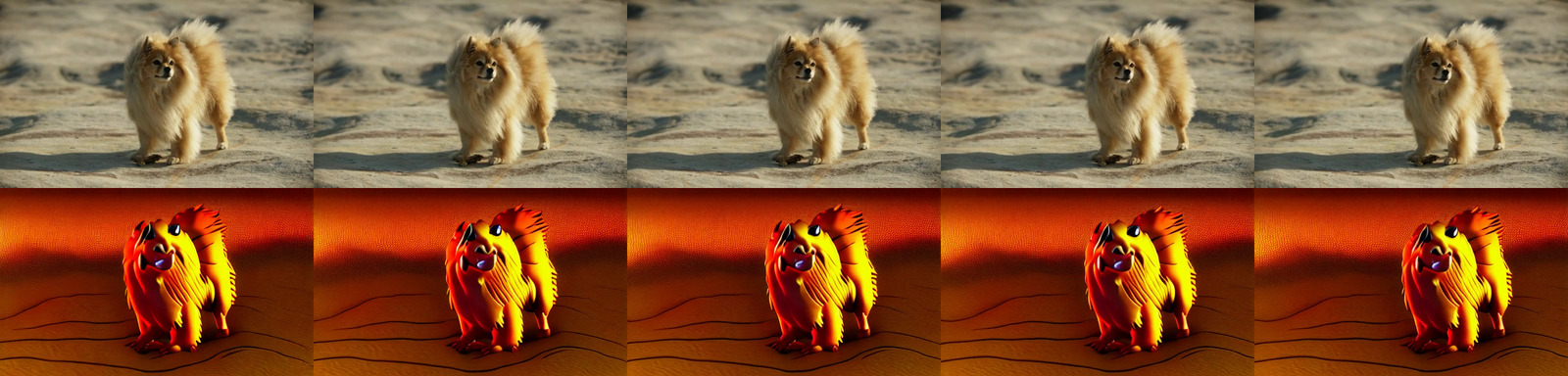}}
\\
\parbox{\imwidthB}{\includegraphics[width=\imwidthB]{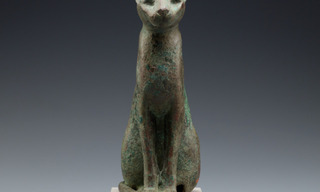}} & 
\parbox{\imwidth}{\includegraphics[width=\imwidth]{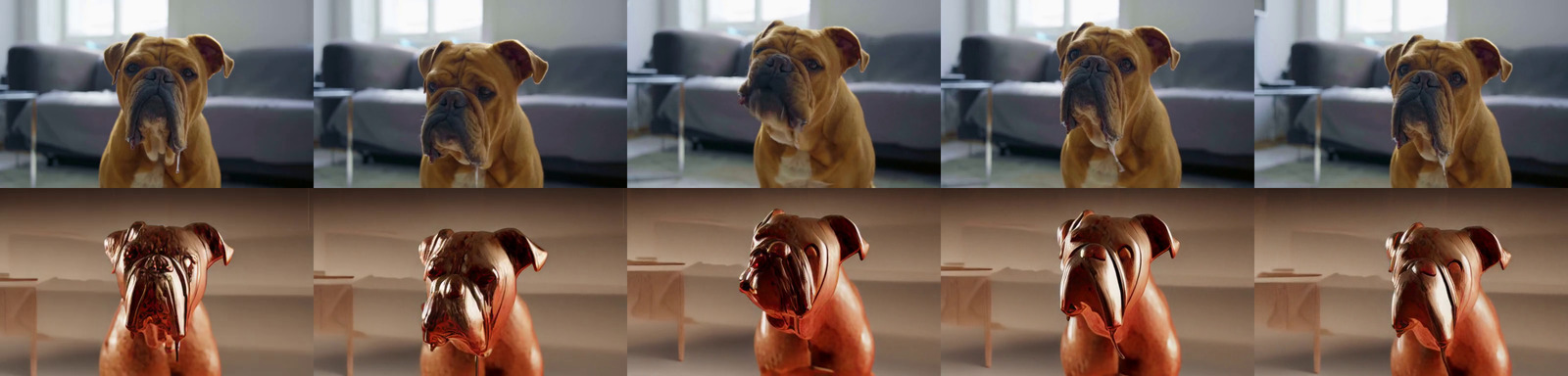}}
\\
\parbox{\imwidthB}{\includegraphics[width=\imwidthB]{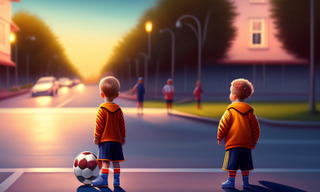}} & 
\parbox{\imwidth}{\includegraphics[width=\imwidth]{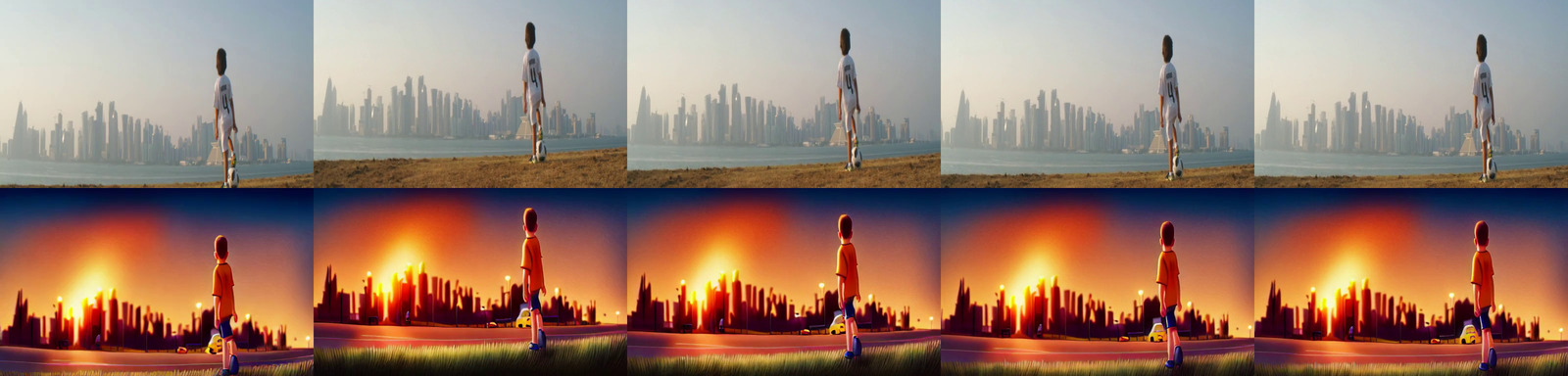}}
\\
\parbox{\imwidthB}{\includegraphics[width=\imwidthB]{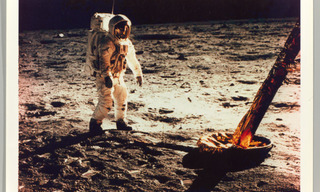}} & 
\parbox{\imwidth}{\includegraphics[width=\imwidth]{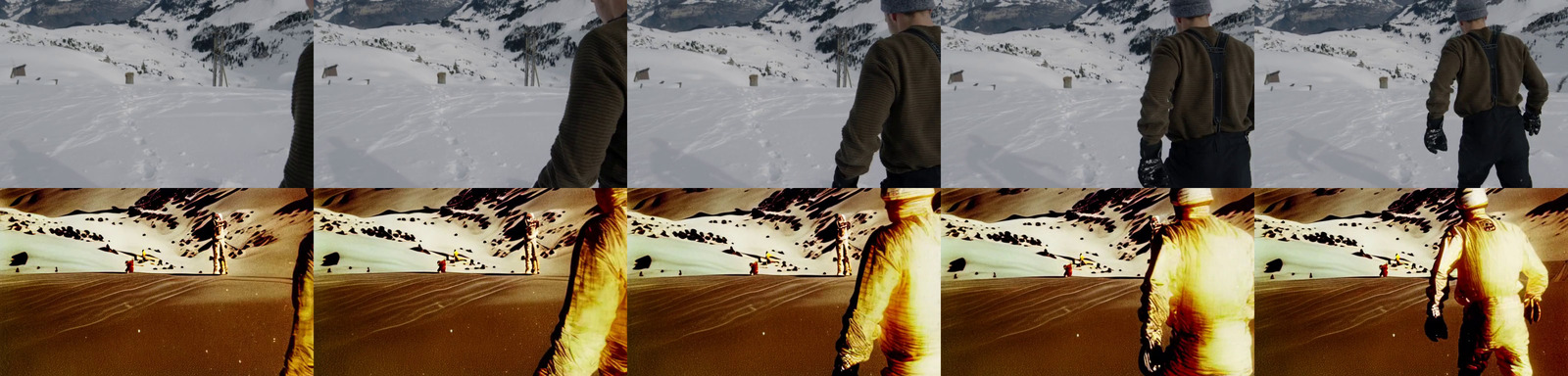}}

\vspace{1em}
\end{tabular}
  \caption{Additional results for image-to-video-editing.}
  \label{fig:imtovideditA}
\end{figure*}%
}

\newcommand{\figimtovideditB}{%
\begin{figure*}
  \centering
\renewcommand{\cellwidth}{0.1\textwidth}
\renewcommand{\imwidth}{0.9\textwidth}
\renewcommand{\imwidthB}{0.1\textwidth}
\renewcommand{\arraystretch}{7}
\begin{tabular}{cc}
  Prompt & Driving Video (top) and Result (bottom) \vspace{-2em}\\ \hline
\parbox{\imwidthB}{\includegraphics[width=\imwidthB]{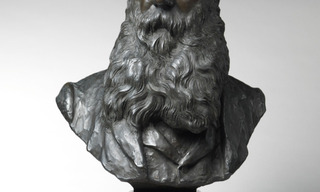}} & 
\parbox{\imwidth}{\includegraphics[width=\imwidth]{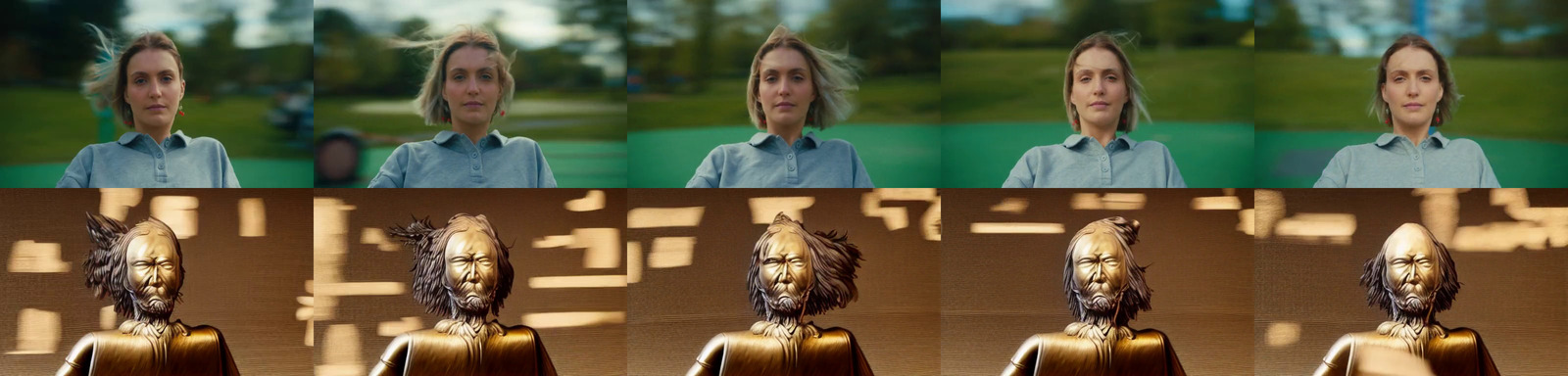}}
\\
\parbox{\imwidthB}{\includegraphics[width=\imwidthB]{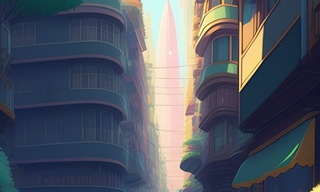}} & 
\parbox{\imwidth}{\includegraphics[width=\imwidth]{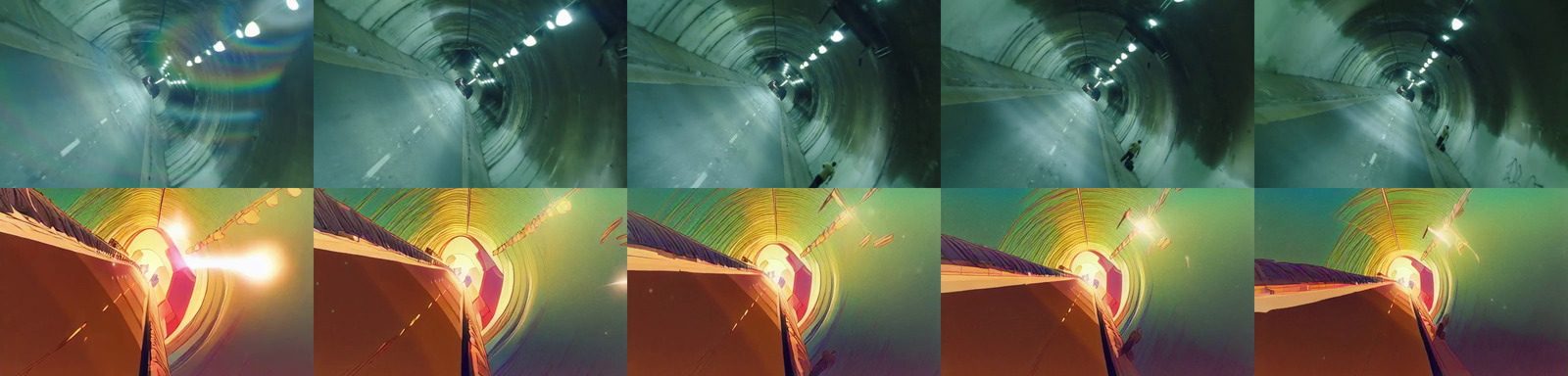}}
\\
\parbox{\imwidthB}{\includegraphics[width=\imwidthB]{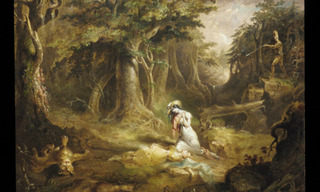}} & 
\parbox{\imwidth}{\includegraphics[width=\imwidth]{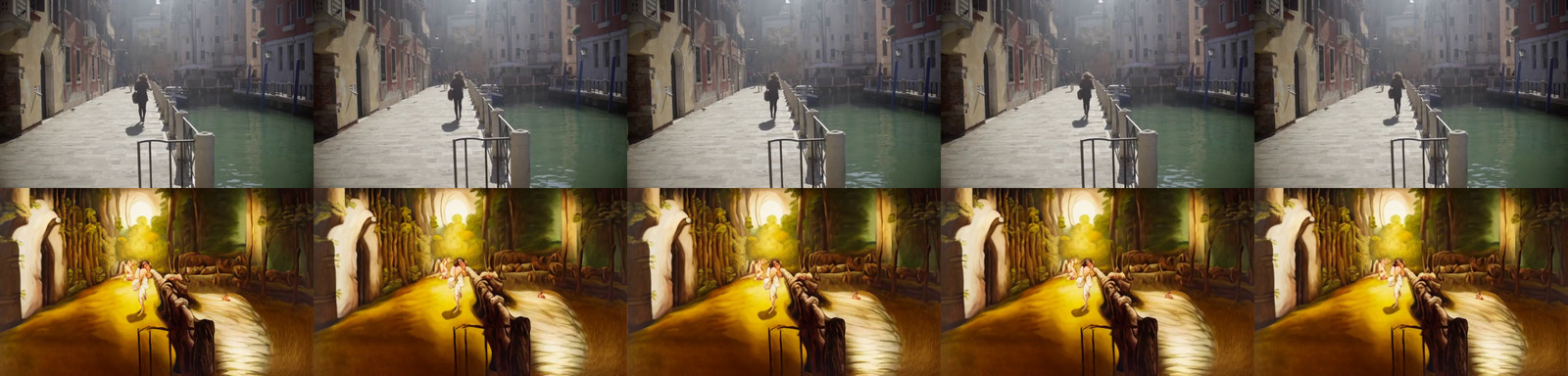}}
\\
\parbox{\imwidthB}{\includegraphics[width=\imwidthB]{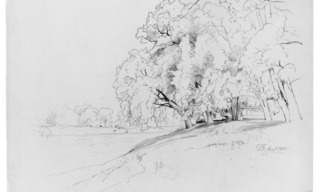}} & 
\parbox{\imwidth}{\includegraphics[width=\imwidth]{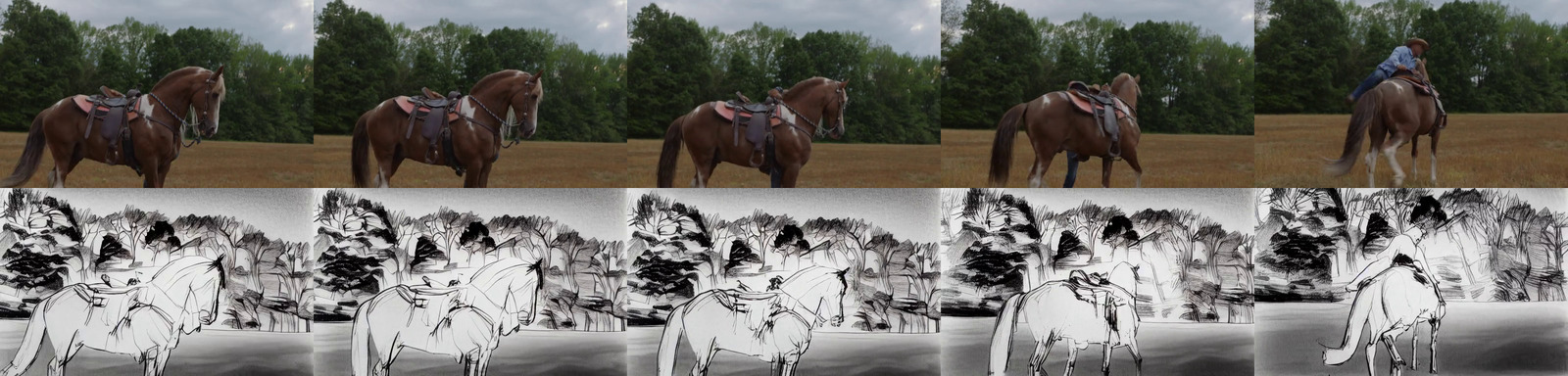}}
\\
\parbox{\imwidthB}{\includegraphics[width=\imwidthB]{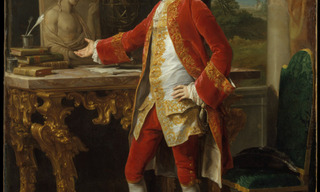}} & 
\parbox{\imwidth}{\includegraphics[width=\imwidth]{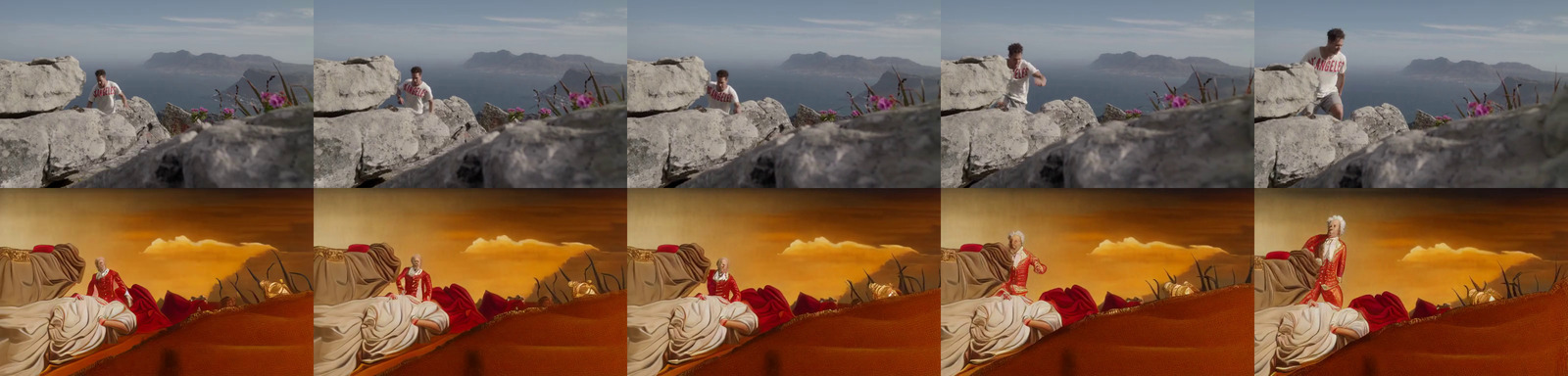}}

\vspace{1em}
\end{tabular}
  \caption{Additional results for image-to-video-editing.}
  \label{fig:imtovideditB}
\end{figure*}%
}

\newcommand{\figimtovideditC}{%
\begin{figure*}
  \centering
\renewcommand{\cellwidth}{0.1\textwidth}
\renewcommand{\imwidth}{0.9\textwidth}
\renewcommand{\imwidthB}{0.1\textwidth}
\renewcommand{\arraystretch}{7}
\begin{tabular}{cc}
  Prompt & Driving Video (top) and Result (bottom) \vspace{-2em}\\ \hline
\parbox{\imwidthB}{\includegraphics[width=\imwidthB]{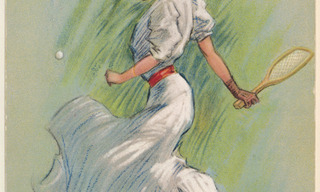}} & 
\parbox{\imwidth}{\includegraphics[width=\imwidth]{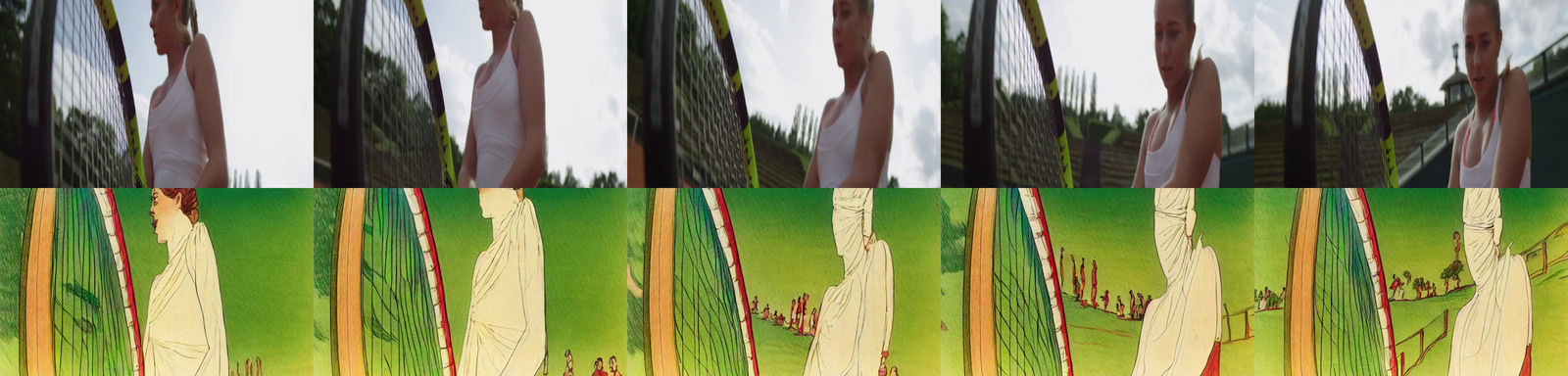}}
\\
\parbox{\imwidthB}{\includegraphics[width=\imwidthB]{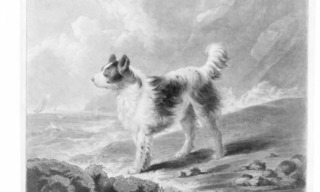}} & 
\parbox{\imwidth}{\includegraphics[width=\imwidth]{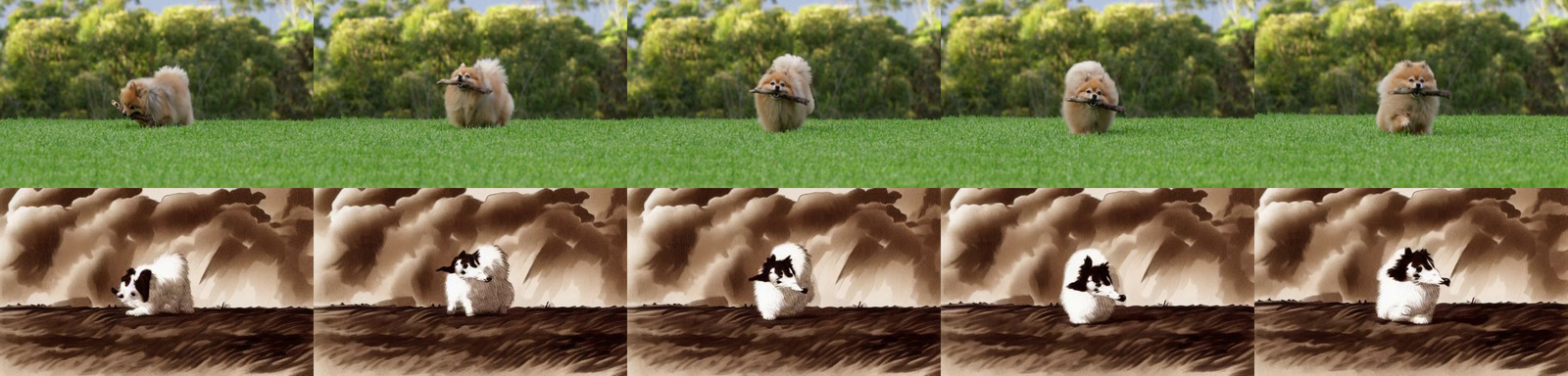}}
\\
\parbox{\imwidthB}{\includegraphics[width=\imwidthB]{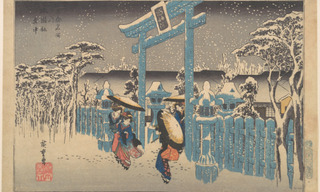}} & 
\parbox{\imwidth}{\includegraphics[width=\imwidth]{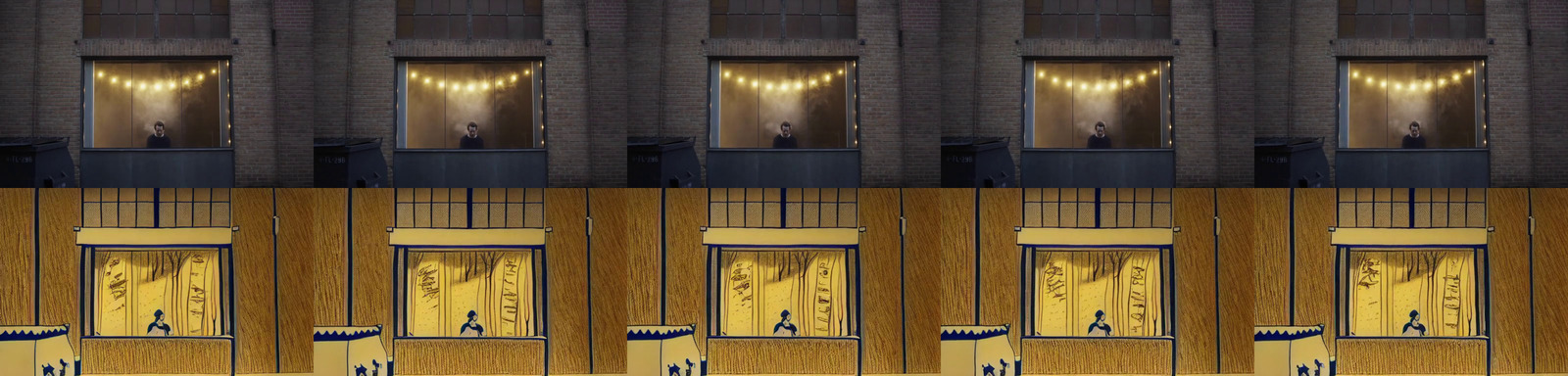}}
\\
\parbox{\imwidthB}{\includegraphics[width=\imwidthB]{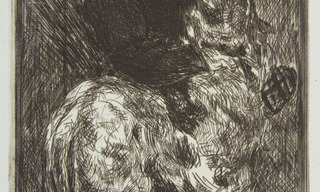}} & 
\parbox{\imwidth}{\includegraphics[width=\imwidth]{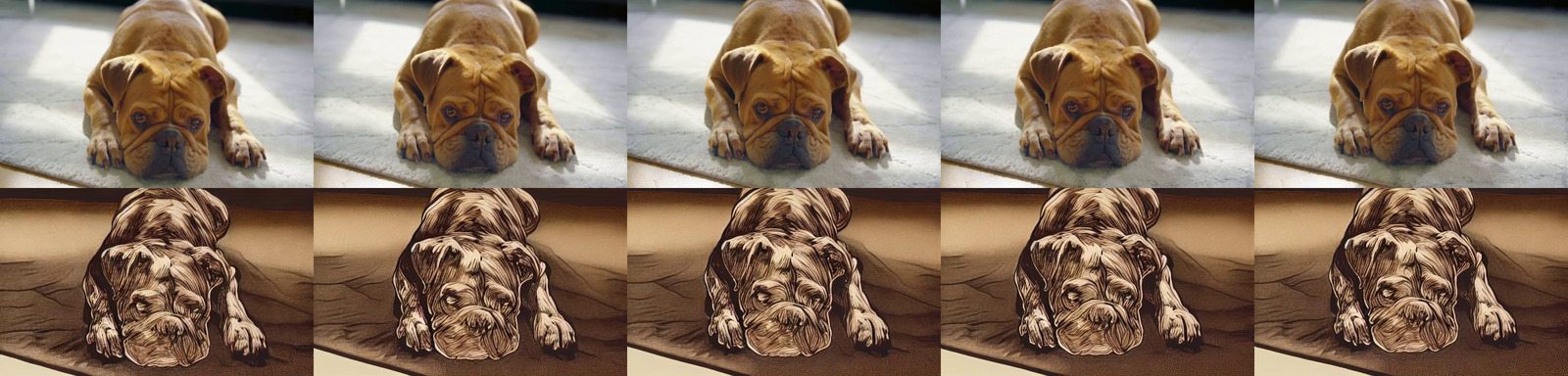}}
\\
\parbox{\imwidthB}{\includegraphics[width=\imwidthB]{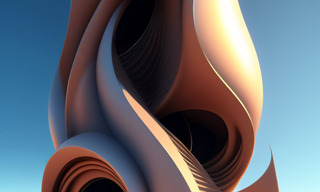}} & 
\parbox{\imwidth}{\includegraphics[width=\imwidth]{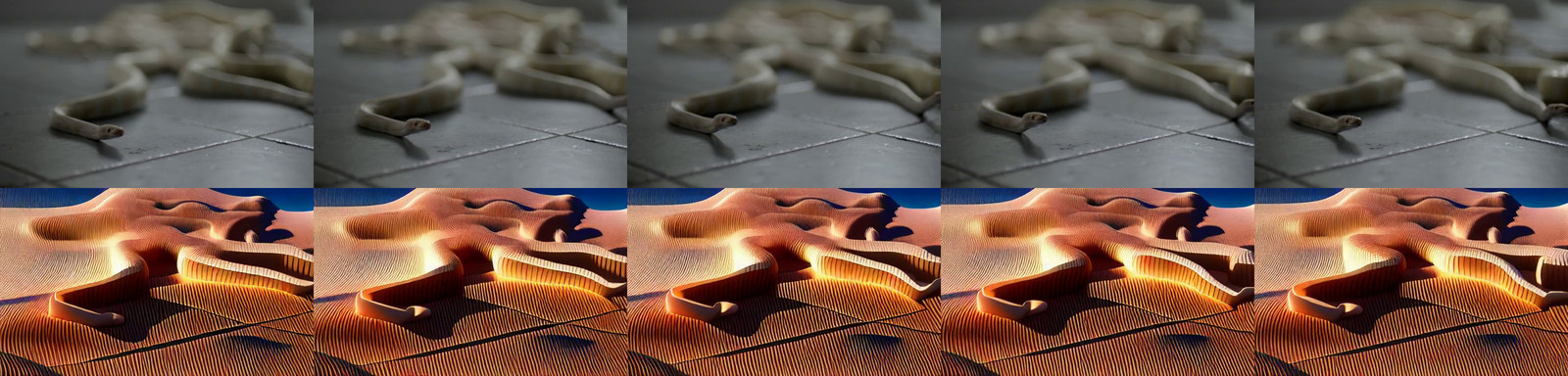}}

\vspace{1em}
\end{tabular}
  \caption{Additional results for image-to-video-editing.}
  \label{fig:imtovideditC}
\end{figure*}%
}

\newcommand{\figimtovideditD}{%
\begin{figure*}
  \centering
\renewcommand{\cellwidth}{0.1\textwidth}
\renewcommand{\imwidth}{0.9\textwidth}
\renewcommand{\imwidthB}{0.1\textwidth}
\renewcommand{\arraystretch}{7}
\begin{tabular}{cc}
  Prompt & Driving Video (top) and Result (bottom) \vspace{-2em}\\ \hline
\parbox{\imwidthB}{\includegraphics[width=\imwidthB]{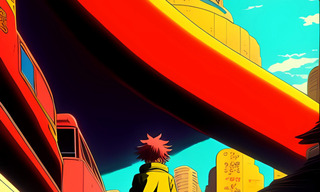}} & 
\parbox{\imwidth}{\includegraphics[width=\imwidth]{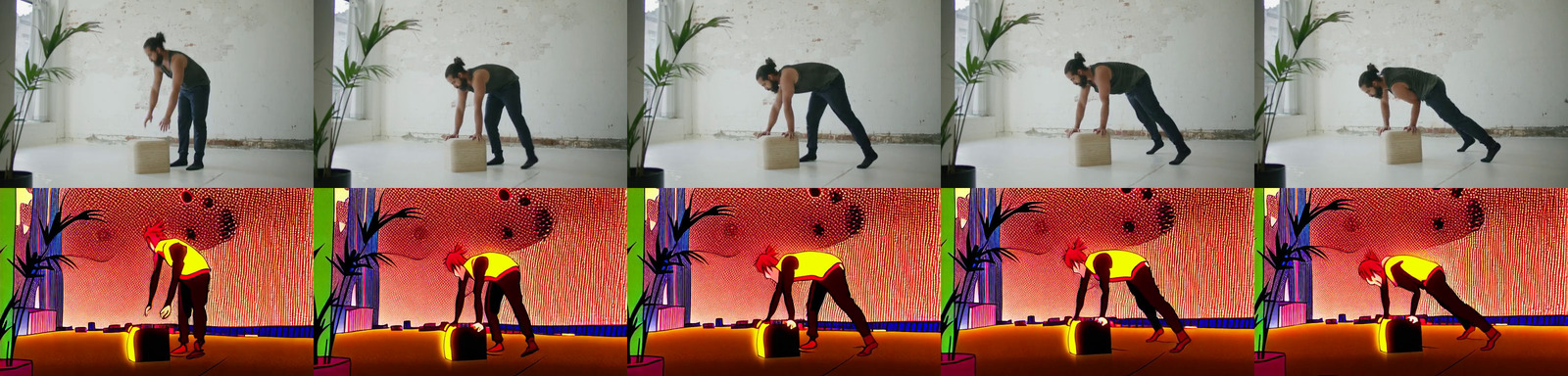}}
\\
\parbox{\imwidthB}{\includegraphics[width=\imwidthB]{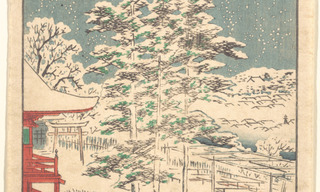}} & 
\parbox{\imwidth}{\includegraphics[width=\imwidth]{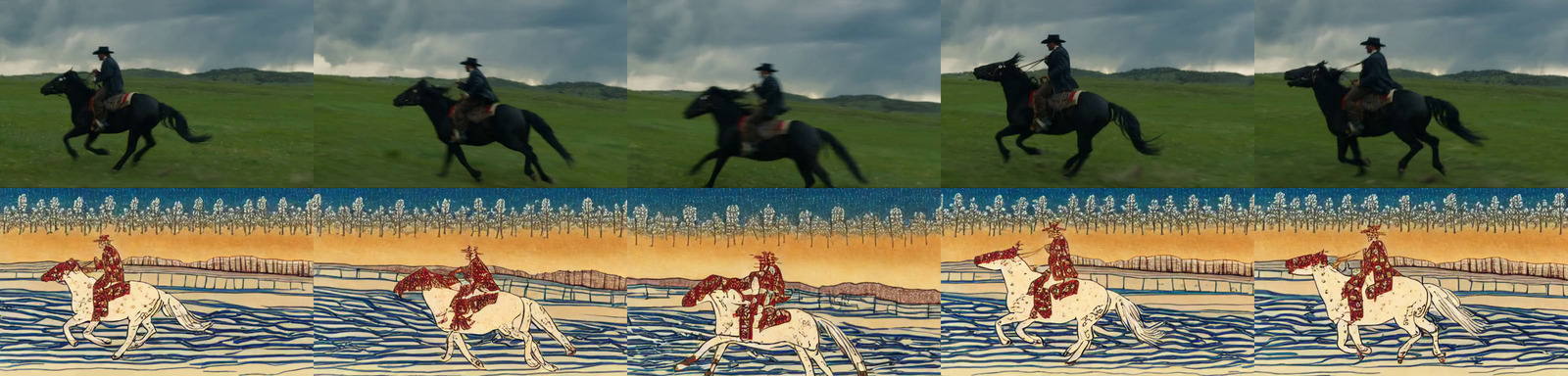}}
\\
\parbox{\imwidthB}{\includegraphics[width=\imwidthB]{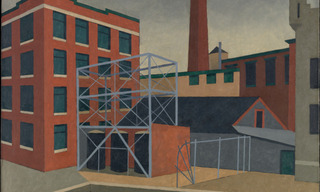}} & 
\parbox{\imwidth}{\includegraphics[width=\imwidth]{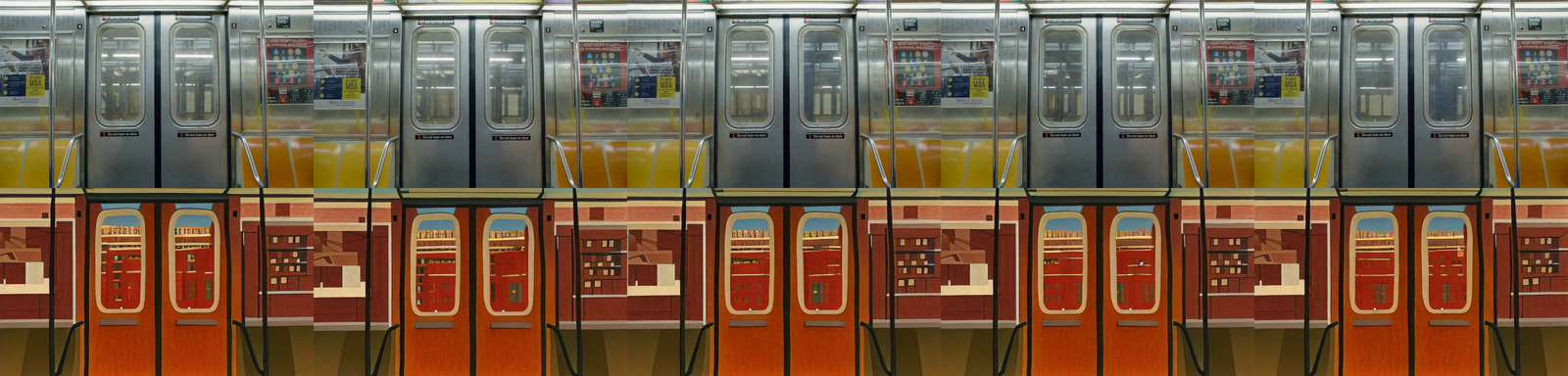}}
\\
\parbox{\imwidthB}{\includegraphics[width=\imwidthB]{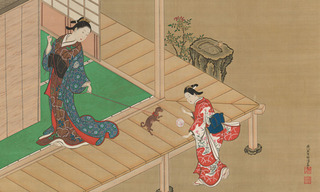}} & 
\parbox{\imwidth}{\includegraphics[width=\imwidth]{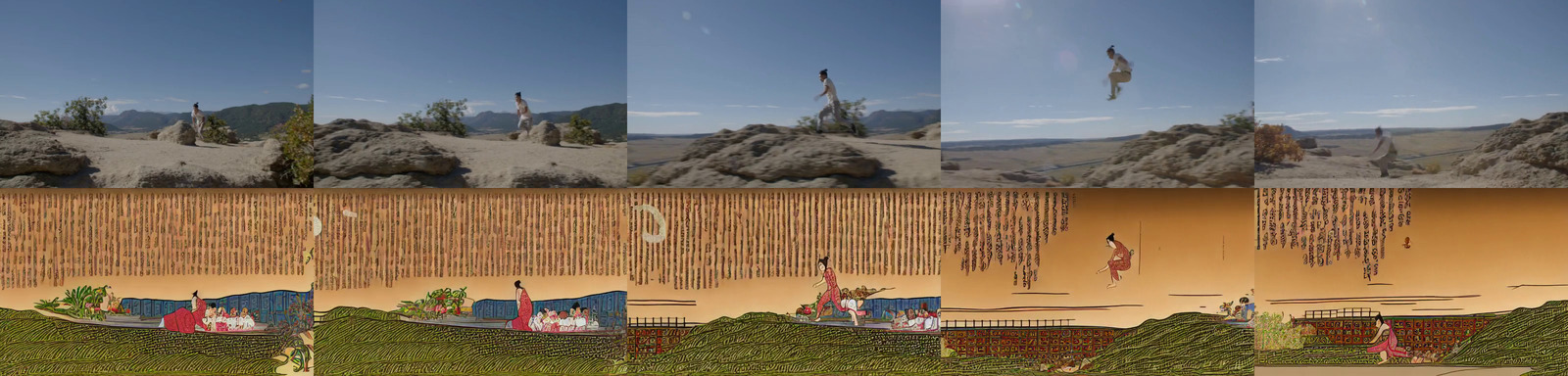}}
\\
\parbox{\imwidthB}{\includegraphics[width=\imwidthB]{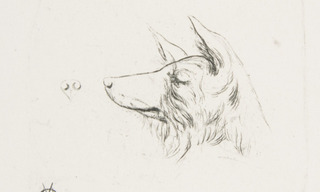}} & 
\parbox{\imwidth}{\includegraphics[width=\imwidth]{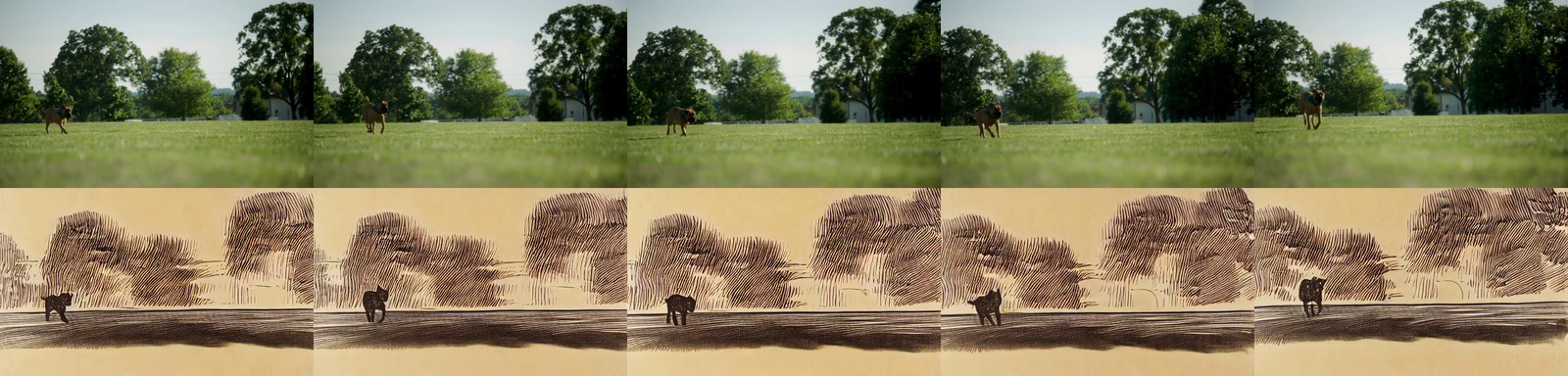}}

\vspace{1em}
\end{tabular}
  \caption{Additional results for image-to-video-editing.}
  \label{fig:imtovideditD}
\end{figure*}%
}
\newcommand{\figimtovideditE}{%
\begin{figure*}
  \centering
\renewcommand{\cellwidth}{0.1\textwidth}
\renewcommand{\imwidth}{0.9\textwidth}
\renewcommand{\imwidthB}{0.1\textwidth}
\renewcommand{\arraystretch}{7}
\begin{tabular}{cc}
  Prompt & Driving Video (top) and Result (bottom) \vspace{-2em}\\ \hline
\parbox{\imwidthB}{\includegraphics[width=\imwidthB]{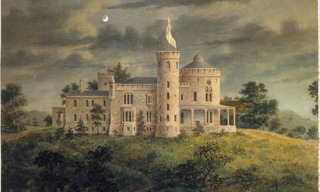}} & 
\parbox{\imwidth}{\includegraphics[width=\imwidth]{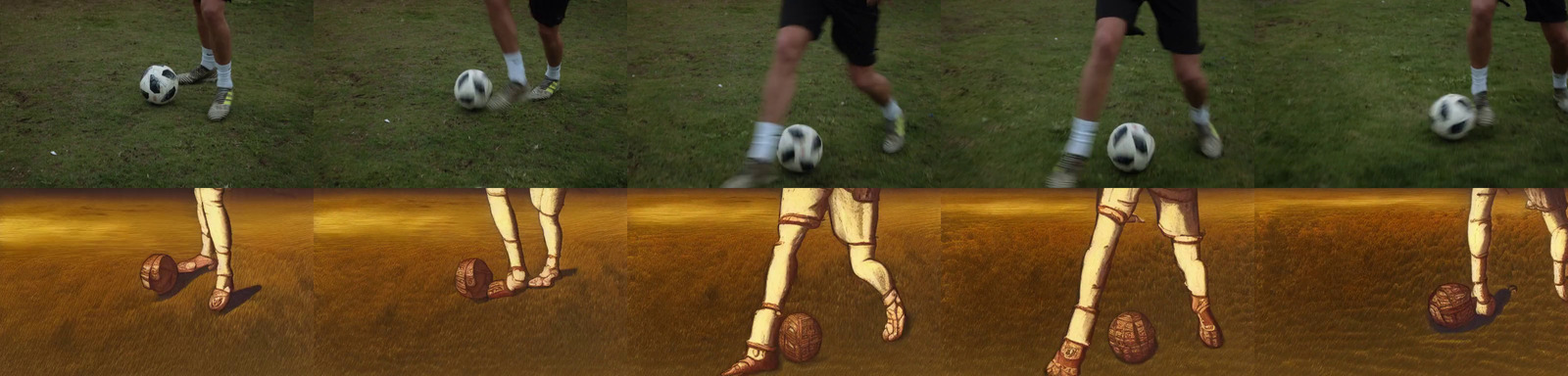}}
\\
\parbox{\imwidthB}{\includegraphics[width=\imwidthB]{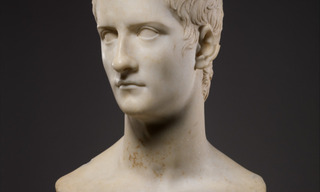}} & 
\parbox{\imwidth}{\includegraphics[width=\imwidth]{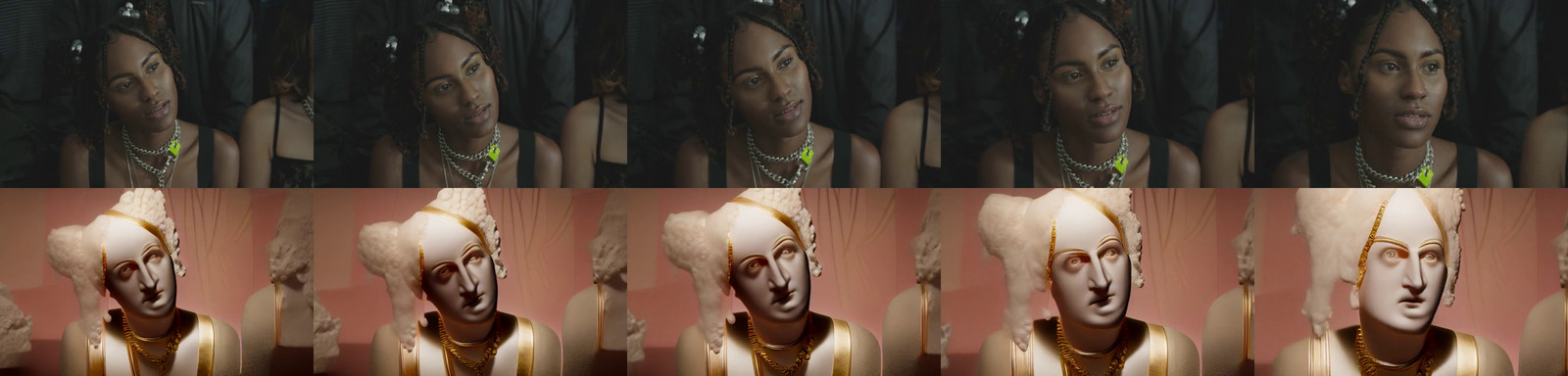}}
\\
\parbox{\imwidthB}{\includegraphics[width=\imwidthB]{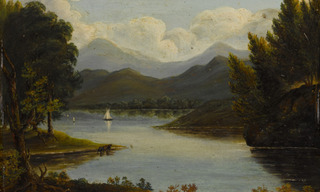}} & 
\parbox{\imwidth}{\includegraphics[width=\imwidth]{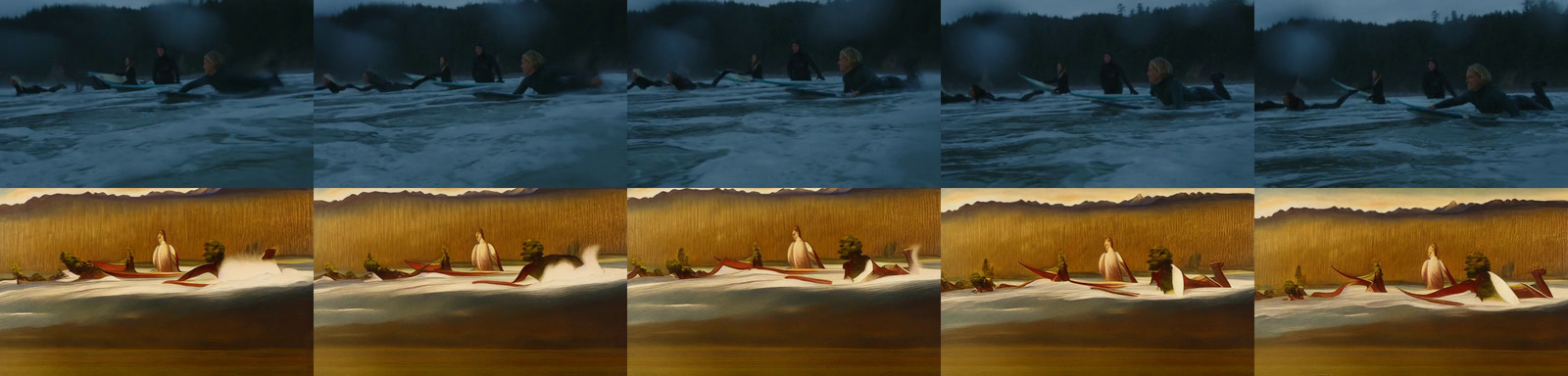}}
\\
\parbox{\imwidthB}{\includegraphics[width=\imwidthB]{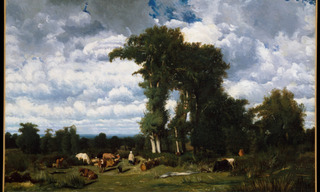}} & 
\parbox{\imwidth}{\includegraphics[width=\imwidth]{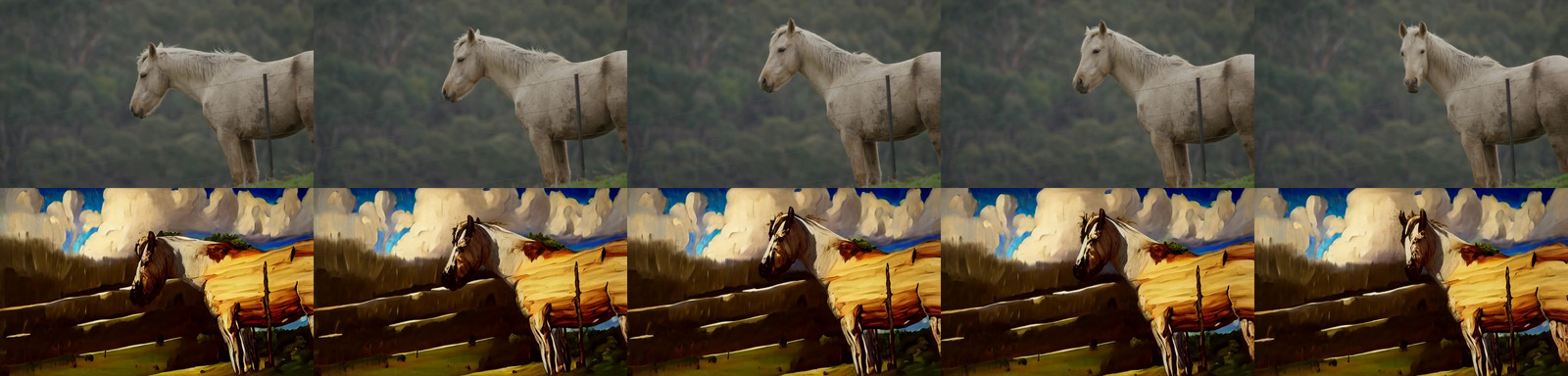}}

\vspace{1em}
\end{tabular}
  \caption{Additional results for image-to-video-editing.}
  \label{fig:imtovideditE}
\end{figure*}%
}

%% file: tabcommands.tex
\newcommand{\metricsplot}{
  \begin{figure}
    \includegraphics[width=\columnwidth]{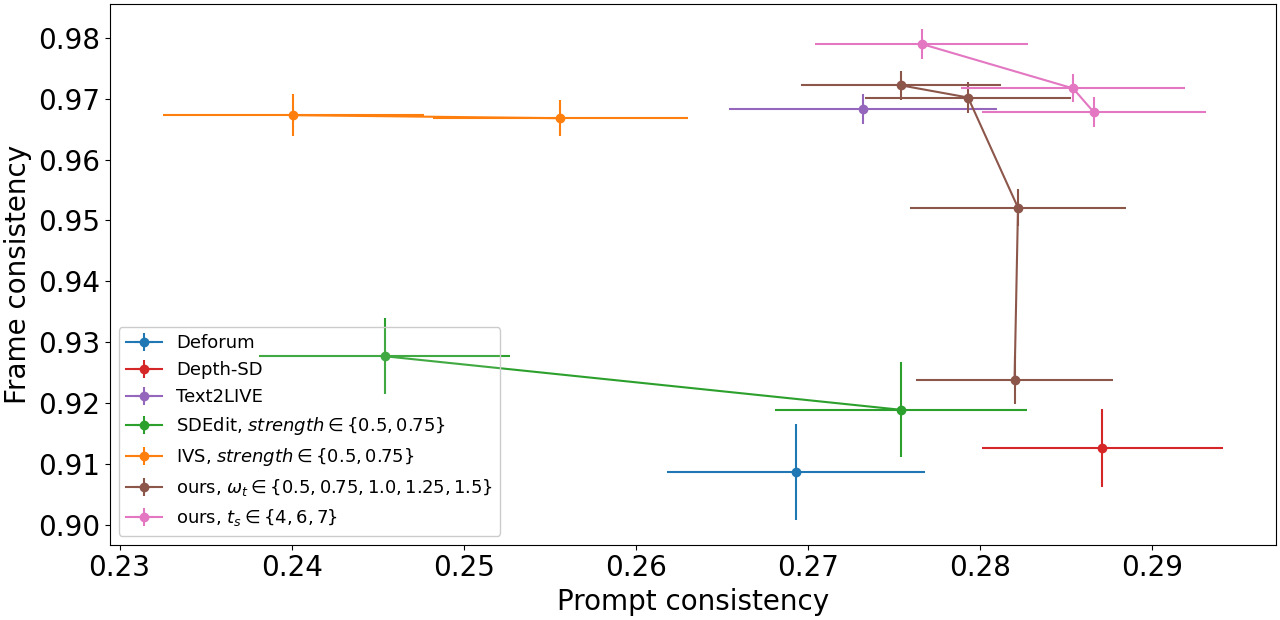}
    \caption{\textbf{Prompt-vs-frame consistency:} Image models such
    as SD-Depth achieve good prompt consistency but fail to produce consistent
    edits across frames. Propagation based approaches such as IVS and Text2Live
    increase frame consistency but fail to provide edits reflecting the prompt
    accurately. Our method achieves the best combination of frame and prompt
    consistency.}
    \label{fig:metrics}
  \end{figure}
}

\newcommand{\preferencesplot}{
  \begin{figure}
    \includegraphics[width=\columnwidth]{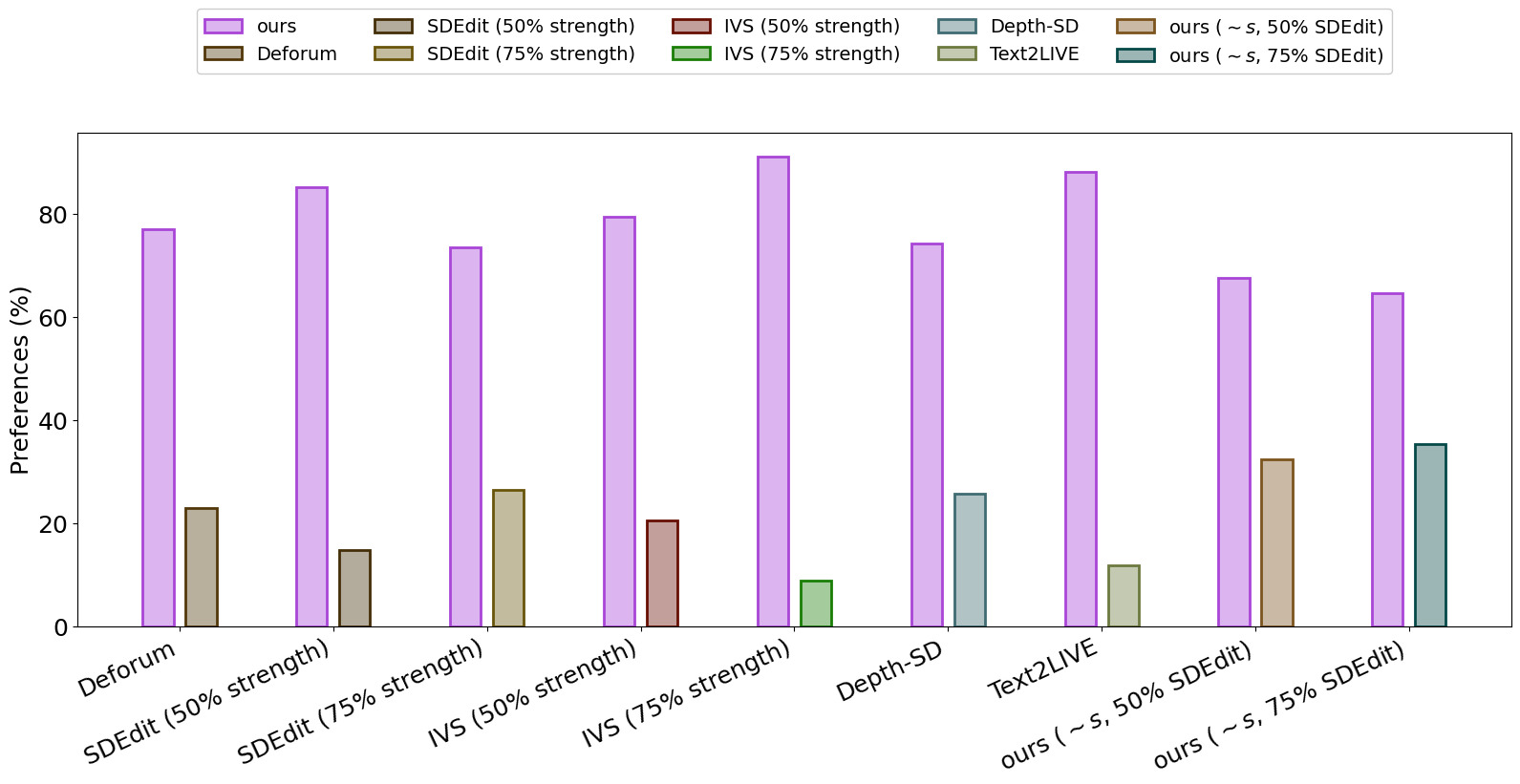}
    \caption{\textbf{User Preferences:} Based on our user study, the results 
    from our model are preferred over the baseline models.}
    \label{fig:preferences}
  \end{figure}
}

\newcommand{\tabmain}{%
\begin{table*}[b]
\renewcommand\cellset{\renewcommand\arraystretch{0.5}%
\setlength\extrarowheight{0pt}}
\renewcommand{\arraystretch}{1.6}

  \begin{tabular*}{\textwidth}{c@{\extracolsep{\fill}}cccc}
    \toprule
    method & \makecell{frame \\ consistency } & \makecell{prompt \\ consistency} & \makecell{ours \\ preferred} \\
    \midrule
Deforum & 0.9087 \scriptsize{$\pm 0.0079$} & 0.2693 \scriptsize{$\pm 0.0075$}  & $77.14\%$ \\
SDEdit, \scriptsize{strength=50\%} & 0.9277 \scriptsize{$\pm 0.0062$} & 0.2454 \scriptsize{$\pm 0.0073$}  & $85.29\%$ \\
SDEdit, \scriptsize{strength=75\%} & 0.9189 \scriptsize{$\pm 0.0078$} & 0.2754 \scriptsize{$\pm 0.0073$}  & $73.53\%$ \\
IVS, \scriptsize{strength=50\%} & 0.9673 \scriptsize{$\pm 0.0035$} & 0.2401 \scriptsize{$\pm 0.0076$} & $79.41\%$ \\
IVS, \scriptsize{strength=75\%} & 0.9668 \scriptsize{$\pm 0.0030$} & 0.2556 \scriptsize{$\pm 0.0074$}  & $91.18\%$ \\
Depth-SD & 0.9126 \scriptsize{$\pm 0.0064$} & 0.2871 \scriptsize{$\pm 0.0070$}  & $74.29\%$ \\
Text2LIVE & 0.9683 \scriptsize{$\pm 0.0025$} & 0.2732 \scriptsize{$\pm 0.0078$} & $88.24\%$ \\
\hline
    ours, \scriptsize{$\sim s$, strength=50\%} & 0.9541 \scriptsize{$\pm 0.0039$} & 0.2703 \scriptsize{$\pm 0.0074$}  & $67.65\%$ \\
    ours, \scriptsize{$\sim s$, strength=75\%} & 0.9482 \scriptsize{$\pm 0.0034$} & 0.2769 \scriptsize{$\pm 0.0062$}  & $64.71\%$ \\
\hline
ours, \scriptsize{$t_s=0$, $\omega_t=1.00$, $\omega=7.50$} & 0.9648 \scriptsize{$\pm 0.0031$} & 0.2805 \scriptsize{$\pm 0.0065$}  & - \\
\hline
ours, \scriptsize{$t_s=0$, $\omega_t=0.50$, $\omega=7.50$} & 0.9238 \scriptsize{$\pm 0.0039$} & 0.2820 \scriptsize{$\pm 0.0057$}  & - \\
ours, \scriptsize{$t_s=0$, $\omega_t=0.75$, $\omega=7.50$} & 0.9521 \scriptsize{$\pm 0.0030$} & 0.2822 \scriptsize{$\pm 0.0063$}  & - \\
ours, \scriptsize{$t_s=0$, $\omega_t=1.25$, $\omega=7.50$} & 0.9702 \scriptsize{$\pm 0.0026$} & 0.2793 \scriptsize{$\pm 0.0060$}  & - \\
ours, \scriptsize{$t_s=0$, $\omega_t=1.50$, $\omega=7.50$} & 0.9722 \scriptsize{$\pm 0.0024$} & 0.2754 \scriptsize{$\pm 0.0058$}  & - \\
\hline
ours, \scriptsize{$t_s=4$, $\omega_t=1.00$, $\omega=7.50$} & 0.9678 \scriptsize{$\pm 0.0025$} & 0.2866 \scriptsize{$\pm 0.0065$}  & - \\
ours, \scriptsize{$t_s=6$, $\omega_t=1.00$, $\omega=7.50$} & 0.9717 \scriptsize{$\pm 0.0023$} & 0.2854 \scriptsize{$\pm 0.0065$}  & - \\
ours, \scriptsize{$t_s=7$, $\omega_t=1.00$, $\omega=7.50$} & 0.9790 \scriptsize{$\pm 0.0025$} & 0.2766 \scriptsize{$\pm 0.0062$}  & - \\
    \bottomrule
  \end{tabular*}
  \caption{Quantiative evaluations corresponding to Fig.~\ref{fig:metrics} and Fig.~\ref{fig:preferences}. $\pm$ denotes standard error obtained
  with a sample size of 35.}
  \label{tab:data}
\end{table*}
}

%% file: macros.tex
\newcommand{\todo}[1]{\textcolor{magenta}{TODO: #1}}
\newcommand{\comment}[1]{\textcolor{blue}{Comment: #1}}

%% file: introduction.tex
\section{Introduction}
\label{sec:introduction}

Visual effects and video editing are ubiquitous in the modern media landscape. 
As such, demand for more intuitive and performant video editing tools has increased as video-centric platforms have been popularized. 
However, editing in the format is still complex and time-consuming due the temporal nature of video data.
State-of-the-art machine learning models have shown great promise in improving the editing process, but methods often balance temporal consistency with spatial detail.

Generative approaches for image synthesis recently experienced a rapid surge in quality and popularity due to the introduction of powerful diffusion models trained on large-scale datasets.
Text-conditioned models, such as DALL-E 2~\cite{ramesh2022dalle} and Stable Diffusion~\cite{rombach2021highresolution}, enable novice users to generate detailed imagery given only a text prompt as input. 
Latent diffusion models especially offer efficient methods for producing imagery via synthesis in a perceptually compressed space. 

Motivated by the progress of diffusion models in image synthesis, we investigate generative models suited for interactive applications in video editing.
Current methods repurpose existing image models by either propagating edits with approaches that compute explicit correspondences~\cite{bar2022text2live} or by finetuning on each individual video~\cite{wu2022tuneavideo}.
We aim to circumvent expensive per-video training and correspondence calculation to achieve fast inference for arbitrary videos.

We propose a controllable structure and content-aware video diffusion model trained on a large-scale dataset of uncaptioned videos and paired text-image data.
We opt to represent structure with monocular depth estimates and content with embeddings predicted by a pre-trained neural network.
Our approach offers several powerful modes of control in its generative process. 
First, similar to image synthesis models, we train our model such that the content of inferred videos, \eg their appearance or style, match user-provided images or text prompts (Fig. \ref{fig:teaser}).
Second, inspired by the diffusion process, we apply an information obscuring
process to the structure representation to enable selecting of how strongly the model adheres to the given structure. 
Finally, we also adjust the inference process via a custom guidance method,
inspired by classifier-free guidance, to enable control over temporal consistency in generated clips.

In summary, we present the following contributions:
\begin{compactitem}
    \item We extend latent diffusion models to video generation by introducing temporal layers into a pre-trained image model and training jointly on images and videos. 
    \item We present a structure and content-aware model that modifies videos guided by example images or texts. Editing is performed entirely at inference time without additional per-video training or pre-processing.
    \item We demonstrate full control over temporal, content and structure consistency. We show for the first time that jointly training on image and video data enables inference-time control over temporal consistency. For structure consistency, training on varying levels of detail in the representation allows choosing the desired setting during inference.
    \item We show that our approach is preferred over several other approaches in a user study.
    \item We demonstrate that the trained model can be further customized to generate more accurate videos of a specific subject by finetuning on a small set of images.
\end{compactitem}

\figoverview

%% file: related-work.tex
\section{Related Work}
\label{sec:related-work}

Controllable video editing and media synthesis is an active area of research. In this section, we review prior work in related areas and connect our method to these approaches. 

\noindent\textbf{Unconditional video generation}
\label{sec:video-generation-related}
Generative adversarial networks (GANs)~\cite{goodfellow2014gans} can learn to synthesize videos based on specific training data~\cite{vondrick2016generating,saito2017temporal,acharya2018towards,Tulyakov:2018:MoCoGAN}. 
These methods often struggle with stability during optimization, and produce fixed-length videos~\cite{vondrick2016generating,saito2017temporal} or longer videos where artifacts accumulate over time~\cite{skorokhodov2021styleganv}. 
\cite{brooks2022generating} synthesize longer videos at high detail with a custom positional encoding and an adversarially-trained model leveraging the encoding, but training is still restricted to small-scale datasets.  
Autoregressive transformers have also been proposed for unconditional video generation~\cite{ge2022long,yan2021videogpt}. However, our focus is on providing user control over the synthesis process whereas these approaches are limited to sampling random content resembling their training distribution.

\noindent\textbf{Diffusion models for image synthesis}
\label{sec:diffusion-models-image-related}
Diffusion models (DMs)~\cite{dickstein2015,song2020score} have recently attracted the attention of researchers and artists alike due to their ability to synthesize detailed imagery \cite{ramesh2022dalle, rombach2021highresolution}, and are now being applied to other areas of content creation such as motion synthesis~\cite{tevet2022human} and 3d shape generation~\cite{zeng2022lion}. 

Other works improve image-space diffusion by changing the parameterization~\cite{ho2020denoising, nichol2021improved,salimans2022progressive}, introducing advanced sampling methods~\cite{song2021denoising,lu2022dpmsolver,kong2021fast,san2021noise,Karras2022edm}, designing more powerful architectures~\cite{balaji2022eDiff-I,ho2022cascaded,vahdat2021score,peebles2022transformerdiffusion}, or conditioning on additional information~\cite{lugmayr2022repaint}.
Text-conditioning, based on embeddings from CLIP~\cite{radford2021learning} or T5~\cite{raffel:2020}, has become a particularly powerful approach for providing artistic control over model output~\cite{saharia2022photorealistic,nichol22glide, ramesh2022dalle,balaji2022eDiff-I,yu2022parti,ding2021cogview}. 
Latent diffusion models (LDMs)~\cite{rombach2021highresolution} perform diffusion in a compressed latent space reducing memory requirements and runtime. 
We extend LDMs to the spatio-temporal domain by introducing temporal connections into the architecture and by training jointly on video and image data. 

\noindent\textbf{Diffusion models for video synthesis}
\label{sec:diffusion-models-video-related}
Recently, diffusion models, masked generative models and autoregressive models have been applied to text-conditioned video synthesis~\cite{ho2022video,ho2022imagenvideo,villegas2022phenaki,zhou2022magicvideo,hong2022cogvideo,Singer2022MakeAVideoTG}.
Similar to \cite{ho2022video} and \cite{Singer2022MakeAVideoTG}, we extend image synthesis diffusion models to video generation by introducing temporal connections into a pre-existing image model. However, rather than synthesizing videos, including their structure and dynamics, from scratch, we aim to provide editing abilities on existing videos. While the inference process of diffusion models enables editing to some degree \cite{meng2021sdedit}, we demonstrate that our model with explicit conditioning on structure is significantly preferred.

\noindent\textbf{Video translation and propagation}
\label{sec:video-translation-related}
Image-to-image translation models, such as pix2pix~\cite{isola2017image,wang2018pix2pixHD}, can process each individual frame in a video, but this produces inconsistency between frames as the model lacks awareness of the temporal neighborhood. 
Accounting for temporal or geometric information, such as flow, in a video can increase consistency across frames when repurposing image synthesis models~\cite{sxela2022warpfusion, deforum}. 
We can extract such structural information to aid our spatio-temporal LDM in text- and image-guided video synthesis. 
Many generative adversarial methods, such as vid2vid \cite{Wang:2018:vid2vid,wang2018fewshotvid2vid}, leverage this type of input to guide synthesis combined with architectures specifically designed for spatio-temporal generation. However, similar to GAN-based approaches for images, results have been mostly limited to singular domains.

Video style transfer takes a reference style image and statistically applies its style to an input video~\cite{ruder2016styletransfer,chen2017coherent,texler2020videostylization}. 
In comparison, our method applies a mix of style and content from an input text prompt or image while being constrained by the extracted structure data. By learning a generative model from data, our approach produces semantically consistent outputs instead of matching feature statistics.

Text2Live~\cite{bar2022text2live} allows editing input videos using text
prompts by decomposing a video into neural layers \cite{kasten2021layered}. Once available, a layered video representation
\cite{ravacha2008unwrap} provides consistent propagation across frames.
SinFusion~\cite{nikankin2022sinfusion} can generate variations and extrapolations of videos by optimizing a diffusion model on a single video. 
Similarly, Tune-a-Video~\cite{wu2022tuneavideo} finetunes an image model converted to video generation on a single video to enable editing.
However, expensive per-video training limits the practicality of these approaches in creative tools. 
We opt to instead train our model on a large-scale dataset permitting inference on any video without individual training.

%% file: method.tex
\section{Method}
\label{sec:method}

For our purposes, it will be helpful to think of a video in terms of its
\textit{content} and \textit{structure}. 
By structure, we refer to characteristics describing its geometry and dynamics, \eg shapes and locations of subjects as well as their temporal changes.
We define content as features describing the appearance and semantics of the video, such as the colors and styles of objects and the lighting of the scene.
The goal of our model is then to edit the content of a video while retaining its structure.

To achieve this, we aim to learn a generative model $p(x \vert s, c)$ of videos
$x$, conditioned on representations of structure, denoted by $s$, and content,
denoted by $c$.
We infer the shape representation $s$ from an input video, and modify it based on a text
prompt $c$ describing the edit.
First, we describe our realization of the generative model as a conditional
latent video diffusion model and, then, we describe our choices for shape and
content representations. Finally, we discuss the optimization process of our model. See Fig. \ref{fig:overview} for an overview.

\subsection{Latent diffusion models}
\label{sec:latent-diffusion}

\paragraph{Diffusion models}
Diffusion models~\cite{dickstein2015} learn to reverse a fixed forward
diffusion process, which is defined as
\begin{equation}
    q(x_t|x_{t - 1}) \coloneqq \mathcal{N}(x_t, \sqrt{1 - \beta_t}x_{t - 1}, \beta_t\mathcal{I}) \; .
\end{equation}
Normally-distributed noise is slowly added to each sample $x_{t - 1}$ to obtain $x_{t}$. 
The forward process models a fixed Markov chain and the noise is dependent on a
variance schedule $\beta_t$ where $t \in \{1, \dots, T\}$, with $T$ being the
total number of steps in our diffusion chain, and $x_0 \coloneqq x$.

\noindent\textbf{Learning to Denoise}
The reverse process is defined according to the following equation with
parameters $\theta$
\begin{gather}\label{eqn:reverse-diffusion}
    p_\theta(x_0) \coloneqq \int p_\theta(x_{0:T})dx_{1:T} \\
    p_\theta(x_{0:T}) = p(x_T) \prod_{t = 1}^T p_\theta(x_{t - 1}|x_t), \\
    p_\theta(x_{t - 1}|x_t) \coloneqq \mathcal{N}(x_{t-1}, \mu_{\theta}(x_t, t), \Sigma_\theta(x_t, t)) \; .
\end{gather}
Using a fixed variance $\Sigma_\theta(x_t, t)$, we are left learning the means
of the reverse process $\mu_{\theta}(x_t, t)$. Training is typically performed
via a reweighted variational bound on the maximum likelihood objective,
resulting in a loss
\begin{equation}
  L \coloneqq \mathbb{E}_{t, q} \lambda_t \Vert \mu_t(x_t, x_0) - \mu_{\theta}(x_t, t) \Vert^2 \; ,
\end{equation}
where $\mu_t(x_t, x_0)$ is the mean of the forward process posterior $q(x_{t-1}
\vert x_t, x_0)$,
which is available in closed form \cite{ho2020denoising}.

\noindent\textbf{Parameterization}
The mean $\mu_{\theta}(x_t, t)$ is then predicted by a UNet architecture
\cite{ronneberger2015u} that receives the noisy input $x_t$ and the diffusion timestep
$t$ as inputs. Instead of directly predicting the mean, different combinations
of parameterizations and weightings, such as $x_0$, $\epsilon$ \cite{ho2020denoising} and
$v$-parameterizations \cite{salimans2022progressive} have been proposed, which can have
significant effects on sample quality. In early experiments, we found it
beneficial to use $v$-parameterization to improve color consistency of
video samples, similar to the findings of \cite{ho2022imagenvideo}, and therefore we use it for all
experiments.

\figblocks

\figtempcontrol

\noindent\textbf{Latent diffusion}
Latent diffusion models~\cite{rombach2021highresolution} (LDMs) take the diffusion
process into the latent space. This provides an improved separation between
compressive and generative learning phases of the model. Specifically, LDMs use
an autoencoder where an encoder $\mathcal{E}$ maps input data $x$ to a lower dimensional
latent code according to $z=\mathcal{E}(x)$ while a decoder $\mathcal{D}$ converts latent
codes back to the input space such that perceptually $x\approx
\mathcal{D}(\mathcal{E}(x))$.

Our encoder downsamples RGB-images $x \in \RR^{3 \times H \times W}$ by a
factor of eight and outputs four channels, resulting in a latent code $z \in
\RR^{4 \times H/8 \times W/8}$.
Thus, the diffusion UNet operates on a much
smaller representation which significantly improves runtime and memory
efficiency. The latter is particularly crucial for video modeling, where the
additional time-axis increases memory costs.

\subsection{Spatio-temporal Latent Diffusion}
\label{sec:architecture}

To correctly model a distribution over video frames, the architecture must 
take relationships between frames into account. At the same time, we
want to jointly learn an image model with shared parameters to benefit from better generalization
obtained by training on large-scale image datasets.

To achieve this, we extend an image architecture by introducing temporal layers, which are
only active for video inputs. All other layers are shared between the image and
video model. The autoencoder remains fixed and processes each frame in a video independently.

The UNet consists of two main building blocks: Residual blocks and transformer
blocks (see Fig.~\ref{fig:blocks}). Similar to \cite{ho2022video,Singer2022MakeAVideoTG}, we
extend them to videos by adding both 1D convolutions across time and 1D
self-attentions across time. In each residual block, we introduce one temporal
convolution after each 2D convolution. Similarly, after each spatial 2D
transformer block, we also include one temporal 1D transformer block, which
mimics its spatial counterpart along the time axis. We also input learnable
positional encodings of the frame index into temporal transformer blocks.

In our implementation, we consider images as videos with a single frame to
treat both cases uniformly. A batched tensor with batch size $b$, number of frames $n$, $c$ channels,
and spatial resolution $w \times h$ (\ie shape $b \times n \times c \times h \times w$) is rearranged to
$(b \cdot n) \times c \times h \times w$ for spatial layers, to $(b \cdot h \cdot w) \times c
\times n$ for temporal convolutions, and to $(b \cdot h \cdot w) \times
n \times c$ for temporal self-attention.

\subsection{Representing Content and Structure}
\label{sec:conditioning}
\noindent\textbf{Conditional Diffusion Models}
Diffusion models are well-suited to modeling conditional distributions such
as $p(x \vert s, c)$. In this case, the forward process $q$ remains unchanged
while the conditioning variables $s, c$ become additional inputs to the model.

We limit ourselves to uncaptioned video data for training due
to the lack of large-scale paired video-text datasets similar in
quality to image datasets such as \cite{schuhmann2021laion}.
Thus, while our goal is to edit an input video based on a text
prompt describing the desired edited video, we have neither training data of
triplets with a video, its edit prompt and the resulting output, nor even pairs of
videos and text captions.

Therefore, during training, we must derive structure and content representations
from the training video $x$ itself, \ie $s=s(x)$ and $c=c(x)$, resulting in a per-example loss of
\begin{equation}
  \lambda_t \Vert \mu_t(\mathcal{E}(x)_t, \mathcal{E}(x)_0) - \mu_{\theta}(\mathcal{E}(x)_t, t, s(x), c(x)) \Vert^2 \; .
\end{equation}

In contrast, during inference, structure $s$ and content $c$ are derived
from an input video $y$ and from a text prompt $t$ respectively. An edited version $x$
of $y$ is obtained by sampling the generative model conditioned on $s(y)$ and $c(t)$:
\begin{equation}
  z \sim p_\theta(z \vert s(y), c(t)), \;\;
  x = \mathcal{D}(z) \; .
\end{equation}

\noindent\textbf{Content Representation}
To infer a content representation from both text inputs $t$ and
video inputs $x$, we follow previous works \cite{ramesh2021dalle,balaji2022eDiff-I} and
utilize CLIP \cite{radford2021learning} image embeddings to represent content. For video
inputs, we select one of the input frames randomly during training. Similar to
\cite{ramesh2021dalle,Singer2022MakeAVideoTG}, one can then train a prior model that allows sampling
image embeddings from text embeddings. This approach enables specifying edits
through image inputs instead of just text. 

Decoder visualizations demonstrate that CLIP embeddings have increased
sensitivity to semantic and stylistic properties
while being more invariant towards precise geometric attributes, such as sizes
and locations of objects \cite{ramesh2022dalle}.
Thus, CLIP embeddings are a fitting representation for content as
structure properties remain largely orthogonal.

\figtexttovidedit

\noindent\textbf{Structure Representation}
A perfect separation of content and structure is difficult.
Prior knowledge about semantic object classes in videos
influences the probability of certain shapes appearing in a video. Nevertheless, we can
choose suitable representations to introduce inductive biases that guide our
model towards the intended behavior while decreasing correlations between structure
and content.

We find that depth estimates extracted from input video frames provide the
desired properties as they encode significantly less content information
compared to simpler structure representations.
For example, edge filters also detect textures in
a video which limits the range of artistic control over content in videos.
Still, a fundamental overlap between content and structure information
remains with our choice of CLIP image embeddings as a content representation
and depth estimates as a structure representation. Depth maps 
reveal the silhouttes of objects which prevents content
edits involving large changes in object shape. 

To provide more control over
the amount of structure to preserve, we propose to
train a model on structure representations with varying amounts of
information.
We employ an information-destroying process based on a blur operator,
which improves stability compared to other approaches such as adding noise.
Similar to the diffusion timestep $t$, we provide the structure blurring level $t_s$
as an input to the model. We note that blurring has also been explored as a forward process for
generative modeling \cite{bansal2023cold}.

While depths map work well for our usecase, our
approach generalizes to other geometric guidance features or combinations of features 
that might be more helpful for other specific applications. For example,
models focusing on human video synthesis might benefit from estimated poses or
face landmarks. 

\noindent\textbf{Conditioning Mechanisms}
We account for the different characteristics of our content and structure with two
different conditioning mechanisms. Since structure represents a significant portion of the
spatial information of video frames, we use concatenation for
conditioning to make effective use of this information. In contrast, attributes
described by the content representation are not tied to particular locations.
Hence, we leverage cross-attention which can effectively
transport this information to any position.

We use the spatial transformer blocks of the UNet architecture for
cross-attention conditioning. Each contains two attention operations,
where the first one perform a spatial self-attention and the second one
a cross attention with keys and values computed from the CLIP image
embedding.

To condition on structure, we first estimate depth maps for all input frames
using the MiDaS DPT-Large model~\cite{Ranftl2019TowardsRM}.
We then apply $t_s$ iterations of blurring and downsampling to the
depth maps, where $t_s$ controls the amount of structure to preserve from the
input video. During training, we randomly sample $t_s$ between $0$ and $T_s$. At
inference, this parameter can be controlled to achieve different editing
effects (see Fig.~\ref{fig:fidelitycontrolsintelcolumn}). 
We resample the perturbed depth map to the resolution of the RGB-frames
and encode it using $\mathcal{E}$. This latent representation of
structure is concatenated with the input $z_t$ given to the UNet. We also
input four channels containing a sinusoidal embedding of $t_s$.

\metricsplot

\noindent\textbf{Sampling}
While Eq.~\eqref{eqn:reverse-diffusion} provides a direct way to sample from the
trained model, many other sampling methods \cite{song2021denoising,lu2022dpmsolver,kong2021fast}
require only a fraction of the number of diffusion timesteps to achieve good sample quality.
We use DDIM \cite{song2021denoising} throughout our experiments. Furthermore, classifier-free diffusion guidance \cite{ho2022classifier} significantly improves sample quality. For a conditional model $\mu_\theta(x_t, t, c)$, this is achieved by training the model to also perform unconditional predictions $\mu_\theta(x_t, t, \emptyset)$ and then adjusting predictions during sampling according to
\begin{equation*}
  \tilde{\mu}_\theta(x_t, t, c) = \mu_\theta(x_t, t, \emptyset) + \omega (\mu_\theta(x_t, t, c) - \mu_\theta(x_t, t, \emptyset))
\end{equation*}
where $\omega$ is the guidance scale that controls the strength.
Based on the intuition that $\omega$ extrapolates the direction between an unconditional and a conditional model, we apply this idea to control temporal consistency of our model. Specifically, since we are training both an image and a video model with shared parameters, we can consider predictions by both models for the same input. Let $\mu_\theta(z_t, t, c, s)$ denote the prediction of our video model, and let $\mu^\pi_\theta(z_t, t, c, s)$ denote the prediction of the image model applied to each frame individually. Taking classifier-free guidance for $c$ into account, we then adjust our prediction according to
\begin{equation}
  \begin{aligned}
    \tilde{\mu}_\theta(z_t, t, c, s) &= \mu^\pi_\theta(z_t, t, \emptyset, s) \\
    &+ \omega_t (\mu_\theta(x_t, t, \emptyset, s) - \mu^\pi_\theta(x_t, t, \emptyset, s)) \\
    &+ \omega (\mu_\theta(x_t, t, c, s) - \mu_\theta(x_t, t, \emptyset, s))
  \end{aligned}
\end{equation}
Our experiments demonstrate that this approach controls temporal consistency in the outputs, see Fig.~\ref{fig:tempcontrol}.

\subsection{Optimization}
\label{sec:optimization}

We train on an internal dataset of 240M images and a custom dataset of 6.4M video clips. We use image batches of
size 9216 with resolutions of $320 \times 320$, $384 \times 320$ and $448 \times 256$, as
well as the same resolutions with flipped aspect ratios.
We sample image batches with a probabilty of 12.5\%. For
the main training, we use video batches containing 8 frames sampled four frames
apart with a resolution of $448 \times 256$ and a total video batch size of
1152.

We train our model in multiple stages. 
First, we initialize model weights based on a pretrained text-conditional
latent diffusion model
\cite{rombach2021highresolution}\footnote{\url{https://github.com/runwayml/stable-diffusion}}.
We change the conditioning from CLIP text embeddings to CLIP image embeddings
and fine-tune for 15k steps on images only.
Afterwards, we introduce temporal connections as described in
Sec.~\ref{sec:architecture} and train jointly on images and videos for 75k steps.
We then add conditioning on structure $s$ with $t_s\equiv 0$ fixed and train for
25k steps. Finally, we resume training with $t_s$ sampled uniformly between $0$
and $7$ for another 10k steps.

\preferencesplot

%% file: results.tex
\section{Results}
\label{sec:results}

\figmaskededit

To evaluate our approach, we use videos from DAVIS ~\cite{davis2017} and various stock footage.
To automatically create edit prompts, we first
run a captioning model \cite{li2022blip} to obtain a description of the original
video content. We then use GPT-3 \cite{brown2020gpt3} to generate edited prompts.

\subsection{Qualitative Results}
We demonstrate that our approach performs well on 
a number of diverse inputs (see Fig.~\ref{fig:texttovidedit}).
Our method handles static shots (first row) as
well as shaky camera motion from selfie videos (second row) without any explicit
tracking of the input videos. We also see that it handles a large variety of
footage such as landscapes and close-ups.
Our approach is not limited to a specific domain of subjects thanks to
its general structure representation based on depth estimates.
The generalization obtained from training simultaneously on large-scale image and video datasets
enables many editing capabilities, including changes to animation styles such
as anime (first row) or claymation (second row), changes in the scene
environment, \eg changing day to sunset (third row) or summer to winter (fourth
row), as well as various changes to characters in a scene, \eg turning a
hiker into an alien (fifth row) or turning a bear in nature into a space bear
walking through the stars (sixth row).

Using content representations through CLIP image embeddings allows users to 
specify content through images. One particular example application is character replacement,
as shown in Fig.~\ref{fig:swapmatrix}. We demonstrate this application using a set of six videos.
For every video in the set, we re-synthesize it five times, each time providing a single content 
image taken from another video in the set. We can retain content 
characteristics with $t_s=3$ despite large differences in their pose and shape.

Lastly, we are given a great deal of flexibilty during inference due to our application of versatile diffusion models. 
We illustrate the use of masked video editing in Fig.~\ref{fig:maskededit}, where our goal
is to have the model predict everything outside the masked area(s) while retaining the original 
content inside the masked area. Notably, this technique resembles approaches for inpainting with 
diffusion models \cite{saharia2021palette,lugmayr2022repaint}.
In Sec.~\ref{sec:control}, we also evaluate the ability of our approach to control 
other characteristics such as temporal consistency and adherence to the input structure.

\subsection{User Study}
\label{sec:user-study}
Text-conditioned video-to-video translation is a nascent area of computer
vision and thus find a limited number of methods to compare against. 
We benchmark against Text2Live~\cite{bar2022text2live}, a recent approach for text-guided video 
editing that employs layered neural atlases~\cite{kasten2021layered}. As a baseline, we 
compare against SDEdit~\cite{meng2021sdedit} in two ways; per-frame generated results and a 
first-frame result propagated by a few-shot video stylization method~\cite{texler2020videostylization} (IVS).
We also include two depth-based versions of Stable Diffusion; one trained with depth-conditioning~\cite{sddepth} and 
one that retains past results based on depth estimates~\cite{deforum}. We also include an ablation: applying SDEdit 
to our video model trained without conditioning 
on a structure representation (ours, $\sim s$).

We judge the success of our method qualitatively based on a user study. We run the user study 
using Amazon Mechanical Turk (AMT) on an evaluation set of 35 representative video editing prompts.
For each example, we ask 5 annotators to compare faithfulness to the video editing prompt 
("Which video better represents the provided edited caption?") between a baseline and our method, 
presented in random order, and use a majority vote for the final result.

The results can be found in Fig.~\ref{fig:preferences}. Across all compared methods, 
results from our approach are preferred roughly 3 out of 4 times. A visual comparison 
among the methods can be found in Fig.~\ref{fig:compare}. We observe that SDEdit is quite 
sensitive to the editing strength. Low values often do not achieve the desired editing effect
and high values change the structure of the input, \eg in
Fig.~\ref{fig:compare} the elephant looks into another direction after the
edit. While the use of a fixed seed is able to keep the overall color of
outputs consistent across frames, both style and structure can change in
unnatural ways between frames as their relationship is not modeled by
image based approaches. Overall, we observe that deforum behaves very similarly.
Propagation of SDEdit outputs with few-shot video
stylization leads to more consistent results, but often introduces propagation
artifacts, especially in case of large camera or subject movements.
Depth-SD produces accurate, structure-preserving
edits on individual frames but without modeling temporal relationships, frames
are inconsistent across time.

The quality of Text2Live outputs varies a lot. Due to its
reliance in Layered Neural Atlases \cite{kasten2021layered}, the outputs tend to be
temporally smooth but it often struggles to perform edits that represent the
edit prompt accurately. A direct comparison is difficult as Text2Live requires input masks and edit prompts for foreground and background. In addition, computing a neural atlas takes about 10 hours whereas our approach requires approximately a minute.

\figswapmatrix
\figfidelitycontrolsintelcolumn

\subsection{Quantitative Evaluation}\label{sec:control}

We quantify trade-offs between frame consistency and prompt consistency with the following two metrics. \\
\noindent \textbf{Frame consistency} We compute CLIP image embeddings on all frames of
output videos and report the average cosine similarity between all pairs
of consecutive frames. \\
\noindent \textbf{Prompt consistency} We compute CLIP image embeddings on all frames of
output videos and the CLIP text embedding of the edit prompt. We report
average cosine similarity between text and image embedding over
all frames.

Fig.~\ref{fig:metrics} shows the results of each model using our frame consistency 
and prompt consistency metrics. 
Our model tends to outperform the baseline models in both aspects (placed 
higher in the upper-right quadrant of the graph). We also notice a slight tradeoff with increasing
the strength parameters in the baseline models: larger strength scales implies higher prompt 
consistency at the cost of lower frame consistency.
Increasing the temporal scale ($\omega_t$) of our model results in higher frame consistency but lower prompt consistency.
We also observe that an increased structure scale ($t_s$) results in higher prompt consistency as the content becomes less determined by the input structure.

\subsection{Customization}
\label{sec:customization}

Customization of pretrained image synthesis models allows users to generate images of custom content, such as people or image styles, based on a small training dataset for finetuning~\cite{ruiz2022dreambooth}. 
To evaluate customization of our depth-conditioned latent video diffusion model,
we finetune it on a set of 15-30 images and produce novel content containing the desired subject. 
During finetuning, half of the batch elements are of the custom subject and the other half are of the original training dataset to avoid overfitting. 

Fig.~\ref{fig:fidelitycontrolsintelcolumn} shows an example with different numbers of customization steps as well as different levels of structure adherence $t_s$. We observe that customization improves fidelity to the style and appearance of the character, such that in combination with higher values for $t_s$ accurate animations are possible despite using a driving video of a person with different characteristics.

%% file: conclusion.tex
\section{Conclusion}
\label{sec:conclusion}

Our latent video diffusion model synthesizes new videos given structure and content information.
We ensure structural consistency by conditioning on depth estimates while content is controlled with images or natural language.
Temporally stable results are achieved with additional temporal connections in the model and joint image and video training.
Furthermore, a novel guidance method, inspired by classifier-free guidance, allows for user control over temporal consistency in outputs. 
Through training on depth maps with varying degrees of fidelity, we expose the ability to adjust the level of structure preservation 
which proves especially useful for model customization. Our quantitative evaluation and user study show that our 
method is highly preferred over related approaches. Future works could investigate other conditioning data,
such as facial landmarks and pose estimates, and additional 3d-priors to improve stability of generated results. We do not intend for the 
model to be used for harmful purposes but realize the risks and hope that further work is aimed at combating abuse of generative models. 

%% file: supp.tex
\FloatBarrier
\clearpage
\appendix
\onecolumn
\renewcommand{\thefigure}{S\arabic{figure}}
\setcounter{figure}{0}
\renewcommand{\thetable}{S\arabic{table}}
\setcounter{table}{0}

\begin{center}
  \textbf{
    \Large Structure and Content-Guided Video Synthesis with Diffusion Models} \\
  \Large 
-- \\
 \textbf{\large Supplementary Material} \\
\hspace{1cm}
\end{center}

We include the raw data of Fig.~\ref{fig:metrics} and Fig.~\ref{fig:preferences} in Tab.~\ref{tab:data}. Fig.~\ref{fig:texttovideditA}-\ref{fig:texttovideditG} contain additional results for text based edits, Fig.~\ref{fig:imtovideditA}-\ref{fig:imtovideditE} for image based edits. Fig.~\ref{fig:compare} shows a qualitative comparison.

\figtexttovideditA
\figtexttovideditB
\figtexttovideditC
\figtexttovideditD
\figtexttovideditE
\figtexttovideditF
\figtexttovideditG

\figimtovideditA
\figimtovideditB
\figimtovideditC
\figimtovideditD
\figimtovideditE

\figcomparative
\tabmain

%% file: ms.bbl
\begin{thebibliography}{10}\itemsep=-1pt

\bibitem{acharya2018towards}
Dinesh Acharya, Zhiwu Huang, Danda~Pani Paudel, and Luc Van~Gool.
\newblock Towards high resolution video generation with progressive growing of
  sliced wasserstein gans.
\newblock {\em arXiv preprint arXiv:1810.02419}, 2018.

\bibitem{sddepth}
Stability AI.
\newblock Stable diffusion depth.
\newblock https://github.com/Stability-AI/stablediffusion, 2022.

\bibitem{balaji2022eDiff-I}
Yogesh Balaji, Seungjun Nah, Xun Huang, Arash Vahdat, Jiaming Song, Karsten
  Kreis, Miika Aittala, Timo Aila, Samuli Laine, Bryan Catanzaro, Tero Karras,
  and Ming-Yu Liu.
\newblock ediff-i: Text-to-image diffusion models with ensemble of expert
  denoisers.
\newblock {\em arXiv preprint arXiv:2211.01324}, 2022.

\bibitem{bansal2023cold}
Arpit Bansal, Eitan Borgnia, Hong-Min Chu, Jie~S. Li, Hamid Kazemi, Furong
  Huang, Micah Goldblum, Jonas Geiping, and Tom Goldstein.
\newblock Cold diffusion: Inverting arbitrary image transforms without noise,
  2023.

\bibitem{bar2022text2live}
Omer Bar-Tal, Dolev Ofri-Amar, Rafail Fridman, Yoni Kasten, and Tali Dekel.
\newblock Text2live: Text-driven layered image and video editing.
\newblock In {\em European Conference on Computer Vision}, pages 707--723.
  Springer, 2022.

\bibitem{brooks2022generating}
Tim Brooks, Janne Hellsten, Miika Aittala, Ting-Chun Wang, Timo Aila, Jaakko
  Lehtinen, Ming-Yu Liu, Alexei~A Efros, and Tero Karras.
\newblock Generating long videos of dynamic scenes.
\newblock 2022.

\bibitem{brown2020gpt3}
Tom Brown, Benjamin Mann, Nick Ryder, Melanie Subbiah, Jared~D Kaplan, Prafulla
  Dhariwal, Arvind Neelakantan, Pranav Shyam, Girish Sastry, Amanda Askell,
  Sandhini Agarwal, Ariel Herbert-Voss, Gretchen Krueger, Tom Henighan, Rewon
  Child, Aditya Ramesh, Daniel Ziegler, Jeffrey Wu, Clemens Winter, Chris
  Hesse, Mark Chen, Eric Sigler, Mateusz Litwin, Scott Gray, Benjamin Chess,
  Jack Clark, Christopher Berner, Sam McCandlish, Alec Radford, Ilya Sutskever,
  and Dario Amodei.
\newblock Language models are few-shot learners.
\newblock In H. Larochelle, M. Ranzato, R. Hadsell, M.F. Balcan, and H. Lin,
  editors, {\em Advances in Neural Information Processing Systems}, volume~33,
  pages 1877--1901. Curran Associates, Inc., 2020.

\bibitem{chen2017coherent}
Dongdong Chen, Jing Liao, Lu Yuan, Nenghai Yu, and Gang Hua.
\newblock Coherent online video style transfer.
\newblock In {\em Proceedings of the IEEE International Conference on Computer
  Vision}, pages 1105--1114, 2017.

\bibitem{deforum}
deforum.
\newblock Deforum stable diffusion.
\newblock https://github.com/deforum/stable-diffusion, 2022.

\bibitem{ding2021cogview}
Ming Ding, Zhuoyi Yang, Wenyi Hong, Wendi Zheng, Chang Zhou, Da Yin, Junyang
  Lin, Xu Zou, Zhou Shao, Hongxia Yang, and Jie Tang.
\newblock Cogview: Mastering text-to-image generation via transformers.
\newblock In M. Ranzato, A. Beygelzimer, Y. Dauphin, P.S. Liang, and J.~Wortman
  Vaughan, editors, {\em Advances in Neural Information Processing Systems},
  volume~34, pages 19822--19835. Curran Associates, Inc., 2021.

\bibitem{ge2022long}
Songwei Ge, Thomas Hayes, Harry Yang, Xi Yin, Guan Pang, David Jacobs, Jia-Bin
  Huang, and Devi Parikh.
\newblock Long video generation with time-agnostic vqgan and time-sensitive
  transformer.
\newblock {\em arXiv preprint arXiv:2204.03638}, 2022.

\bibitem{goodfellow2014gans}
Ian Goodfellow, Jean Pouget-Abadie, Mehdi Mirza, Bing Xu, David Warde-Farley,
  Sherjil Ozair, Aaron Courville, and Yoshua Bengio.
\newblock Generative adversarial nets.
\newblock In Z. Ghahramani, M. Welling, C. Cortes, N. Lawrence, and K.Q.
  Weinberger, editors, {\em Advances in Neural Information Processing Systems},
  volume~27. Curran Associates, Inc., 2014.

\bibitem{ho2022imagenvideo}
Jonathan Ho, William Chan, Chitwan Saharia, Jay Whang, Ruiqi Gao, Alexey
  Gritsenko, Diederik~P Kingma, Ben Poole, Mohammad Norouzi, David~J Fleet,
  et~al.
\newblock Imagen video: High definition video generation with diffusion models.
\newblock {\em arXiv preprint arXiv:2210.02303}, 2022.

\bibitem{ho2020denoising}
Jonathan Ho, Ajay Jain, and Pieter Abbeel.
\newblock Denoising diffusion probabilistic models.
\newblock In H. Larochelle, M. Ranzato, R. Hadsell, M.F. Balcan, and H. Lin,
  editors, {\em Advances in Neural Information Processing Systems}, volume~33,
  pages 6840--6851. Curran Associates, Inc., 2020.

\bibitem{ho2022cascaded}
Jonathan Ho, Chitwan Saharia, William Chan, David~J Fleet, Mohammad Norouzi,
  and Tim Salimans.
\newblock Cascaded diffusion models for high fidelity image generation.
\newblock {\em J. Mach. Learn. Res.}, 23:47--1, 2022.

\bibitem{ho2022classifier}
Jonathan Ho and Tim Salimans.
\newblock Classifier-free diffusion guidance, 2022.

\bibitem{ho2022video}
Jonathan Ho, Tim Salimans, Alexey Gritsenko, William Chan, Mohammad Norouzi,
  and David~J Fleet.
\newblock Video diffusion models.
\newblock {\em arXiv:2204.03458}, 2022.

\bibitem{hong2022cogvideo}
Wenyi Hong, Ming Ding, Wendi Zheng, Xinghan Liu, and Jie Tang.
\newblock Cogvideo: Large-scale pretraining for text-to-video generation via
  transformers.
\newblock {\em arXiv preprint arXiv:2205.15868}, 2022.

\bibitem{isola2017image}
Phillip Isola, Jun-Yan Zhu, Tinghui Zhou, and Alexei~A Efros.
\newblock Image-to-image translation with conditional adversarial networks.
\newblock In {\em Proceedings of the IEEE conference on computer vision and
  pattern recognition}, pages 1125--1134, 2017.

\bibitem{Karras2022edm}
Tero Karras, Miika Aittala, Timo Aila, and Samuli Laine.
\newblock Elucidating the design space of diffusion-based generative models.
\newblock In {\em Proc. NeurIPS}, 2022.

\bibitem{kasten2021layered}
Yoni Kasten, Dolev Ofri, Oliver Wang, and Tali Dekel.
\newblock Layered neural atlases for consistent video editing.
\newblock {\em ACM Transactions on Graphics (TOG)}, 40(6):1--12, 2021.

\bibitem{kong2021fast}
Zhifeng Kong and Wei Ping.
\newblock On fast sampling of diffusion probabilistic models.
\newblock {\em arXiv preprint arXiv:2106.00132}, 2021.

\bibitem{li2022blip}
Junnan Li, Dongxu Li, Caiming Xiong, and Steven Hoi.
\newblock Blip: Bootstrapping language-image pre-training for unified
  vision-language understanding and generation.
\newblock In {\em ICML}, 2022.

\bibitem{lu2022dpmsolver}
Cheng Lu, Yuhao Zhou, Fan Bao, Jianfei Chen, Chongxuan Li, and Jun Zhu.
\newblock {DPM}-solver: A fast {ODE} solver for diffusion probabilistic model
  sampling in around 10 steps.
\newblock In Alice~H. Oh, Alekh Agarwal, Danielle Belgrave, and Kyunghyun Cho,
  editors, {\em Advances in Neural Information Processing Systems}, 2022.

\bibitem{lugmayr2022repaint}
Andreas Lugmayr, Martin Danelljan, Andres Romero, Fisher Yu, Radu Timofte, and
  Luc Van~Gool.
\newblock Repaint: Inpainting using denoising diffusion probabilistic models.
\newblock In {\em Proceedings of the IEEE/CVF Conference on Computer Vision and
  Pattern Recognition}, pages 11461--11471, 2022.

\bibitem{meng2021sdedit}
Chenlin Meng, Yang Song, Jiaming Song, Jiajun Wu, Jun{-}Yan Zhu, and Stefano
  Ermon.
\newblock Sdedit: Image synthesis and editing with stochastic differential
  equations.
\newblock {\em CoRR}, abs/2108.01073, 2021.

\bibitem{nichol2021improved}
Alexander~Quinn Nichol and Prafulla Dhariwal.
\newblock Improved denoising diffusion probabilistic models.
\newblock In {\em International Conference on Machine Learning}, pages
  8162--8171. PMLR, 2021.

\bibitem{nichol22glide}
Alexander~Quinn Nichol, Prafulla Dhariwal, Aditya Ramesh, Pranav Shyam, Pamela
  Mishkin, Bob Mcgrew, Ilya Sutskever, and Mark Chen.
\newblock {GLIDE}: Towards photorealistic image generation and editing with
  text-guided diffusion models.
\newblock In Kamalika Chaudhuri, Stefanie Jegelka, Le Song, Csaba Szepesvari,
  Gang Niu, and Sivan Sabato, editors, {\em Proceedings of the 39th
  International Conference on Machine Learning}, volume 162 of {\em Proceedings
  of Machine Learning Research}, pages 16784--16804. PMLR, 17--23 Jul 2022.

\bibitem{nikankin2022sinfusion}
Yaniv Nikankin, Niv Haim, and Michal Irani.
\newblock Sinfusion: Training diffusion models on a single image or video.
\newblock {\em arXiv preprint arXiv:2211.11743}, 2022.

\bibitem{peebles2022transformerdiffusion}
William Peebles and Saining Xie.
\newblock Scalable diffusion models with transformers, 2022.

\bibitem{davis2017}
Jordi Pont-Tuset, Federico Perazzi, Sergi Caelles, Pablo Arbel\'aez, Alexander
  Sorkine-Hornung, and Luc {Van Gool}.
\newblock The 2017 davis challenge on video object segmentation.
\newblock {\em arXiv:1704.00675}, 2017.

\bibitem{radford2021learning}
Alec Radford, Jong~Wook Kim, Chris Hallacy, Aditya Ramesh, Gabriel Goh,
  Sandhini Agarwal, Girish Sastry, Amanda Askell, Pamela Mishkin, Jack Clark,
  et~al.
\newblock Learning transferable visual models from natural language
  supervision.
\newblock In {\em International Conference on Machine Learning}, pages
  8748--8763. PMLR, 2021.

\bibitem{raffel:2020}
Colin Raffel, Noam Shazeer, Adam Roberts, Katherine Lee, Sharan Narang, Michael
  Matena, Yanqi Zhou, Wei Li, and Peter~J. Liu.
\newblock Exploring the limits of transfer learning with a unified text-to-text
  transformer.
\newblock {\em J. Mach. Learn. Res.}, 21(1), jun 2022.

\bibitem{ramesh2022dalle}
Aditya Ramesh, Prafulla Dhariwal, Alex Nichol, Casey Chu, and Mark Chen.
\newblock Hierarchical text-conditional image generation with clip latents,
  2022.

\bibitem{ramesh2021dalle}
Aditya Ramesh, Mikhail Pavlov, Gabriel Goh, Scott Gray, Chelsea Voss, Alec
  Radford, Mark Chen, and Ilya Sutskever.
\newblock Zero-shot text-to-image generation.
\newblock In Marina Meila and Tong Zhang, editors, {\em Proceedings of the 38th
  International Conference on Machine Learning}, volume 139 of {\em Proceedings
  of Machine Learning Research}, pages 8821--8831. PMLR, 18--24 Jul 2021.

\bibitem{Ranftl2019TowardsRM}
Ren{\'e} Ranftl, Katrin Lasinger, David Hafner, Konrad Schindler, and Vladlen
  Koltun.
\newblock Towards robust monocular depth estimation: Mixing datasets for
  zero-shot cross-dataset transfer.
\newblock {\em IEEE Transactions on Pattern Analysis and Machine Intelligence},
  44:1623--1637, 2019.

\bibitem{ravacha2008unwrap}
Alex Rav-Acha, Pushmeet Kohli, Carsten Rother, and Andrew~William Fitzgibbon.
\newblock Unwrap mosaics: a new representation for video editing.
\newblock {\em ACM SIGGRAPH 2008 papers}, 2008.

\bibitem{rombach2021highresolution}
Robin Rombach, Andreas Blattmann, Dominik Lorenz, Patrick Esser, and Björn
  Ommer.
\newblock High-resolution image synthesis with latent diffusion models, 2021.

\bibitem{ronneberger2015u}
Olaf Ronneberger, Philipp Fischer, and Thomas Brox.
\newblock U-net: Convolutional networks for biomedical image segmentation.
\newblock In {\em International Conference on Medical image computing and
  computer-assisted intervention}, pages 234--241. Springer, 2015.

\bibitem{ruder2016styletransfer}
Manuel Ruder, Alexey Dosovitskiy, and Thomas Brox.
\newblock Artistic style transfer for videos.
\newblock In Bodo Rosenhahn and Bjoern Andres, editors, {\em Pattern
  Recognition}, pages 26--36, Cham, 2016. Springer International Publishing.

\bibitem{ruiz2022dreambooth}
Nataniel Ruiz, Yuanzhen Li, Varun Jampani, Yael Pritch, Michael Rubinstein, and
  Kfir Aberman.
\newblock Dreambooth: Fine tuning text-to-image diffusion models for
  subject-driven generation.
\newblock {\em arXiv preprint arXiv:2208.12242}, 2022.

\bibitem{sxela2022warpfusion}
Alexander S.
\newblock Disco diffusion v5.2 - warp fusion.
\newblock https://github.com/Sxela/DiscoDiffusion-Warp, 2022.

\bibitem{saharia2021palette}
Chitwan Saharia, William Chan, Huiwen Chang, Chris~A. Lee, Jonathan Ho, Tim
  Salimans, David~J. Fleet, and Mohammad Norouzi.
\newblock Palette: Image-to-image diffusion models, 2021.

\bibitem{saharia2022photorealistic}
Chitwan Saharia, William Chan, Saurabh Saxena, Lala Li, Jay Whang, Emily
  Denton, Seyed Kamyar~Seyed Ghasemipour, Burcu~Karagol Ayan, S~Sara Mahdavi,
  Rapha~Gontijo Lopes, et~al.
\newblock Photorealistic text-to-image diffusion models with deep language
  understanding.
\newblock {\em arXiv preprint arXiv:2205.11487}, 2022.

\bibitem{saito2017temporal}
Masaki Saito, Eiichi Matsumoto, and Shunta Saito.
\newblock Temporal generative adversarial nets with singular value clipping.
\newblock In {\em Proceedings of the IEEE international conference on computer
  vision}, pages 2830--2839, 2017.

\bibitem{salimans2022progressive}
Tim Salimans and Jonathan Ho.
\newblock Progressive distillation for fast sampling of diffusion models.
\newblock In {\em International Conference on Learning Representations}, 2022.

\bibitem{san2021noise}
Robin San-Roman, Eliya Nachmani, and Lior Wolf.
\newblock Noise estimation for generative diffusion models.
\newblock {\em arXiv preprint arXiv:2104.02600}, 2021.

\bibitem{schuhmann2021laion}
Christoph Schuhmann, Richard Vencu, Romain Beaumont, Robert Kaczmarczyk,
  Clayton Mullis, Aarush Katta, Theo Coombes, Jenia Jitsev, and Aran
  Komatsuzaki.
\newblock Laion-400m: Open dataset of clip-filtered 400 million image-text
  pairs.
\newblock {\em arXiv preprint arXiv:2111.02114}, 2021.

\bibitem{Singer2022MakeAVideoTG}
Uriel Singer, Adam Polyak, Thomas Hayes, Xiaoyue Yin, Jie An, Songyang Zhang,
  Qiyuan Hu, Harry Yang, Oron Ashual, Oran Gafni, Devi Parikh, Sonal Gupta, and
  Yaniv Taigman.
\newblock Make-a-video: Text-to-video generation without text-video data.
\newblock {\em ArXiv}, abs/2209.14792, 2022.

\bibitem{skorokhodov2021styleganv}
Ivan Skorokhodov, Sergey Tulyakov, and Mohamed Elhoseiny.
\newblock Stylegan-v: A continuous video generator with the price, image
  quality and perks of stylegan2, 2021.

\bibitem{dickstein2015}
Jascha Sohl-Dickstein, Eric Weiss, Niru Maheswaranathan, and Surya Ganguli.
\newblock Deep unsupervised learning using nonequilibrium thermodynamics.
\newblock In Francis Bach and David Blei, editors, {\em Proceedings of the 32nd
  International Conference on Machine Learning}, volume~37 of {\em Proceedings
  of Machine Learning Research}, pages 2256--2265, Lille, France, 07--09 Jul
  2015. PMLR.

\bibitem{song2021denoising}
Jiaming Song, Chenlin Meng, and Stefano Ermon.
\newblock Denoising diffusion implicit models.
\newblock In {\em International Conference on Learning Representations}, 2021.

\bibitem{song2020score}
Yang Song, Jascha Sohl-Dickstein, Diederik~P Kingma, Abhishek Kumar, Stefano
  Ermon, and Ben Poole.
\newblock Score-based generative modeling through stochastic differential
  equations.
\newblock {\em arXiv preprint arXiv:2011.13456}, 2020.

\bibitem{tevet2022human}
Guy Tevet, Sigal Raab, Brian Gordon, Yonatan Shafir, Amit~H Bermano, and Daniel
  Cohen-Or.
\newblock Human motion diffusion model.
\newblock {\em arXiv preprint arXiv:2209.14916}, 2022.

\bibitem{texler2020videostylization}
Ond\v{r}ej Texler, David Futschik, Michal Ku\v{c}era, Ond\v{r}ej Jamri\v{s}ka,
  \v{S}\'{a}rka Sochorov\'{a}, Menglei Chai, Sergey Tulyakov, and Daniel
  S\'{y}kora.
\newblock Interactive video stylization using few-shot patch-based training.
\newblock {\em ACM Transactions on Graphics}, 39(4):73, 2020.

\bibitem{Tulyakov:2018:MoCoGAN}
Sergey Tulyakov, Ming-Yu Liu, Xiaodong Yang, and Jan Kautz.
\newblock {MoCoGAN}: Decomposing motion and content for video generation.
\newblock In {\em IEEE Conference on Computer Vision and Pattern Recognition
  (CVPR)}, pages 1526--1535, 2018.

\bibitem{vahdat2021score}
Arash Vahdat, Karsten Kreis, and Jan Kautz.
\newblock Score-based generative modeling in latent space.
\newblock {\em Advances in Neural Information Processing Systems},
  34:11287--11302, 2021.

\bibitem{villegas2022phenaki}
Ruben Villegas, Mohammad Babaeizadeh, Pieter-Jan Kindermans, Hernan Moraldo,
  Han Zhang, Mohammad~Taghi Saffar, Santiago Castro, Julius Kunze, and Dumitru
  Erhan.
\newblock Phenaki: Variable length video generation from open domain textual
  description, 2022.

\bibitem{vondrick2016generating}
Carl Vondrick, Hamed Pirsiavash, and Antonio Torralba.
\newblock Generating videos with scene dynamics.
\newblock {\em Advances in neural information processing systems}, 29, 2016.

\bibitem{wang2018fewshotvid2vid}
Ting-Chun Wang, Ming-Yu Liu, Andrew Tao, Guilin Liu, Jan Kautz, and Bryan
  Catanzaro.
\newblock Few-shot video-to-video synthesis.
\newblock In {\em Advances in Neural Information Processing Systems (NeurIPS)},
  2019.

\bibitem{Wang:2018:vid2vid}
Ting-Chun Wang, Ming-Yu Liu, Jun-Yan Zhu, Guilin Liu, Andrew Tao, Jan Kautz,
  and Bryan Catanzaro.
\newblock Video-to-video synthesis.
\newblock In {\em Conference on Neural Information Processing Systems
  (NeurIPS)}, 2018.

\bibitem{wang2018pix2pixHD}
Ting-Chun Wang, Ming-Yu Liu, Jun-Yan Zhu, Andrew Tao, Jan Kautz, and Bryan
  Catanzaro.
\newblock High-resolution image synthesis and semantic manipulation with
  conditional gans.
\newblock In {\em Proceedings of the IEEE Conference on Computer Vision and
  Pattern Recognition}, 2018.

\bibitem{wu2022tuneavideo}
Jay~Zhangjie Wu, Yixiao Ge, Xintao Wang, Stan~Weixian Lei, Yuchao Gu, Wynne
  Hsu, Ying Shan, Xiaohu Qie, and Mike~Zheng Shou.
\newblock Tune-a-video: One-shot tuning of image diffusion models for
  text-to-video generation.
\newblock {\em arXiv preprint arXiv:2212.11565}, 2022.

\bibitem{yan2021videogpt}
Wilson Yan, Yunzhi Zhang, Pieter Abbeel, and Aravind Srinivas.
\newblock Videogpt: Video generation using vq-vae and transformers.
\newblock {\em arXiv preprint arXiv:2104.10157}, 2021.

\bibitem{yu2022parti}
Jiahui Yu, Yuanzhong Xu, Jing~Yu Koh, Thang Luong, Gunjan Baid, Zirui Wang,
  Vijay Vasudevan, Alexander Ku, Yinfei Yang, Burcu~Karagol Ayan, Ben
  Hutchinson, Wei Han, Zarana Parekh, Xin Li, Han Zhang, Jason Baldridge, and
  Yonghui Wu.
\newblock Scaling autoregressive models for content-rich text-to-image
  generation, 2022.

\bibitem{zeng2022lion}
Xiaohui Zeng, Arash Vahdat, Francis Williams, Zan Gojcic, Or Litany, Sanja
  Fidler, and Karsten Kreis.
\newblock Lion: Latent point diffusion models for 3d shape generation.
\newblock In {\em Advances in Neural Information Processing Systems (NeurIPS)},
  2022.

\bibitem{zhou2022magicvideo}
Daquan Zhou, Weimin Wang, Hanshu Yan, Weiwei Lv, Yizhe Zhu, and Jiashi Feng.
\newblock Magicvideo: Efficient video generation with latent diffusion models,
  2022.

\end{thebibliography}
